%% file: bare_jrnl.tex
\documentclass[journal]{IEEEtran}

\ifCLASSINFOpdf
\else
\fi

\newif\ifguidelines
\guidelinestrue   %

\ifguidelines
    \newcommand{\guide}[1]{\textcolor{blue}{\textbf{[Guideline:} #1\textbf{]}}}
\else
    \newcommand{\guide}[1]{}
\fi

\usepackage{tikz}
\usetikzlibrary{arrows.meta, positioning, shapes.geometric, fit}
\usepackage{booktabs}
\usepackage{multirow}
\usepackage{array}
\usepackage{todonotes}
\usepackage{amssymb}
\usepackage{amsfonts}

\usepackage{tabularx}
\usepackage{xcolor}
\newcommand{\rev}[1]{#1}
\usepackage[utf8]{inputenc}

\usetikzlibrary{decorations.pathreplacing}
\usetikzlibrary{shapes.geometric, arrows, positioning, fit}

\usetikzlibrary{backgrounds, calc, decorations.pathmorphing}
\usepackage{pifont}
\newcommand{\cmark}{\ding{51}} %
\newcommand{\xmark}{\ding{55}} %
\usepackage{amsmath}

\usepackage{hyperref}
\usepackage{tcolorbox}
\usepackage{tabularx}
\usepackage{array}

\usepackage[normalem]{ulem}
\usepackage{colortbl}
\newif\ifshowchanges
\showchangestrue
\ifshowchanges
  
  \newcommand{\deleted}[1]{{\color{red}\sout{#1}}}

\else
  
  \newcommand{\deleted}[1]{}

\fi

\usepackage{enumitem}
\usepackage[acronym,nonumberlist]{glossaries}
\makeglossaries

\newacronym{ai}{AI}{Artificial Intelligence}
\newacronym{fm}{FM}{Foundation Models}
\newacronym{llm}{LLM}{Large Language Model}
\newacronym{wm}{WM}{World Models}
\newacronym{dt}{DT}{Digital Twins}
\newacronym{rl}{RL}{Reinforcement Learning}
\newacronym{drl}{DRL}{Deep Reinforcement Learning}
\newacronym{marl}{MARL}{Multi-Agent Reinforcement Learning}
\newacronym{fl}{FL}{Federated Learning}
\newacronym{opt}{OPT}{Optimization}
\newacronym{grpo}{GRPO}{Group Relative Policy Optimization}
\newacronym{rag}{RAG}{Retrieval-Augmented Generation}
\newacronym{ppo}{PPO}{Proximal Policy Optimization }
\newacronym{randr}{R\&R}{Representation \& Reasoning}
\newacronym{pands}{P\&S}{Prediction \& Simulation}
\newacronym{ac}{AC}{Adaptive Control}
\newacronym{ds}{DS}{Decision Synthesis}

\newacronym{arc}{ARC}{Autonomous Reinforcement Coordination}
\newacronym{fm-rl}{FM-RL}{Foundation Model--Reinforcement Learning}

\newacronym{ran}{RAN}{Radio Access Network}
\newacronym{rrm}{RRM}{Radio Resource Management}
\newacronym{ric}{RIC}{RAN Intelligent Controller}
\newacronym{son}{SON}{Self-Organizing Network}
\newacronym{ru}{RU}{Radio Unit}
\newacronym{du}{DU}{Distributed Unit}
\newacronym{cu}{CU}{Central Unit}
\newacronym{phy}{PHY}{Physical Layer}
\newacronym{mac}{MAC}{Medium Access Control}
\newacronym{rrc}{RRC}{Radio Resource Control}

\newacronym{qos}{QoS}{Quality of Service}
\newacronym{sla}{SLA}{Service Level Agreement}
\newacronym{urllc}{URLLC}{Ultra-Reliable Low-Latency Communications}
\newacronym{mmtc}{mMTC}{Massive Machine-Type Communications}
\newacronym{embb}{eMBB}{Enhanced Mobile Broadband}
\newacronym{kpi}{KPIs}{Key Performance Indicators}
\newacronym{eeirp}{EEIRP}{Expected Equivalent Isotropically Radiated Power}

\begin{document}
\title{
Hybrid AI for AI-Native RAN Control: A Functional Perspective
}

\author{ \IEEEauthorblockN{ Sabrine Aroua\IEEEauthorrefmark{1}, Osama Arouk\IEEEauthorrefmark{1}, Pegah Alizadeh\IEEEauthorrefmark{2} } \\ \IEEEauthorblockA{ \IEEEauthorrefmark{1} Ericsson Research, France\\ Email: \{sabrine.aroua, osama.arouk\}@ericsson.com\\ \IEEEauthorrefmark{2} Université Lumière Lyon 2, Université Claude Bernard Lyon 1, ERIC, 69007, Lyon, France\\ Email: pegah.alizadeh@univ-lyon2.fr} }

\markboth{}{}

\maketitle

\begin{abstract}
AI-native radio access network (RAN) control involves interdependent decisions across network entities, protocol layers, and timescales.
Understanding how AI and conventional methods can jointly support these decisions requires a common basis for comparing their functional roles. This survey introduces a technique-agnostic taxonomy of RAN decision-making organized around four complementary capabilities: Representation and Reasoning, Prediction and Simulation, Adaptive Control, and Decision Synthesis. These capabilities are defined by their roles in the control process, independently of the techniques used to implement them. Through this perspective, we examine representative AI and model-based approaches, including foundation models, world models, reinforcement learning, and optimization. We then apply the taxonomy to representative RAN control solutions across slow, medium, fast, and cross-timescale applications to identify which capabilities they implement and how these capabilities are combined in hybrid AI systems. We also review datasets, benchmarks,
simulators, digital twins, and experimental platforms according to their support for the decision-making process and reproducible evaluation. The review highlights a gap between the availability of resources for closed-loop experimentation and the standardization of tasks, metrics, and evaluation protocols needed for meaningful comparisons. Finally, we identify research directions for hybrid AI in AI-native RAN, including capability integration, multi-agent and cross-timescale coordination, constraint handling, and reproducible closed-loop evaluation. A curated list of surveyed works and evaluation
resources is available at \url{https://github.com/pegahani/Hybrid-AI-RAN-Survey}.

\end{abstract}


\IEEEpeerreviewmaketitle

\section{Introduction}
\label{sec:introduction}
\input{sections/introduction/main}

\section{AI and Model-Based Techniques for Decision Making}
\label{sec:general_view}
\input{sections/general_view/main2}

\section{Decision-Making Requirements for AI-Native RAN Control}
\label{sec:architecture}
\input{sections/architecture/main}

\section{Functional Building Blocks of Hybrid AI for RAN Control}
\label{sec:functionalBlocks}

\input{sections/functional_block_new/main}

\input{sections/multi_timescale_applications/main}

\input{sections/benchmarks_datasets/main2}

\input{sections/future_directions/main2}

\section{Conclusion} \label{sec:conclusion} 

This survey presented a functional perspective on AI-native RAN control through four complementary capabilities: Representation and Reasoning (R\&R), Prediction and Simulation (P\&S), Adaptive Control (AC), and Decision Synthesis (DS). This framework provides a common lens for understanding how foundation models, world models, reinforcement learning, optimization, and related approaches contribute to autonomous network operation and complement one another within the overall decision-making process. 

Our analysis revealed an uneven distribution of capabilities across the current research landscape. While Adaptive Control has received significant attention, Prediction and Simulation and Decision Synthesis remain comparatively underexplored. Moreover, existing solutions largely address individual control functions in isolation, providing limited support for coordinated intelligence across multiple agents, protocol layers, and timescales. A related challenge lies in evaluation, where reproducible methodologies for assessing integrated closed-loop RAN control remain limited despite the growing availability of datasets, simulators, digital twins, and experimental platforms. 

These observations suggest that the key challenge is no longer the development of individual AI techniques, but their integration into coherent decision-making systems capable of reasoning, predicting, adapting, and acting under operational constraints. The proposed framework provides a foundation for analyzing, designing, and evaluating such systems, contributing to the broader vision of proactive, trustworthy, and scalable AI-native RANs capable of autonomous operation in increasingly complex network environments.

\section*{Acknowledgment}
The authors used ChatGPT (OpenAI) and Claude (Anthropic) to assist in searching for and collecting relevant literature, improving the language and clarity of the manuscript, checking consistency across revisions, and preparing some of the figures. The ideas, mathematical formulations, and modeling presented in this article are entirely the authors' own. All AI-assisted content, including the selection and accuracy of references, was carefully reviewed, verified, and revised by the authors, who take full responsibility for the content of this article.

\ifCLASSOPTIONcaptionsoff
  \newpage
\fi

\appendix
\section{Acronyms} \label{app:acronyms} %
\printglossary[ type=\acronymtype]

\bibliographystyle{IEEEtran}
\bibliography{IEEEabrv,references}

\end{document}

%% file: sections/introduction/main.tex
\subsection{Motivation}
\input{sections/introduction/motivation_f}
\subsection{Comparison with existing surveys}
\label{existingSurvey}
\input{sections/introduction/existingSurvey2}

\subsection{Scope, contributions, and organization}
\label{contribution}
\input{sections/introduction/unique_contributions}

%% file: sections/introduction/motivation_f.tex
\label{motivation}
The control and optimization of \gls{ran}s is a central challenge in wireless systems \cite{luong2019applications, chen2020reinforcement}.
In 5G networks, \gls{ran}s must support services with distinct and potentially competing requirements, ranging from the ultra-reliability and low latency of \gls{urllc} to the massive connectivity of \gls{mmtc} and the dynamic traffic demands of \gls{embb} \cite{popovski20185g, parvez2018survey}. Meeting these requirements calls for coordinating heterogeneous resources and control objectives
across services while satisfying latency, reliability, and feasibility constraints \cite{olwal2016survey}. Rule-based control, \gls{son} functions, and task-specific optimization have played a key role in 4G and 5G network management. However, these mechanisms become
harder to maintain and coordinate as service requirements and control objectives evolve, and actions in one control loop affect the operating conditions of others.

The increasing programmability of the RAN creates opportunities to
address these challenges through intelligent control applications.
In particular, O-RAN enables applications to operate across control
timescales, increasing the need to manage their interactions
\cite{oran_architecture_2026}. Alongside this architectural evolution,
6G research advances the vision of AI-native networks, in which
\gls{ai} participates directly in decision-making processes distributed
across protocol layers, network entities, and control domains
\cite{mustafovski2026ai}. The breadth of these requirements motivates
hybrid \gls{ai} approaches that combine complementary AI and model-based
methods. This survey examines representative approaches, including
\gls{fm}, \gls{wm}, \gls{rl}, and \gls{opt}, according to how they
support the interpretation of network context, anticipation of
candidate outcomes, adaptation of control policies, and synthesis
of operationally feasible decisions.

This evolution gives RAN control an inherently multi-agent, cross-layer, and multi-timescale nature. Schedulers, controllers, and management functions operate across protocol layers, including \gls{phy}, \gls{mac}, and \gls{rrc}, relying on heterogeneous and often partial observations of the network state, such as traffic load, user mobility, and radio conditions \cite{chen2020reinforcement, ahmadi20195g, peterson20205g}. At the management level, \gls{son} functions can modify cell configurations, mobility parameters, and resource allocation policies, thereby influencing the behavior of lower-layer control mechanisms. Meanwhile, control decisions span a broad range of temporal scales, from microsecond-level scheduling and link adaptation to longer-term resource orchestration, network slicing, and energy optimization \cite{yan2025near, navidan2026toward}.

These dimensions are strongly interdependent. Decisions made by one agent can affect the observations, objectives, and feasible actions of others, while slower control loops shape the operating conditions under which faster loops execute. As a result, achieving efficient network operation requires more than isolated optimization of individual functions. 
It requires a systematic understanding of how complementary control capabilities can be coordinated across agents, protocol layers, and timescales; anticipating network evolution; adapting strategies to changing conditions; and maintaining coherent decision making across the RAN.

Taken together, these challenges define the central problem addressed in this survey: \textbf{Enabling RAN control systems to understand network behavior, anticipate future dynamics, adapt their control policies, and synthesize timely and feasible decisions across multiple agents, protocol layers, and temporal scales.}

\subsection{Research Gap}
\label{Research_Gap}

RAN control research has explored different aspects of the network decision-making process. Classical \gls{opt} selects actions subject to explicit constraints and can provide optimality guarantees under appropriate assumptions. \gls{rl} learns control policies through interaction with the environment, often targeting specific objectives and decision timescales \cite{heo2025dynamic, khetarpal2022continualrl, korkmaz2024survey}.
Hierarchical \gls{rl} introduces temporal abstraction across decision scales \cite{pateria2021hierarchical, machado2023temporal, yu2025reinforcement}, while multi-agent \gls{rl} addresses coordination among distributed agents operating under partial observability \cite{huh2023multi, ma2023hierarchical}. More recently, foundation models and world models have broadened the range of approaches for reasoning over heterogeneous network information and predicting network dynamics \cite{sun2025survey,zhang2026exploring}.

Existing surveys examine AI paradigms, RAN architectures, and control applications from complementary perspectives, with less emphasis on how techniques from different paradigms contribute to and interact within a RAN control loop \cite{yang2025decision, Polese_2023_Understanding_ORAN}. This makes it difficult to compare which capabilities existing systems implement, how they combine them, and how those combinations vary across control timescales.

To address this gap, we develop a functional taxonomy of RAN control capabilities and use it to classify representative studies according to the roles their methods perform within the control loop. By comparing these classifications across RAN applications, network functions, and control timescales, we identify recurring combinations of capabilities and analyze how they are integrated within hybrid AI solutions. This analysis provides a structured perspective on hybrid AI for AI-native RAN, clarifies how different AI paradigms contribute to the control process, and highlights open research questions regarding the integration of control capabilities across applications, agents, and timescales.

\subsection{Functional Decomposition of RAN Control}
\label{fromToMotovation2}

The perspective above suggests a shift in how RAN control should be conceptualized. An \gls{ai}-native RAN can be understood as a collection of interacting control loops that observe and act on a dynamic network environment.
Such a system must first perceive its environment by gathering and interpreting information from network entities, users, services, and the surrounding radio environment. Building on this situational awareness, it must reason about the observed state, anticipate future dynamics, adapt its behavior to changing conditions, and make decisions that influence subsequent observations through actuation.

In such environments, decisions made by one network entity can modify the operating conditions observed by others, while actions taken at one timescale can influence the objectives, constraints, and available resources of control processes operating at different temporal horizons. Intelligence therefore emerges not from an individual control algorithm, but from the coordinated interaction of perception, reasoning, prediction, adaptation, and decision-making capabilities operating over shared and evolving network dynamics. These interactions motivate an analysis of how complementary capabilities realized by AI and conventional methods contribute to RAN control across agents, protocol layers, and timescales.

This perspective motivates a functional decomposition of RAN intelligence into four complementary decision-making capabilities. These capabilities operate on information provided through network perception, which supplies awareness of network conditions, user behavior, service requirements, and environmental dynamics. Unlike the capabilities described below, perception is not treated as a decision-making function. Rather, it serves as a foundational enabler that provides the observations and contextual information upon which the decision-making process relies.

\gls{randr} establishes a coherent understanding of heterogeneous observations and contextual information; \gls{pands} anticipates network evolution and the consequences of alternative control strategies; \gls{ac} adapts control mechanisms to changing conditions and feedback; and \gls{ds} transforms available information and control mechanisms into operationally feasible decisions. These capabilities are not tied to specific technologies or necessarily implemented as independent modules. Rather, they define functional roles that can be realized through different, potentially overlapping AI and model-based paradigms.

\gls{fm} can support representation and reasoning over heterogeneous network information \cite{sun2025survey}, while \gls{wm} and \gls{dt} can provide predictive and simulation-based capabilities \cite{zhang2026exploring,duran2025toward}. \gls{rl} can support adaptive control through interaction and feedback, while optimization can provide explicit mechanisms for objective-driven adaptation and constraint-aware decision synthesis \cite{li2022applications,zhou2026convergence}. Importantly, these paradigms can contribute to multiple capabilities and may need to be combined to address the coupled nature of RAN control. Their composition and interaction, however, remain difficult to characterize within a common framework when viewed through technology-specific perspectives.

This survey therefore adopts a functional perspective and examines \textbf{how AI and model-based techniques contribute to the RAN decision-making process and how their capabilities are combined}. Section~\ref{sec:architecture} discusses the requirements for RAN control across agents, protocol layers, and timescales. Section~IV describes the four functional capabilities and illustrates how they can interact within and across control loops.

%% file: sections/introduction/existingSurvey2.tex
\begin{table*}[t]
\centering
\caption{Comparison of representative surveys on AI-native RAN decision-making.}
\label{tab:survey_comparison}
\footnotesize
\renewcommand{\arraystretch}{1.08}
\setlength{\tabcolsep}{3pt}

\begin{tabularx}{\textwidth}{@{}p{2.0cm}ccccXcccX@{}}
\toprule
\multirow{2}{*}{\textbf{Survey}}
& \multicolumn{4}{c}{\textbf{AI Paradigms}}
& \multirow{2}{*}{\shortstack[l]{\textbf{Primary}\\\textbf{Perspective}}}
& \multicolumn{3}{c}{\textbf{Decision-Making Aspects}}
& \multirow{2}{*}{\shortstack[l]{\textbf{Functional}\\\textbf{Taxonomy Scope}}} \\
\cmidrule(lr){2-5}\cmidrule(lr){7-9}
& \gls{fm} & \gls{wm} & \gls{rl} & \gls{opt}
& & \shortstack{Multi-\\Agent} & \shortstack{Multi-\\Timescale} & RAN & \\
\midrule

Cai et al.~\cite{Cai_2025_Tutorial_on_LLM_RL_Wireless}
& \cmark & \xmark & \cmark & \xmark
& LLM-enhanced wireless control
& \xmark & \xmark & Partial
& LLM roles in RL workflows \\

Yang et al.~\cite{yang2025decision}
& \cmark & \xmark & \xmark & \xmark
& LLM-based wireless decision-making
& Partial & \xmark & \xmark
& Reward modeling for LLM-based decisions \\

Zhang et al.~\cite{Zhang_2026_Multi_Modal_FM_Survey}
& \cmark & \xmark & \xmark & \xmark
& Multimodal FMs for prediction and control
& \xmark & \xmark & Partial
& FM roles in wireless control \\

Mienye et al.~\cite{Mienye_2026_DRL_FM_Survey}
& \cmark & \xmark & \cmark & \xmark
& Integration of FMs and deep RL
& Partial & \xmark & \xmark
& Roles within FM--RL integration \\

Wei et al.~\cite{Wei_2025_LLM_For_Wireless_Management}
& \cmark & \xmark & \xmark & Partial
& LLM-based network management
& \xmark & \xmark & Partial
& Functions in LLM-enabled optimization workflows \\

Zhao et al.~\cite{zhao2025edge}
& \cmark & \cmark & \xmark & \xmark
& World models and agents for edge intelligence
& \xmark & \xmark & \xmark
& \textemdash \\

Liu et al.~\cite{liu2024survey}
& \xmark & \xmark & \xmark & \cmark
& Optimization methods for wireless systems
& Partial & \xmark & \xmark
& \textemdash \\

Barker et al.~\cite{barker2026ainative}
& \cmark & Partial & \cmark & \xmark
& AI-enabled Open RAN and RIC
& Partial & Partial & \cmark
& \textemdash \\

Ameur et al.~\cite{ameur2026agentic}
& \cmark & \xmark & \xmark & \xmark
& Agentic AI for 5G/6G
& \cmark & \xmark & Partial
& \textemdash \\

Polese et al.~\cite{Polese_2023_Understanding_ORAN}
& \xmark & \xmark & \xmark & \xmark
& O-RAN architecture and control loops
& \xmark & Partial & \cmark
& \textemdash \\

Belter et al.~\cite{belter2026assured}
& \xmark & \xmark & \xmark & \xmark
& Assurance of interacting control loops
& \cmark & \cmark & Partial
& \textemdash \\

Sahin et al.~\cite{sahin2026ainos}
& \cmark & \xmark & \xmark & \xmark
& AI-native network operating systems
& Partial & Partial & Partial
& \textemdash \\

\midrule
\textbf{This work}
& \cmark & \cmark & \cmark & \cmark
& \textbf{Hybrid AI in AI-native RAN}
& \cmark & \cmark & \cmark
& \textbf{Technique-agnostic taxonomy of RAN decision-making functions and their combinations} \\
\bottomrule
\end{tabularx}

\par\smallskip
\begin{minipage}{\textwidth}
\footnotesize
The \emph{Functional Taxonomy} column indicates which decision-making
functions each survey uses to classify reviewed studies. A dash denotes the absence of an explicit classification by decision-making function.
\end{minipage}
\end{table*}

Table~\ref{tab:survey_comparison} compares representative surveys
related to AI-native \gls{ran} decision-making. It summarizes their
coverage of AI paradigms, their treatment of multi-agent and
multi-timescale control, their relevance to the RAN, and the scope
of any functional taxonomy they apply. The surveyed works can be
grouped into four categories according to their primary perspective.

\begin{itemize}
    \item \textbf{AI paradigm-centric surveys.}
    These surveys examine \gls{fm}s and \gls{llm}s
    \cite{yang2025decision,Zhang_2026_Multi_Modal_FM_Survey,Wei_2025_LLM_For_Wireless_Management},
    the integration of \gls{fm}s and \gls{rl}
    \cite{Cai_2025_Tutorial_on_LLM_RL_Wireless,Mienye_2026_DRL_FM_Survey},
    \gls{wm}s \cite{zhao2025edge}, and optimization
    \cite{liu2024survey}. Some use functional classifications
    within specific workflows. For example, Cai et al. classify
    the roles of LLMs in RL workflows
    \cite{Cai_2025_Tutorial_on_LLM_RL_Wireless}, while Wei et al.
    organize LLM-enabled optimization around problem formulation,
    solution generation, and verification
    \cite{Wei_2025_LLM_For_Wireless_Management}.

    \item \textbf{Architecture-centric surveys.}
    These works examine AI-native RAN and O-RAN through their
    components, interfaces, and control loops, including how
    intelligent control applications are deployed across the RAN
    \cite{barker2026ainative,Polese_2023_Understanding_ORAN}.

    \item \textbf{Agentic and autonomous control.}
    Surveys in this category examine capabilities such as reasoning,
    planning, tool use, and multi-agent coordination, and map their
    applications to telecommunications domains and use cases
    \cite{ameur2026agentic}.

    \item \textbf{Control-loop governance and assurance.}
    These works examine how autonomous decisions and interacting
    control loops can be governed, verified, and assured during
    network operation \cite{belter2026assured,sahin2026ainos}.
\end{itemize}

The distinguishing feature of this survey is a technique-agnostic
functional taxonomy of RAN decision-making. The taxonomy takes the
decision-making process as its starting point and defines four
functions independently of the methods used to implement them.
We apply the same categories to representative RAN control
solutions to identify which functions they implement, which AI
or conventional methods perform those functions, and how the
functions are combined. This common basis for comparison lets us
identify recurring combinations of capabilities for hybrid AI in
AI-native RAN across applications and control timescales.

%% file: sections/introduction/unique_contributions.tex
Before defining the scope of this survey, it is important to clarify the meaning of AI-native \gls{ran}. The term is sometimes used to describe networks that provide communication and computing infrastructure for AI services, including distributed training, inference, and edge AI applications \cite{11320975}. In this survey, however, AI-native \gls{ran} refers to a network in which AI is an intrinsic component of the control process itself. The focus is therefore on AI-enabled network intelligence, where AI systems participate directly in monitoring, reasoning, prediction, adaptation, and decision making, rather than on supporting the execution of AI workloads. 

The main contributions of this survey are as follows:
\begin{itemize}

\item \textbf{A technique-agnostic functional taxonomy of RAN
decision-making.} We define four decision-making functions
and use them to characterize the roles of AI and conventional
methods in RAN control. The taxonomy accommodates different
methods that perform the same function and individual methods
that contribute to multiple functions.

\item \textbf{A formulation of distributed RAN decision-making.}
We formulate RAN control as a set of interacting decision
processes across network entities, protocol layers, and
timescales. The formulation describes how partial observations,
coupled network dynamics, and operational constraints shape
decisions within and across control loops.

\item \textbf{A functional analysis of RAN control applications.}
We apply the taxonomy to representative systems across slow-,
medium-, fast-, and cross-timescale settings
(Section~\ref{sec:applications}). We identify which functions
each system implements, which methods perform them, and how
the functions work together. Among the reviewed systems,
\gls{ac} appears more frequently than \gls{pands} and \gls{ds},
and explicit \gls{ds} is uncommon in fast-timescale applications.

\item \textbf{A review of resources for evaluating AI-native RAN
control.} We classify datasets, benchmarks, digital twins,
simulators, dynamics models, and interactive environments by
the parts of the decision-making process they support. We also
distinguish resources that enable closed-loop experiments from
benchmarks that specify common tasks, metrics, and evaluation
protocols.

\item \textbf{Open challenges and research directions.}
We identify research questions concerning cross-timescale
coordination, scalable multi-agent decision-making, safety and
governance, the integration of decision-making functions, and
reproducible closed-loop evaluation.

\end{itemize}

\begin{figure*}
\centering 
\includegraphics[width=1\linewidth]{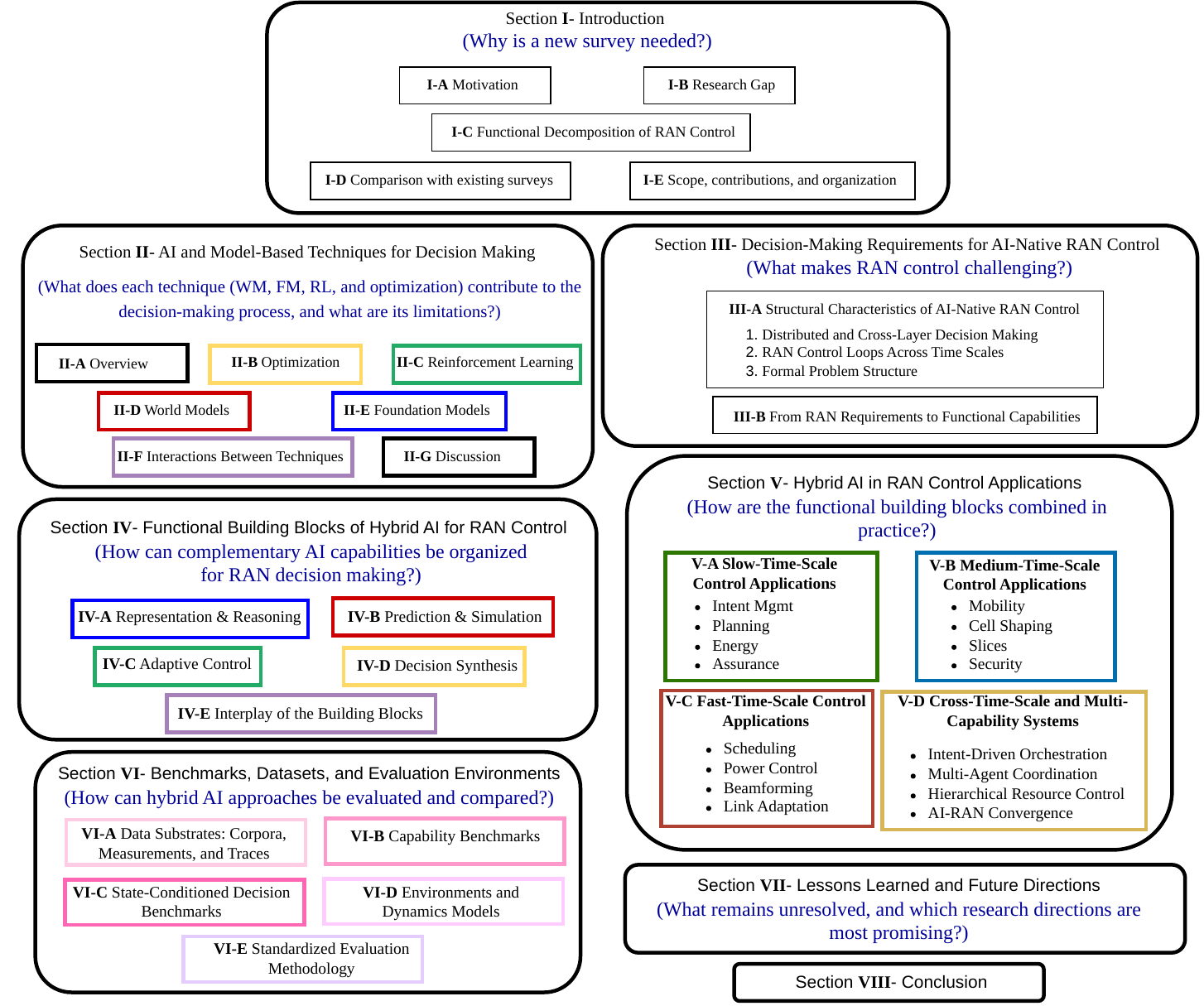} 
\caption{Survey roadmap illustrating the progression from AI and model-based decision-making techniques, to the requirements and functional capabilities of AI-native \gls{ran} control, their realization in representative applications, the associated evaluation ecosystem, and the resulting open research challenges.} 
\label{fig:survey_roadmap} 
\end{figure*}

Fig.~\ref{fig:survey_roadmap} shows how the survey is organized.
Sections~\ref{sec:general_view} and \ref{sec:architecture} review the techniques and requirements for RAN
control. Section~\ref{sec:functionalBlocks} describes the four capabilities of the functional taxonomy and explains how they interact. Section~\ref{sec:applications} applies the taxonomy to representative RAN control solutions across timescales. Sections~\ref{sec:benchmarksDatasets} and \ref{sec:lessonsLearned} review evaluation resources and discuss open research directions.

%% file: sections/general_view/main2.tex
\subsection{Overview}
\label{overveiw1}

Before turning to AI-native \gls{ran} control, we first examine the broader decision-making literature. A natural question is how major AI and model-based techniques, including \gls{opt}, \gls{rl}, \gls{wm}s, and \gls{fm}s, relate to one another and which combinations have already been explored.

Our objective is not to identify a universal architecture or to argue that all these techniques should be combined. Instead, we examine their respective roles in decision-making systems, their limitations when used in isolation, and how they have been combined in the literature. Table~\ref{tab:capabilities_complementarities} summarizes the primary contributions, limitations, and complementary roles of these techniques. This analysis provides the foundation for examining which combinations are most relevant to AI-native \gls{ran} control and why.

For each technique, we discuss two questions: (i) what does it contribute to the decision-making process, and (ii) what limitations prevent it from addressing the complexity of modern \gls{ran} control on its own?

\subsection{Optimization} \label{subsec:optimization}

\gls{opt} provides an explicit mechanism for selecting actions under objectives and constraints. Decision-making is formulated as the search for a feasible solution that optimizes a predefined objective while satisfying operational requirements~\cite{bertsekas2019reinforcement}. This capability makes optimization particularly attractive when decisions must satisfy strict resource, performance, or reliability constraints. 

Its effectiveness, however, depends on the availability of sufficiently accurate models of objectives, constraints, and system dynamics. In large-scale and dynamic environments, such models may be difficult to construct and maintain. Moreover, optimization does not inherently provide representation learning, reasoning over heterogeneous information, predictive modeling, or adaptation through interaction. In the \gls{ran}, it underpins classical interference-control and resource-allocation designs~\cite{peng2015recent}, whose models must be revisited whenever traffic, topology, or objectives change.

\subsection{Reinforcement Learning} 
\label{subsec:rl}

\gls{rl} addresses sequential decision-making by learning policies through interaction with the environment~\cite{sutton2018reinforcement,arulkumaran2017deep}. Rather than relying on a predefined decision model, it continuously improves behavior using feedback from past actions. This makes it particularly attractive in dynamic environments where system behavior is difficult to model explicitly.

Despite these advantages, \gls{rl} is often sample-inefficient, may struggle to satisfy hard constraints, and typically provides limited guarantees regarding safety, reliability, or feasibility. It also lacks native mechanisms for handling heterogeneous semantic inputs or exploiting rich prior knowledge. In the \gls{ran}, it has been applied to handover-parameter tuning, link adaptation and power control, and slicing~\cite{alizadeh2025offline,guo2023reinforcement,yan2025near}.

\subsection{World Models} 
\label{subsec:wm}

\gls{wm}s learn predictive representations of environment dynamics, enabling anticipation of future states and evaluation of the consequences of candidate actions \cite{ha2018worldmodels,hafner2020mastering,hafner2023mastering,Moerland_2023_Model_based_RL}. By explicitly representing how the environment evolves, they support planning, simulation, and look-ahead decision-making.

Their usefulness depends on the accuracy of the learned dynamics. Inaccurate models can lead to poor predictions and suboptimal decisions. Furthermore, while world models are effective for anticipating future outcomes, they do not naturally perform semantic reasoning, adapt policies through interaction, or enforce operational constraints. In the \gls{ran}, learned world models have recently been used to plan control decisions from historical network data and to train slice-allocation policies on imagined trajectories~\cite{bouton2026mobile,zhou2025world}.

\subsection{Foundation Models} 
\label{subsec:fm}

\gls{fm}s are pretrained on large-scale datasets to learn transferable representations and reasoning capabilities that can generalize across tasks \cite{bommasani2021opportunities}. Their ability to process heterogeneous and potentially multimodal information makes them particularly effective for understanding context, extracting structured representations, and supporting high-level reasoning.

However, foundation models do not inherently model environment dynamics, learn from closed-loop interaction, or guarantee that generated actions satisfy operational constraints. Their outputs may be semantically meaningful without being directly actionable or feasible. In the \gls{ran}, they have been used to process operator intents and to guide \gls{rl}-based slicing~\cite{Habib2025LlmBasedIntentProcessing,lotfi2025prompt}.

\subsection{Interactions Between Techniques} \label{subsec:technique_interactions}

\textbf{Have these techniques already been combined?} Yes. The literature already contains numerous examples of combining optimization, reinforcement learning, world models, and foundation models. However, these combinations are typically limited to a subset of techniques and are often driven by specific application needs rather than a holistic view of the decision-making process.

World models and reinforcement learning form the foundation of model-based \gls{rl}, where learned dynamics are used to improve planning and policy learning \cite{ha2018worldmodels,hafner2020mastering,hafner2023mastering,Moerland_2023_Model_based_RL}. Reinforcement learning and optimization have long been connected through optimal control and approximate dynamic programming \cite{bertsekas2019reinforcement}. Foundation models have recently been integrated with reinforcement learning, either by providing representations, initializing policies, or directly treating control as a sequence modeling problem \cite{reed2022gato,driess2023palme,Mienye_2026_DRL_FM_Survey,chen2021decisiontransformer}. 

Combinations involving foundation models and world models remain comparatively less mature, although recent work has explored transformer-based world models capable of learning predictive environment dynamics \cite{micheli2023transformers}. In an AI-native RAN workflow, such an interaction can take the following form: a \gls{fm} translates an operator intent into candidate configurations, and a \gls{wm} or digital twin predicts their impact on network performance before deployment, so that only configurations with acceptable predicted outcomes are enacted. Early instances of this pattern validate \gls{llm}-processed intents against predicted traffic~\cite{Habib2025LlmBasedIntentProcessing}, evaluate \gls{llm}-generated network designs through ray-tracing simulation~\cite{qiu2024large}, and use a multimodal \gls{wm} to generate imagined network trajectories~\cite{chai2025MobiWorld}. %
Optimization has also been integrated into learning systems through differentiable optimization layers \cite{amos2017optnet,kotary2021endtoend}, while foundation models are more commonly coupled to optimization through external solvers that enforce constraints or compute decisions \cite{sudhakara2025constrained,tan2025tool}. 

More tightly integrated combinations remain relatively rare. While some model-based reinforcement learning systems combine prediction, planning, and decision-making at runtime \cite{chua2018deep}, few works simultaneously exploit the complementary strengths of representation learning, predictive modeling, adaptive learning, and constrained decision-making. 

\subsection{Discussion} 
\label{subsec:discussion_general_view}

The preceding review shows that optimization, reinforcement learning, world models, and foundation models provide complementary capabilities for decision-making. However, the presence of multiple techniques in a system does not necessarily imply effective integration. Existing approaches often combine selected capabilities for specific tasks, while the allocation of responsibilities, the exchange of information, and the coordination of decisions remain dependent on the application and architecture.

For AI-native RAN control, this distinction is particularly important. Network operation requires decisions that account for changing network conditions, anticipate future outcomes, adapt to observed feedback, and satisfy operational constraints. Addressing these requirements involves not only selecting suitable AI techniques, but also determining how their capabilities should interact across decision-making functions and timescales.

This motivates the hybrid AI perspective adopted in this survey. Rather than treating optimization, reinforcement learning, world models, and foundation models as competing alternatives or assuming that they should always be combined, we examine how their complementary capabilities can be organized according to the functions they serve. The next section builds on this perspective to establish a functional view of hybrid AI for AI-native RAN control.

\begin{table*}[t]
    \centering
    \caption{Capabilities and complementarities of AI and model-based techniques for decision making.}
    \label{tab:capabilities_complementarities}
    \footnotesize
    \renewcommand{\arraystretch}{1.1}
    \setlength{\tabcolsep}{4pt}
    \begin{tabular}{p{1.8cm} p{2.8cm} p{3.0cm} p{3.2cm}}
        \toprule
        \textbf{Technique} & \textbf{Primary Contribution} &
        \textbf{Limitations in Isolation} & \textbf{Complementary Role} \\
        \midrule
        \textbf{\gls{opt}} &
        Objective-driven decision making under explicit constraints &
        Depends on problem formulation, available models, and computational tractability &
        Selects actions that optimize objectives while accounting for constraints and learned predictions \\

        \textbf{\gls{rl}} &
        Adaptive sequential decision making through interaction and feedback &
        Can be sample-inefficient and does not inherently guarantee safety or constraint satisfaction &
        Learns and refines policies using interaction, simulated experience, and predictive models \\

        \textbf{\gls{wm}s} &
        Predictive modeling, simulation, and look-ahead evaluation &
        Prediction errors and model bias can degrade planning and decision quality &
        Supports planning and policy learning through predictions of future outcomes \\

        \textbf{\gls{fm}s} &
        Transferable representations, contextual understanding, and reasoning capabilities &
        Do not inherently provide reliable dynamics models, closed-loop adaptation, or constraint satisfaction &
        Interpret heterogeneous information and provide high-level guidance or structured inputs to decision-making components \\
        \bottomrule
    \end{tabular}
\end{table*}

%% file: sections/architecture/main.tex
\input{sections/architecture/why_difficult2}

\input{sections/architecture/general_architecture}

%% file: sections/architecture/why_difficult2.tex
\subsection{Structural Characteristics of AI-Native RAN Control}
\label{subsec:whyDifficult}
\begin{figure*} \centering \includegraphics[width=0.95\linewidth]{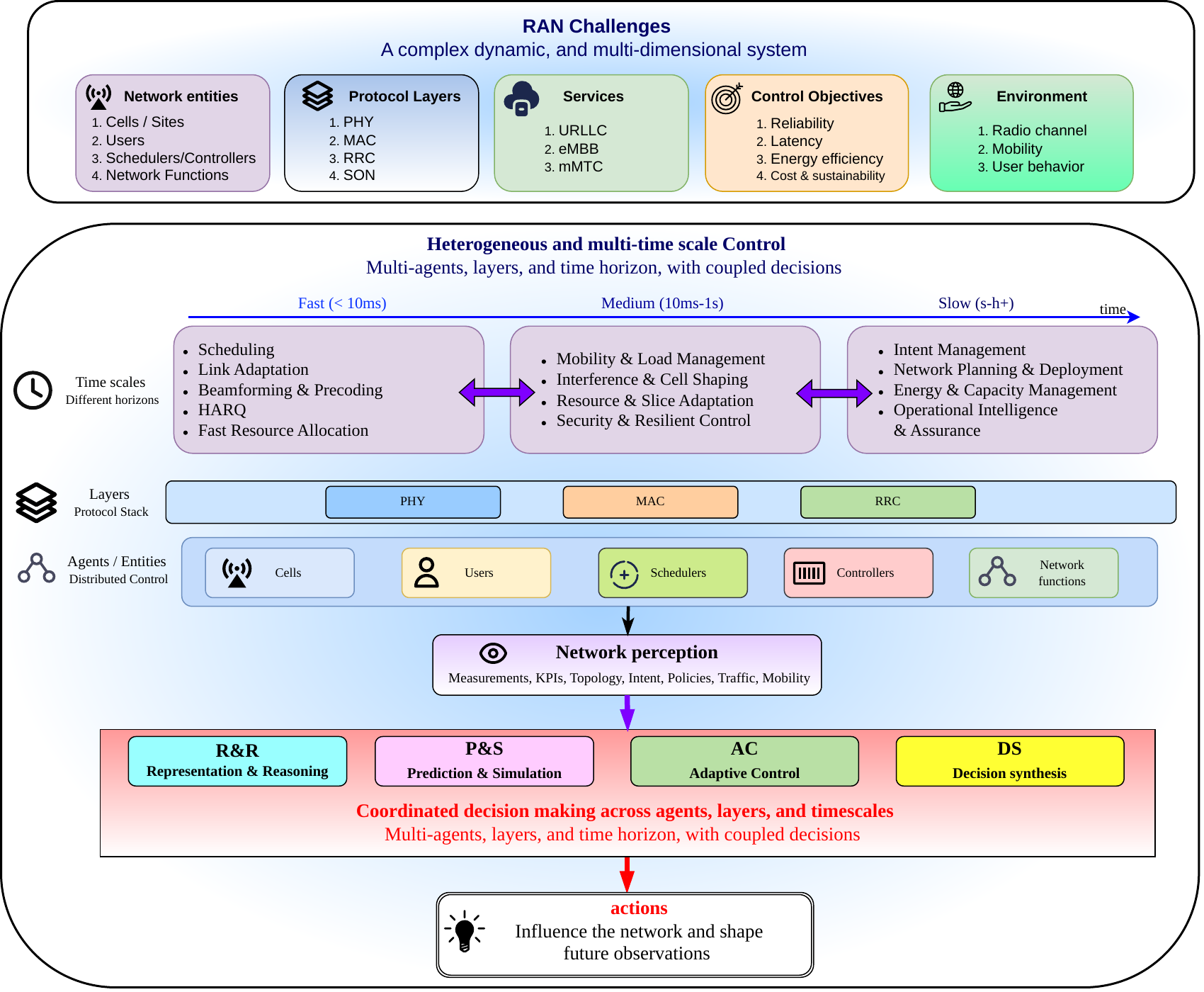} \caption{ AI-native RAN control as a distributed decision-making system. Network entities operating across protocol layers and timescales observe a dynamic environment and interact through coupled control loops. These interactions give rise to the need for four complementary capabilities: Representation \& Reasoning (R\&R), Prediction \& Simulation (P\&S), Adaptive Control (AC), and Decision Synthesis (DS), which together support coordinated decision making across heterogeneous network conditions, control objectives, and timescales. } \label{fig:challenges} \end{figure*}

Figure~\ref{fig:challenges} illustrates the perspective adopted in this survey. AI-native RAN control is viewed as a distributed decision-making system operating across multiple agents, protocol layers, and timescales. The decision-making process is enabled by an underlying perception layer that continuously gathers and interprets information from the network and its environment. Building on these observations, the controller must understand the current state, anticipate future dynamics, adapt its behavior, and ultimately synthesize executable decisions.

Three structural characteristics shape this system: distributed and multi-agent decision making, cross-layer dependencies, and multi-timescale operation. Together, they create interdependent decisions under partial information, heterogeneous operational constraints, and different temporal requirements. Consequently, effective RAN control requires coordinating decisions across entities, functions, and timescales rather than optimizing each control function independently.

\subsubsection{Distributed and Cross-Layer Dependencies} 

\gls{ran} control decisions are distributed across base stations, antennas, user equipment, and centralized or distributed controllers, each operating with different observations, control authority, and objectives. For example, scheduling, handover management, and interference coordination decisions often involve multiple network entities and therefore require coordination to avoid conflicts and ensure efficient resource utilization \cite{OnlineFS, alizadeh2025offline, wu2006tilting, forsberg2023network}. Consequently, no single entity necessarily possesses sufficient information or authority to optimize network-wide performance independently. 

Decisions are also coupled across protocol layers and network functions. For example, physical-layer resource allocation directly affects achievable data rates, latency, and interference conditions, while higher-layer functions such as cell configuration, traffic steering, and network slicing determine the operating conditions under which lower-layer controllers operate \cite{xu2025fully, lacava2023programmable, zhou2025world}. As a result, locally optimal actions may degrade overall network performance or conflict with decisions made elsewhere in the system. 

Effective coordination must therefore address partial observability, interdependent actions, and potentially conflicting objectives operating at different temporal horizons. These challenges motivate complementary capabilities for interpreting heterogeneous observations, anticipating the consequences of candidate actions, adapting control policies to changing conditions, and synthesizing feasible decisions that support coherent network-wide behavior.

\subsubsection{RAN Control Loops Across Timescales}

In addition to being distributed and cross-layer, \gls{ran} control is inherently multi-timescale. Radio propagation, mobility, traffic demand, and network evolution operate at different rates, making a single control loop insufficient. Modern \gls{ran} architectures therefore organize decision making through multiple interacting control loops, each characterized by its own latency requirements, information availability, and objectives \cite{o_ran_whitepaper_2020,Polese_2023_Understanding_ORAN,fourati2021comprehensive}. 

Following the typical O-RAN control-loop timing bands, we distinguish fast control (real-time, RT), medium control (near-real-time, Near-RT), and slow control (non-real-time, Non-RT) \cite{o_ran_whitepaper_2020, oran_architecture_2026}.
The timing boundaries given below are approximate, and individual applications may operate within narrower ranges.

\paragraph{Fast Timescale (RT, below 10~ms)}

Fast control loops operate on rapidly varying physical-layer observations. Representative functions include link adaptation, HARQ, beamforming, power control, and per-slot scheduling \cite{3gpp38213,3gpp38214,3gpp38321,mismar2019deep}. These decisions are subject to stringent latency and computational constraints and often require tightly coordinated radio-resource management.

\paragraph{Medium Timescale (Near-RT, 10~ms to less than 1~s)}

Medium-timescale control adapts network behavior to evolving traffic, mobility, interference, and service conditions. Representative functions include mobility management, traffic steering, load balancing, admission control, and interference coordination \cite{wang2018handover,lacava2023programmable,addali2019dynamic}. Decisions at this level rely on aggregated measurements and can exploit prediction and adaptive optimization.

\paragraph{Slow Timescale (Non-RT, 1~s and above)}

Slow control loops optimize longer-term network objectives using historical observations and network-wide information. Representative functions include network slicing, coverage and capacity optimization, energy management, and cell configuration \cite{rodriguez20205g,hoffmann2024energy,aroua2025adaptive,Polese_2023_Understanding_ORAN}. Here, network slicing refers to slice planning and dimensioning; the adaptation of resources within existing slices operates at the medium timescale. %
Within this band, applications such as network planning and energy management may operate on timescales of minutes to hours. These decisions define the operating conditions under which faster controllers subsequently operate. 

Although operating at different temporal horizons, these control loops remain coupled. Long-term decisions constrain and guide medium- and fast-timescale control, while measurements collected at lower layers are reported upward to monitor and retrain higher-level policies~\cite{Polese_2023_Understanding_ORAN}. This hierarchical organization is reflected in O-RAN, whose Non-RT RIC, Near-RT RIC, and lower-layer RAN functions support control and optimization at different temporal scales, subject to their respective functional and latency requirements \cite{o_ran_whitepaper_2020,Polese_2023_Understanding_ORAN}.

The coexistence of multiple agents, protocol layers, and timescales creates heterogeneous requirements in terms of information processing, anticipation, adaptation, computational complexity, and constraint handling. No single AI or model-based technique addresses all of these requirements in isolation. Understanding the resulting capability requirements is therefore essential for the design of AI-native \gls{ran} control systems.

\subsubsection{Formal Problem Structure} The structural characteristics discussed above can be represented through a generic sequential decision process. Consider a \gls{ran} composed of $N$ interacting decision entities $\mathcal{G}=\{g_1,\ldots,g_N\}$, coordinated through centralized, hierarchical, or distributed control architectures \cite{o_ran_whitepaper_2020,Polese_2023_Understanding_ORAN}.

Let $s_t\in\mathcal S$ denote the global network state at time $t$, where $\mathcal S$ is the state space. Since information is inherently distributed, each decision entity observes only a partial view of the network,

\[ o_t^i=\mathcal O_i(s_t), \]
where $\mathcal O_i:\mathcal S\rightarrow\Omega_i$ is the observation function associated with entity $g_i$, $\Omega_i$ denotes its observation space, and $o_t^i\in\Omega_i$ the corresponding observation.

Each entity selects an action $a_t^i\in\mathcal A_i$, where $\mathcal A_i$ denotes its action space. The resulting joint action is

\[ a_t=(a_t^1,\ldots,a_t^N). \]

The collective observation and action spaces are given by

\[ \Omega=\Omega_1\times\cdots\times\Omega_N, \qquad \mathcal A=\mathcal A_1\times\cdots\times\mathcal A_N. \]

The network evolves according to 

\[ s_{t+1}=f(s_t,a_t,\xi_t), \]

where $f(\cdot)$ denotes the generally unknown network dynamics, capturing interactions among decision entities, protocol layers, and control loops operating at different timescales. The random variable $\xi_t$ represents uncertainty arising from traffic fluctuations, user mobility, channel variability, and other exogenous factors affecting network behavior. 

The joint action naturally decomposes into decisions operating at fast, medium, and slow timescales,

\[ \begin{aligned} a_t &= \Big\{ a_t^{(\tau_s)}, a_t^{(\tau_m)}(a_t^{(\tau_s)}), a_t^{(\tau_f)}(a_t^{(\tau_m)},a_t^{(\tau_s)}) \Big\},\\ \tau_f &\ll \tau_m \ll \tau_s, \end{aligned} \]

where $\tau_f$, $\tau_m$, and $\tau_s$ denote representative fast, medium, and slow control intervals, respectively. Slower-timescale decisions may define objectives, constraints, resource allocations, or policy parameters that influence the action space and operating conditions of faster control loops.

The objective is to determine a policy $\pi:\mathcal H\rightarrow\mathcal A$ that maps observation histories $h_t=(o_{0:t},a_{0:t-1})\in\mathcal H$ to actions, since partial observability generally requires memory of past observations, so as to maximize a long-term network utility while satisfying operational constraints:

\begin{equation} \begin{aligned} \max_{\pi}\quad & \mathbb{E}\!\left[ \sum_{t=0}^{T}\gamma^t R(s_t,a_t) \right] \\ \text{s.t.}\quad & a_t\in\mathcal F(s_t), \\ & \mathbb{E}\!\left[ \textstyle\sum_{t=0}^{T}\gamma^t C_k(s_t,a_t) \right]\le \bar C_k, \quad k=1,\ldots,K, \\ 
& o_t^i=\mathcal O_i(s_t), \qquad \forall i. \end{aligned} \label{eq:control_objective} \end{equation}

Here, $R(s_t,a_t)$ denotes an aggregate network utility that may combine multiple, potentially conflicting objectives, including throughput, latency, reliability, fairness, and energy efficiency. The parameter $\gamma\in(0,1]$ is the discount factor, $T$ is the decision horizon, and $\mathcal F(s_t)\subseteq\mathcal A$ denotes the set of feasible actions under the current network state and operational constraints. While $\mathcal F(s_t)$ captures instantaneous constraints, the functions $C_k$ with budgets $\bar C_k$ capture long-term constraints, such as average energy consumption or service-level agreement violation rates, that can only be satisfied over the decision horizon.

This formulation captures key characteristics of AI-native \gls{ran} control, including distributed observations, uncertainty, coupled network dynamics, multiple control timescales, and operational constraints. It also exposes the fundamental requirements that any controller must satisfy. To make effective decisions, controllers must interpret heterogeneous observations, anticipate future network evolution, adapt their behavior as operating conditions change, and generate feasible actions that satisfy objectives and constraints. The next subsection derives these requirements more systematically and establishes a functional decomposition of AI-native \gls{ran} control.

%% file: sections/architecture/general_architecture.tex
\subsection{From RAN Requirements to Functional Capabilities}
\label{subsec:generalArchitecture}
The structural characteristics of AI-native \gls{ran} control discussed in the previous subsection imply a set of requirements that no single AI or model-based technique can satisfy in isolation. Instead, effective operation requires complementary capabilities for understanding network conditions, anticipating their evolution, adapting decisions to changing environments, and ensuring that resulting actions are operationally feasible.

Motivated by these requirements, we adopt a functional view of hybrid AI for \gls{ran} control. Rather than organizing the discussion around specific techniques such as \gls{fm}s, \gls{wm}s, \gls{rl}, or \gls{opt}, we focus on the functions that a decision-making system must perform. These functions should be interpreted as conceptual roles rather than prescribed software modules. Depending on the implementation, a single model may realize multiple functions, while the same function may be distributed across several components. Similar functional roles can be identified across a wide range of modern AI systems and decision-making architectures \cite{ha2018worldmodels,hafner2023mastering,schrittwieser2020muzero,reed2022gato,driess2023palme,wang2025founder}. 

The following design principles motivate this decomposition and clarify why the resulting functions remain useful even when their implementations are tightly integrated.

\noindent\textbf{Design Principle 1: Understanding and decision making are distinct functions.} Many modern systems learn representations jointly with downstream decision policies, while others explicitly expose intermediate representations that support multiple tasks. Regardless of the implementation, transforming heterogeneous observations into meaningful and actionable information serves a different purpose from selecting control actions. We therefore distinguish understanding the environment from acting upon it \cite{bommasani2021opportunities,yang2023foundationdecision,reed2022gato,driess2023palme}.

\noindent\textbf{Design Principle 2: Anticipating the future is different from deciding how to act.} Effective control often requires reasoning not only about the current network state but also about its future evolution. Although prediction and decision making are frequently coupled, especially in model-based learning systems, predictive models can simultaneously support planning, optimization, and policy learning. The ability to anticipate future dynamics therefore constitutes a distinct requirement of AI-native \gls{ran} control \cite{ha2018worldmodels,schrittwieser2020muzero,hafner2023mastering,Moerland_2023_Model_based_RL}.

\noindent\textbf{Design Principle 3: Adaptation remains essential in dynamic environments.} Network conditions, traffic demands, user behavior, and radio environments evolve continuously over time. Consequently, decision systems must be capable of updating their behavior in response to observations and feedback. This adaptive capability remains important regardless of whether it is achieved through reinforcement learning, online optimization, imitation learning, or other approaches \cite{bertsekas2019reinforcement,reed2022gato,driess2023palme}.

\noindent\textbf{Design Principle 4: Operational feasibility requires explicit consideration.} Generating a desirable action is not sufficient in operational networks. Decisions must satisfy protocol constraints, resource limitations, service requirements, and operational policies, while remaining consistent with higher-level safety and governance objectives. In distributed and multi-agent environments, feasibility is further complicated by the need to coordinate independently generated decisions across entities, protocol layers, and timescales. Consequently, mechanisms are required to reconcile, synchronize, and validate candidate actions before execution, using optimization, constraint handling, policy enforcement, runtime assurance, or related techniques \cite{amos2017optnet,kotary2021endtoend,elmachtoub2022smart,belter2026assured}.

These principles reveal four recurring requirements of AI-native \gls{ran} control: understanding the current network state, anticipating future behavior, adapting decisions based on feedback, and producing feasible actions. Each requirement corresponds to one element of the formulation in~\eqref{eq:control_objective}: partial observability ($o_t^i=\mathcal O_i(s_t)$) requires constructing a usable representation of the network state; the unknown and stochastic dynamics $f(\cdot)$ and $\xi_t$ require anticipating the consequences of actions; learning the policy $\pi$ from interaction requires adaptation based on feedback; and the feasibility and long-term constraints, together with the coupling of decisions across entities and timescales, require that candidate actions be reconciled before execution. These four elements are distinct parts of the control problem, which is why the decomposition comprises four capabilities. %
In the following subsection, these requirements are formalized as four complementary functional capabilities that provide a common framework for analyzing hybrid AI systems for \gls{ran} control.

It is important to distinguish these capabilities from perception. Perception does not constitute a decision-making capability in itself. Rather, it transforms observations collected from the network and its environment into situational awareness that can support subsequent decision making. In AI-native RANs, perception may integrate network telemetry, radio measurements, topology information, service requirements, operator intents, and other contextual information to provide a coherent understanding of the network state, as illustrated in Fig.~\ref{fig:challenges}. Recent work on perception-aided AI-RANs similarly argues that perception transforms heterogeneous sensing data into semantic and contextual representations that can be exploited by higher-level reasoning, prediction, adaptation, and control functions \cite{marasinghe2026perception}.

%% file: sections/functional_block_new/main.tex
The previous section showed that AI-native \gls{ran} control requires four complementary capabilities: understanding network conditions, anticipating their evolution, adapting decisions to changing environments, and ensuring that resulting actions remain operationally feasible. In this section, these requirements are formalized as four functional building blocks: Representation \& Reasoning (\gls{randr}), Prediction \& Simulation (\gls{pands}), Adaptive Control (\gls{ac}), and Decision Synthesis (\gls{ds}). 

These building blocks are functional roles within the decision-making
process rather than implementation modules: a single AI technique may
support several of them, and the same role may be realized by different
combinations of foundation models, world models, reinforcement learning,
optimization, and other model-based approaches. Accordingly, for each
building block we characterize its role in AI-native RAN control, the
information it consumes, the outputs it produces, and how it interacts with
the other blocks, before reviewing representative realizations. This yields
a common framework for analyzing heterogeneous AI systems and understanding
how their complementary capabilities can be combined.

\input{sections/functional_block_new/RR}

\input{sections/functional_block_new/PS}

\input{sections/functional_block_new/AC}

\input{sections/functional_block_new/DS}

\subsection{Interplay of the Building Blocks}
\label{subsec:blockInterplay}
Fig.~\ref{fig:functional_loop} summarizes how the four building blocks interact. Rather than executing in a fixed order, the blocks exchange information through a shared decision context. \gls{randr} contributes the task-relevant representation $z_t$ derived from observations $o_t$ and context $c_t$, and \gls{pands} contributes estimates of future outcomes $\hat{y}_{t:t+H}$. \gls{ac} draws on this context, together with feedback $\rho_t$, to update the decision mechanism that produces candidate decisions $\mathcal{U}^{\mathrm{AC}}_t$, and \gls{ds} turns these candidates into an executable action $a_t\in\mathcal{F}(s_t)$, optionally evaluating them with \gls{pands} before enforcement. The only direct hand-over is therefore from \gls{ac} to \gls{ds}: \gls{ac} improves how decisions are generated, whereas \gls{ds} determines which decisions are enacted. The blocks also need not operate at the same rate; for example, \gls{ds} acts at every decision, whereas \gls{ac} may update the decision mechanism far less frequently. In a given system, any block may be absent or merged with another, as the applications reviewed in Section~\ref{sec:applications} illustrate.

Such a loop exists at each control timescale. As discussed in Section~\ref{subsec:whyDifficult}, loops at different timescales are coupled: slower loops shape the context $c_t$ of faster ones through policies, objectives, and constraints, while faster loops return feedback that refines higher-level decisions. The four building blocks therefore describe the decision process at every level of the RAN control hierarchy, and the coupling between levels takes place through their context and feedback.

\input{sections/functional_block_new/fig_pipeline}

%% file: sections/functional_block_new/RR.tex
\subsection{Representation \& Reasoning}
\label{subsec:representationReasoning}
\subsubsection{Functional Role} 

In the formulation of Section~\ref{subsec:whyDifficult}, each decision entity observes only a partial view of the network through $o_t^i=\mathcal{O}_i(s_t)$. These observations originate from different network entities, protocol layers, and management systems, and may differ substantially in format, semantics, temporal resolution, and reliability, ranging from per-beam user counts obtained from beam sweeps~\cite{aroua2025adaptive} to management-level KPIs~\cite{alizadeh2025offline}. 

The first role of an AI-native controller is therefore to transform heterogeneous observations into a coherent description of the current network state: 

\[ 
z_t=\Phi(o_t,c_t), 
\] 
where $o_t$ denotes the available observations, $c_t$ captures contextual information such as topology, configuration, history, or operator objectives, and $z_t$ represents a task-relevant description of the decision context.

Beyond representation, the \gls{randr} capability provides reasoning mechanisms that relate information originating from different entities, layers, and timescales. This enables the controller to identify relevant dependencies and determine which information is most useful for the current task~\cite{battaglia2018relational}, as well as to interpret objectives such as operator intents~\cite{Habib2025LlmBasedIntentProcessing}. Together, representation and reasoning provide the informational foundation upon which prediction, adaptation, and decision synthesis operate. 

\subsubsection{Representative Realizations} 
The \gls{randr} capability can be realized through techniques that support either representation, reasoning, or both. 

\textit{Representation-focused approaches.} 
Early methods relied on engineered state abstractions derived from domain knowledge~\cite{kaelbling1998planning}, or built task-relevant state abstractions adaptively during learning~\cite{samejima1999adaptive}. More recent approaches learn compact representations directly from data using recurrent architectures, latent-state models, and other representation-learning techniques \cite{wierstra2010recurrent,baisero2018learning,igl2018deep}. Hybrid methods further combine learned representations with structured priors, for example through graph-based and neuro-symbolic approaches \cite{madbhavi2023graph,jin2112graph,zhang2024neuro}. 

\textit{Reasoning-enabled approaches.} Beyond constructing representations, a decision system may need to relate information across entities, layers, timescales, objectives, and constraints. This can be achieved through symbolic reasoning, knowledge-graph reasoning, neuro-symbolic methods, or more recently through foundation models capable of performing contextual and semantic reasoning over heterogeneous inputs \cite{yang2023foundationdecision,bommasani2021opportunities}. In AI-native \gls{ran}s, such models can jointly interpret measurements, \gls{kpi}, topology information, operator intents, and domain knowledge, enabling reasoning at different levels of abstraction. Telecom-oriented adaptations such as TelecomGPT and multimodal wireless foundation models further illustrate this trend \cite{zou2025telecomgpt,alikhani2024large,jiao2025ai2mmum}. 

The \gls{randr} process can also be enriched through mechanisms such as \gls{rag} and prompting, which incorporate external knowledge, objectives, and constraints into the representation and reasoning process \cite{gao2023retrieval,brown2020language,sahoo2024systematic}.

%% file: sections/functional_block_new/PS.tex
\subsection{Prediction \& Simulation}
\label{subsec:predictionSimulation}

\subsubsection{Functional Role}

The formulation of Section~\ref{subsec:whyDifficult} assumes that the network evolves according to

\[ s_{t+1}=f(s_t,a_t,\xi_t), \]

where the dynamics $f(\cdot)$ are generally unknown and capture interactions across network entities, protocol layers, and control timescales. Consequently, an understanding of the current network state alone is insufficient for effective decision making. Controllers must also anticipate how the network may evolve and how alternative actions may affect future outcomes.

The Prediction \& Simulation (P\&S) capability fulfills this role by extending the decision context beyond the current state. Building on the representation produced by R\&R, it estimates

\[ \hat{y}_{t:t+H} = \Psi ( z_t, a_{t:t+H-1}, c_t, \xi_t ), \]

where $\Psi(\cdot)$ denotes a predictive or simulation model and $H$ is the prediction horizon, and $\xi_t$ captures exogenous and stochastic factors that the controller does not determine, such as traffic arrivals or channel fading.

Depending on the task, the output may correspond to future network states and KPIs, such as resource utilization and throughput~\cite{bouton2026mobile}, or to traffic demand, mobility patterns, radio conditions, interference levels, rewards, or other variables relevant to control objectives. While prediction estimates likely future outcomes from available information, simulation evaluates network evolution under hypothetical actions or operating conditions. Together, these capabilities support forecasting, planning, and candidate-action evaluation without requiring direct experimentation on the operational network \cite{ha2018worldmodels,schrittwieser2020muzero,hafner2023mastering,Moerland_2023_Model_based_RL}.

By making future network behavior accessible to the controller, \gls{pands} enables proactive rather than purely reactive decision making.

\subsubsection{Representative Realizations} 
The \gls{pands} capability can be realized through simulation-based, predictive, or hybrid approaches. 

\textit{Simulation-based approaches} explicitly reproduce network behavior under specified operating conditions. Traditional network simulators model traffic, radio propagation, protocol behavior, and resource allocation, enabling candidate decisions to be evaluated before deployment \cite{nardini2024enabling,adday2024investigating}. \glspl{dt} extend this idea by maintaining virtual representations continuously synchronized with operational measurements and network data \cite{tran2025network,zaki2026reference}.

\textit{Predictive approaches} focus on forecasting future variables or outcomes from historical and real-time observations. Early methods relied on statistical models and on analytical performance models, such as queueing analyses that predict the achievable quality of service, whereas modern approaches increasingly employ neural architectures, transformers, graph-based models, and latent-state representations capable of capturing complex temporal and spatial dependencies \cite{young2000stochastic,rashid2009opportunistic,cui2019traffic,habib2024transformer}.

A particularly relevant class of predictive models is represented by \gls{wm}s. Rather than forecasting a single variable, they learn a model of the system dynamics,

\[ \hat{z}_{t+1}=\Psi(z_t,a_t,\xi_t), \]
i.e., a one-step special case of $\Psi(\cdot)$ that predicts the next latent state, which can be repeatedly queried under alternative actions to generate imagined trajectories \cite{ha2018worldmodels,hafner2023mastering}. This enables planning, counterfactual evaluation, and multi-step reasoning without direct interaction with the physical network, although model errors can accumulate over long horizons~\cite{Moerland_2023_Model_based_RL,bouton2026mobile}.

\textit{Hybrid approaches} combine explicit domain knowledge with data-driven predictive models to improve fidelity and to balance interpretability and adaptability~\cite{tran2025network,zaki2026reference}. Such combinations are particularly attractive in RAN environments, where partial system knowledge is often available but complete analytical models remain impractical~\cite{zaki2026reference}.

%% file: sections/functional_block_new/AC.tex
\subsection{Adaptive Control}
\label{subsec:adaptiveControl}

\subsubsection{Functional Role}

Understanding the current network state and anticipating its evolution are necessary but not sufficient for effective control. Network conditions, traffic demand, user behavior, and operational objectives evolve over time, causing previously effective strategies to become suboptimal \cite{coronado2022zero,luong2019applications}. The role of \gls{ac} is therefore to continuously adjust how decisions are generated in response to feedback and changing conditions. Building on the representation produced by \gls{randr} and the predictive information generated by \gls{pands}, the adaptation process can be expressed as

\begin{equation} \theta_{t+1} = \mathcal{A} ( \theta_t, z_t, \hat{y}_{t:t+H}, \rho_t, c_t ), \label{eq:adaptation_operator} \end{equation}
$\theta_t$ denotes the elements of the decision-making mechanism subject to adaptation and $\rho_t$ captures feedback from previous decisions.

Depending on the implementation, $\theta_t$ may represent controller parameters, policies, objective functions, decision thresholds, or higher-level strategies. Feedback may include rewards, network KPIs, prediction errors, or constraint violations~\cite{annaswamy2023adaptive,sutton2018reinforcement,stooke2020responsive}. Combined with predictive information, this enables the controller to improve its future behavior.

Importantly, \gls{ac} does not necessarily determine the final action to be executed. Instead, it updates the mechanisms that generate candidate decisions. The responsibility for selecting, coordinating, validating, and deploying actions belongs to the \gls{ds} capability. \gls{ac} therefore focuses on improving the decision-making process itself.

\subsubsection{Representative Realizations} The \gls{ac} capability can be realized through explicit adaptation rules, optimization-based adaptation, learning-based adaptation, or hybrid combinations.

\textit{Explicit adaptation.} In explicit adaptation, update rules are defined through engineering knowledge, control laws, or expert-designed policies, 

\[ \theta_{t+1}=g(\theta_t,z_t,\rho_t,c_t). \]

Examples include classical adaptive control and self-tuning mechanisms~\cite{aastrom1995adaptive,annaswamy2023adaptive}, as well as rule-based network optimization such as load-triggered tuning of handover parameters~\cite{addali2019dynamic}. These approaches offer interpretability and predictable behavior, but rely on the availability of suitable adaptation rules. 

\textit{Optimization-based adaptation.} Optimization-based approaches update controller parameters by solving an optimization problem,

\[ \theta_{t+1} = \arg\max_{\theta} J(\theta;z_t,\hat{y}_{t:t+H},\rho_t,c_t). \]

Examples include online optimization~\cite{chen2017online} and model predictive control, which re-optimizes the control action at every sampling instant~\cite{mayne2000constrained} and, in its coalitional variant, periodically re-selects which local controllers cooperate~\cite{masero2020coalitional}. Optimization belongs to AC when it updates the decision mechanism itself; when it directly computes deployable actions, it is more naturally associated with DS. 

\textit{Learning-based adaptation.} Learning-based approaches infer how control mechanisms should evolve from data, experience, or interaction with the environment,

\[ \theta_{t+1} = \mathcal{A}_{\phi} ( \theta_t, z_t, \hat{y}_{t:t+H}, \rho_t, c_t ). \]

Representative examples include reinforcement learning, deep reinforcement learning, and meta-learning \cite{luong2019applications,sutton2018reinforcement,zhao2024meta}. These approaches are particularly attractive when effective adaptation rules are difficult to specify analytically.

\textit{Hybrid approaches.} Hybrid approaches combine multiple adaptation mechanisms. Learning may estimate unknown system characteristics within an optimization framework~\cite{dahrouj2021overview}, optimization may constrain learned policies~\cite{stooke2020responsive}, and domain knowledge may guide adaptation~\cite{aroua2025adaptive}. This allows a controller to keep adapting online without its updates drifting outside the safe operating region of the network.

%% file: sections/functional_block_new/DS.tex
\subsection{Decision Synthesis}
\label{subsec:decisionSynthesis}
\subsubsection{Functional Role}

Representation, prediction, and adaptation can generate candidate decisions, but these decisions cannot always be deployed directly. In AI-native \gls{ran}s, multiple controllers, agents, applications, and control loops operate concurrently across entities, protocol layers, and timescales. As a result, independently generated decisions may compete for shared resources, pursue conflicting objectives, or violate operational constraints~\cite{peng2015recent,giannopoulos2025comix}. 

The role of \gls{ds} is to transform candidate decisions into executable network actions. While \gls{ac} determines how decisions are generated and updated, \gls{ds} determines which decisions are ultimately enacted. \gls{ds} is therefore a functional role rather than a technique: optimization is one way to realize it, alongside rule-based validation, arbitration, and conflict resolution. Moreover, whereas optimization typically generates a decision from a problem formulation, \gls{ds} operates on candidate decisions produced by other controllers and determines which of them, or which combination, is executed. %
In doing so, 
it operationalizes the feasibility constraint $a_t \in \mathcal{F}(s_t)$ of the global control formulation in~\eqref{eq:control_objective}, where $\mathcal{F}(s_t)$ denotes 
the set of actions that satisfy the applicable technical, operational, policy, regulatory, and safety requirements in the current network state \cite{singh2026regulations}.

Given a set of candidate decisions $\mathcal{U}^{\mathrm{AC}}_t$ generated by one or more controllers, the \gls{ds} process can be expressed as 

\begin{equation} a_t = \mathcal{D} \left( \mathcal{U}^{\mathrm{AC}}_t, z_t, \hat{y}_{t:t+H}, \rho_t, c_t \right), \qquad a_t \in \mathcal{F}(s_t), \label{eq:decision_synthesis} \end{equation}
where $\mathcal{D}(\cdot)$ denotes the synthesis mechanism. The process may exploit the current network representation $z_t$, predictive information $\hat{y}_{t:t+H}$, feedback $\rho_t$, and contextual information $c_t$, including policies, objectives, and governance requirements.

\gls{pands} can support \gls{ds} by estimating the consequences of candidate decisions before deployment~\cite{giannopoulos2025comix,bouton2026mobile}. Such information enables alternative actions to be compared in terms of throughput, latency, energy consumption, reliability, interference, or constraint violations.

Conceptually, \gls{ds} consists of three functions: 

\begin{equation} \mathcal{U}^{\mathrm{AC}}_t \xrightarrow{\mathcal{V}} \mathcal{U}^{\mathrm{valid}}_t \xrightarrow{\mathcal{R}} \mathcal{U}^{\mathrm{resolved}}_t \xrightarrow{\mathcal{S}} a_t \in \mathcal{F}(s_t), \label{eq:ds_operations} \end{equation}

where $\mathcal{V}$ denotes validation, $\mathcal{R}$ reconciliation, and $\mathcal{S}$ synthesis. Validation checks compliance with constraints and policies, reconciliation resolves interactions among decisions, and synthesis produces the final executable action. 

A representative hybrid realization is proposed in \cite{11771692}, where a Lagrangian \gls{ppo} controller generates candidate power-allocation decisions and a rule-based compliance layer validates them against \gls{eeirp} regulatory constraints. Actions that violate the constraints are replaced by a predefined fallback allocation before execution. This illustrates how adaptive learning and explicit governance mechanisms can be combined within the \gls{ds} function to ensure safe deployment.

\gls{ds} is therefore the point at which network actuation occurs, linking AI-driven decision making to operational network control. More generally, it serves as the governance layer that transforms candidate decisions into actions that are feasible, coordinated, and deployable.

\subsubsection{Representative Realizations} 

The \gls{ds} capability can be realized through rule-based, optimization-based, learning-assisted, or hybrid approaches.

\textit{Rule-based decision synthesis.} Rule-based approaches rely on predefined policies, priorities, admission rules, or governance procedures to determine executable actions~\cite{Adamczyk2023ConflictMitigationORAN,polese2026beyond}. They are particularly attractive when constraints and conflict-resolution mechanisms can be specified explicitly. 

\textit{Optimization-based decision synthesis.} Optimization-based approaches formulate synthesis as a constrained optimization problem~\cite{boyd2004convex},

\[ a_t= \arg\max_{a\in\mathcal{F}(s_t)} J(a). \] 

Examples include resource allocation, scheduling, network slicing, and power-control optimization~\cite{peng2015recent}. Such approaches naturally reconcile competing objectives while enforcing feasibility constraints.

\textit{Learning-assisted decision synthesis.} Learning-assisted approaches use predictive or learned models to evaluate candidate actions, estimate their consequences, or guide action selection~\cite{dalal2018safelayer}. In these systems, learning supports the synthesis process, while operational constraints remain enforced by the surrounding decision framework~\cite{alshiekh2018shielding}.

\textit{Hybrid approaches.} Hybrid approaches combine optimization, learning, domain knowledge, and governance mechanisms. Learned models may propose candidate actions or surrogate objectives, while optimization and policy-enforcement mechanisms ensure feasibility and compliance~\cite{kotary2021endtoend,Adamczyk2023ConflictMitigationORAN}. In this way, learning widens the set of candidate actions considered, while the final choice remains subject to explicit, verifiable constraints.

%% file: sections/functional_block_new/fig_pipeline.tex
\begin{figure*}[t]
\centering
\resizebox{0.8\textwidth}{!}{%
\begin{tikzpicture}[
  font=\small,
  blk/.style={draw, thick, rounded corners=4pt, align=center,
              text width=2.7cm, minimum height=1.45cm, inner sep=3pt},
  hub/.style={draw, thick, rounded corners=4pt, align=center,
              text width=4.2cm, minimum height=1.45cm, inner sep=3pt, fill=gray!8},
  strip/.style={draw, thick, rounded corners=4pt, align=center,
                minimum height=0.7cm, inner sep=3pt},
  ts/.style={draw, thick, rounded corners=3pt, align=center,
             text width=2.7cm, minimum height=1.0cm, inner sep=2pt, fill=gray!6},
  fwd/.style={line width=1.0pt, -{Latex[length=2.6mm]}},
  aux/.style={line width=0.8pt, -{Latex[length=2.4mm]}},
  opt/.style={line width=0.9pt, dash pattern=on 4pt off 2.5pt, -{Latex[length=2.4mm]}},
  lbl/.style={font=\scriptsize, align=center, fill=white, inner sep=1.5pt}
]

\draw[thick, rounded corners=6pt, gray!70] (0,-0.25) rectangle (13.4,6.55);
\node[font=\scriptsize\itshape, anchor=north west, gray!40!black] at (0.1,6.5)
  {one control loop (at any timescale)};

\node[strip, fill=gray!4, minimum width=12.6cm] (cx) at (6.7,5.6)
  {\textbf{Context} $c_t$: topology, configuration, history, operator intents, policies and constraints};

\node[hub]                  (hub) at (6.7,3.55) {\textbf{Shared decision}\\\textbf{context}\\[2pt]$z_t,\ \hat{y}_{t:t+H}$};
\node[blk, fill=blue!8]     (rr)  at (2.0,3.55) {\textbf{Representation}\\\textbf{\& Reasoning}\\[1pt]{\scriptsize\itshape understand the network}};
\node[blk, fill=yellow!14]  (ps)  at (11.4,3.55){\textbf{Prediction}\\\textbf{\& Simulation}\\[1pt]{\scriptsize\itshape anticipate its evolution}};
\node[blk, fill=green!12]   (ac)  at (4.4,1.05) {\textbf{Adaptive Control}\\[1pt]{\scriptsize\itshape improve how decisions are made}};
\node[blk, fill=red!8]      (ds)  at (9.0,1.05) {\textbf{Decision Synthesis}\\[1pt]{\scriptsize\itshape decide what is enacted}};

\draw[fwd] (rr) -- node[lbl, above]{$z_t$} (hub);
\draw[fwd] ([yshift=3mm]hub.east) -- node[lbl, above]{$z_t$} ([yshift=3mm]ps.west);
\draw[fwd] ([yshift=-3mm]ps.west) -- node[lbl, below]{$\hat{y}_{t:t+H}$} ([yshift=-3mm]hub.west -| hub.east);
\draw[aux] (cx.south -| rr.north) -- (rr.north);
\draw[aux] (cx.south -| hub.north) -- node[lbl, right]{intents, policies} (hub.north);

\draw[fwd] ([xshift=-1.5cm]hub.south) -- (ac.north);
\draw[fwd] ([xshift=1.5cm]hub.south) -- (ds.north);

\draw[fwd] (ac) -- node[lbl, above]{candidates}node[lbl, below]{$\mathcal{U}^{\mathrm{AC}}_t$} (ds);

\draw[opt] (ps.south) |- node[lbl, pos=0.3, right]{what-if\\evaluation} (ds.east);

\node[strip, fill=gray!14, minimum width=13.4cm] (net) at (6.7,-1.35)
  {\textbf{RAN environment}: network state $s_t$, evolving as $s_{t+1}=f(s_t,a_t\rev{,\xi_t})$}; %
\draw[fwd] (net.north -| 0.25,0) -- node[lbl, pos=0.1, right]{observations $o_t$} (0.25,3.55) -- (rr.west);
\draw[fwd] (net.north -| ac.south) -- node[lbl, pos=0.25, right]{feedback $\rho_t$} (ac.south);
\draw[fwd] (ds.south) -- node[lbl, pos=0.75, right]{action $a_t\in\mathcal{F}(s_t)$} (net.north -| ds.south);

\node[font=\scriptsize\itshape, gray!40!black] at (16.0,6.35) {loops nested across timescales};
\node[ts] (slow) at (16.0,5.15) {\textbf{Slow}\\{\scriptsize s--h+}};
\node[ts] (med)  at (16.0,3.05) {\textbf{Medium}\\{\scriptsize 10\,ms--1\,s}};
\node[ts] (fast) at (16.0,0.95) {\textbf{Fast}\\{\scriptsize $<$\,10\,ms}};
\foreach \u/\l in {slow/med, med/fast}{
  \draw[aux] ([xshift=-6mm]\u.south) -- ([xshift=-6mm]\l.north);
  \draw[aux] ([xshift=6mm]\l.north)  -- ([xshift=6mm]\u.south);
}
\node[lbl, anchor=east] at ([xshift=-7mm]$(slow.south)!0.5!(med.north)$) {policies,\\constraints\\$\rightarrow c_t$};
\node[lbl, anchor=west] at ([xshift=7mm]$(slow.south)!0.5!(med.north)$) {feed-\\back};
\node[lbl, anchor=east] at ([xshift=-7mm]$(med.south)!0.5!(fast.north)$) {policies,\\constraints\\$\rightarrow c_t$};
\node[lbl, anchor=west] at ([xshift=7mm]$(med.south)!0.5!(fast.north)$) {feed-\\back};
\draw[thick, dotted, gray!60!black] (13.45,3.05) -- (med.west);
\node[font=\scriptsize\itshape, gray!40!black, align=center] at (16.0,-0.45) {each level runs\\the loop on the left};

\end{tikzpicture}%
}
\caption{Information dependencies among the four functional building blocks within one control loop (left), and the nesting of such loops across timescales (right). The blocks exchange information through a shared decision context rather than in a fixed order, and any of them may be absent in a given system; the dashed arrow denotes optional what-if evaluation.}
\label{fig:functional_loop}
\end{figure*}
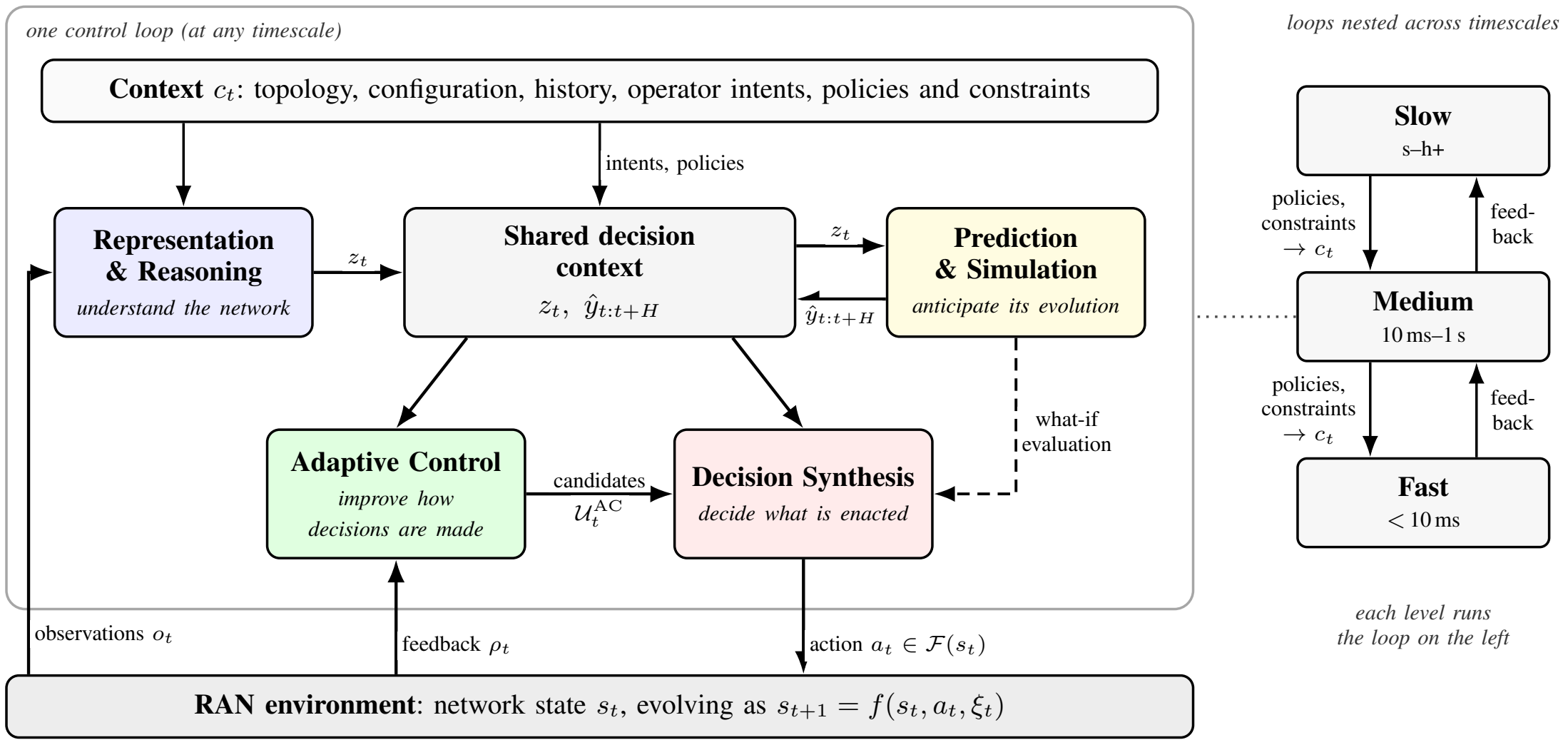

%% file: sections/multi_timescale_applications/main.tex
\section{Hybrid AI in RAN Control Applications}
\label{sec:applications}

The previous sections introduced a functional perspective on AI-native \gls{ran} control based on four complementary capabilities: \gls{randr}, \gls{pands}, \gls{ac}, and \gls{ds}. In this section, we use this perspective to analyze representative RAN control applications and examine how these capabilities are realized in practice.

Rather than organizing the literature around specific AI techniques, we focus on the capabilities required by different classes of RAN decision-making problems. The works discussed are representative rather than exhaustive: we focus on recent systems, most of them published since 2024, that address RAN control decisions and explicitly realize at least one of the four capabilities, and include earlier works only where they illustrate a capability pattern not covered by recent ones.

Although these applications address different objectives, ranging from network planning \cite{networkPlanning} and mobility management \cite{gures2020comprehensive} to scheduling, beam management, and network assurance \cite{haque2023survey}, they share a common characteristic: their decisions operate over different temporal horizons. We therefore organize the discussion into slow-, medium-, and fast-timescale control applications, followed by architectures that coordinate decision making across multiple timescales. Works are grouped according to the application class they address; the timescale denotes the typical decision horizon of that class, and individual implementations may operate faster or slower than this nominal horizon.

This analysis highlights how the requirements of a given application influence the roles of representation and reasoning, prediction and simulation, adaptive control, and decision synthesis. While many existing solutions emphasize one or two capabilities, systems that explicitly integrate all four remain comparatively uncommon. Understanding these capability patterns provides valuable insight into the current state of AI-native \gls{ran} control and the directions along which future hybrid AI systems are likely to evolve.
\input{sections/multi_timescale_applications/slow-time-scale}

\input{sections/multi_timescale_applications/medium-time-scale}
\input{sections/multi_timescale_applications/fast-time-scale}

\input{sections/multi_timescale_applications/multi-time-scale}

%% file: sections/multi_timescale_applications/slow-time-scale.tex
\subsection{Slow-Timescale Control Applications}
\label{subsec:slow_timescale}

Slow-timescale applications address strategic network optimization and long-horizon decision making, typically over minutes, hours, or longer periods. Their relaxed latency requirements permit extensive computation, network-wide reasoning, predictive evaluation, and the use of large historical datasets, digital twins, and computationally intensive AI models. Consequently, they provide a natural environment for foundation models, long-horizon prediction, planning, and optimization. 

At this timescale, the dominant functional capabilities are typically \gls{randr} and \gls{pands}, reflecting the need to interpret high-level objectives, reason over heterogeneous information, and anticipate long-term network evolution. \gls{ac} and \gls{ds} become increasingly important when multiple objectives, candidate strategies, or operational constraints must be reconciled. As shown throughout this subsection, different application classes instantiate different subsets of the four functional capabilities, with systems jointly realizing \gls{randr}, \gls{pands}, \gls{ac}, and \gls{ds} remaining comparatively rare.

The main application classes considered at this timescale are summarized in Tables~\ref{tab:intent_management_mapping}--\ref{tab:assurance_mapping}.

\begin{table}[t] 
\caption{Instantiation of the four functional capabilities in intent-driven network automation and policy management.} \label{tab:intent_management_mapping} 
\centering 
\renewcommand{\arraystretch}{1.1} 
\begin{tabular}{lcccc} 
\toprule 
\textbf{Work} & 
\textbf{R\&R} & 
\textbf{P\&S} & 
\textbf{AC} & 
\textbf{DS} \\ 
\midrule 
Mekrache \& Ksentini~\cite{mekrache2024llm} 
& \cmark & \xmark & \xmark & \xmark \\ 
Habib et al.~\cite{Habib2025LlmBasedIntentProcessing} 
& \cmark & \cmark & \cmark & \xmark \\ 
ARC~\cite{shokrnezhad2025autonomous} 
& \cmark & \xmark & \cmark & \cmark \\ 
IAN~\cite{Guo2025IAN} 
& \cmark & \xmark & \cmark & \xmark \\ 
LAMeTA~\cite{Liu2026LAMeTA} 
& \cmark & \xmark & \cmark & \xmark \\ 
Mekrache et al.~\cite{mekrache2024intent} 
& \cmark & \xmark & \xmark & \cmark \\
\bottomrule 
\end{tabular} 
\end{table}

\begin{table}[t] \caption{Instantiation of the four functional capabilities in network planning, deployment, and configuration.} \label{tab:planning_mapping} 
\centering \renewcommand{\arraystretch}{1.1} 
\begin{tabular}{lcccc} 
\toprule 
\textbf{Work} & \textbf{R\&R} & \textbf{P\&S} & \textbf{AC} & \textbf{DS} \\ 
\midrule 
LMCO~\cite{qiu2024large} 
& \cmark & \cmark & \xmark & \cmark \\ 
Quan \textit{et al.}~\cite{quan2025large} 
& \cmark & \cmark & \xmark & \cmark \\ 
Sevim \textit{et al.}~\cite{Sevim2024LLMAssistedWirelessDeployment} 
& \cmark & \xmark & \cmark & \xmark \\ 
Hribar \textit{et al.}~\cite{hribar2026radioenvironmentmappingworld} 
& \cmark & \cmark & \xmark & \xmark \\ 
Al-Tahmeesschi \textit{et al.}~\cite{al2024multi} 
& \xmark & \xmark & \cmark & \xmark \\ 
\bottomrule 
\end{tabular} 
\end{table}

\begin{table}[t] \caption{Instantiation of the four functional capabilities in energy and capacity management.} \label{tab:energy_mapping} 
\centering \renewcommand{\arraystretch}{1.1} 
\begin{tabular}{lcccc} 
\toprule 
\textbf{Work} & \textbf{R\&R} & \textbf{P\&S} & \textbf{AC} & \textbf{DS} \\ 
\midrule 
MobiWorld~\cite{chai2025MobiWorld} 
& \cmark & \cmark & \cmark & \xmark \\ 
Bouton et al.~\cite{bouton2026mobile} 
& \xmark & \cmark & \xmark & \cmark \\ 
Sleep-mode RL~\cite{el2022energy, masrur2025energy, liang2026scalable} 
& \xmark & \xmark & \cmark & \xmark \\ 
\bottomrule 
\end{tabular}
\end{table}

\begin{table}[t] 
\caption{Instantiation of the four functional capabilities in network operations, assurance, and knowledge management.}
\label{tab:assurance_mapping} 
\centering 
\renewcommand{\arraystretch}{1.1} 
\begin{tabular}{lcccc} 
\toprule 
\textbf{Work} & \textbf{R\&R} & \textbf{P\&S} & \textbf{AC} & \textbf{DS} \\ 
\midrule
Ali et al.~\cite{Ali2023Security} 
& \cmark & \xmark & \cmark & \xmark \\ 

Zhang et al.~\cite{zhang2025automatedDetection} 
& \cmark & \xmark & \xmark & \xmark \\ 

Sana et al.~\cite{Sana_2025_TeleLogs} 
& \cmark & \xmark & \xmark & \xmark \\ 

He et al.~\cite{11618692} 
& \cmark & \xmark & \xmark & \xmark \\ 

Wang et al.~\cite{wang2026fault} 
& \cmark & \xmark & \xmark & \xmark \\ 
\bottomrule 
\end{tabular} 
\end{table}

\textbf{Intent-driven network automation and policy management.} Intent management is a naturally slow-timescale application because operator intents, business objectives, and service policies evolve over timescales ranging from minutes to days. These applications therefore have sufficient time to perform semantic interpretation, reasoning, validation, and planning before network actions are deployed.
\gls{randr} is present in every work listed in Table~~\ref{tab:intent_management_mapping}, while \gls{ac}
appears in several systems that combine intent interpretation with feedback-driven control. \gls{randr} is the dominant functional block, since the primary objective is to translate natural-language intents into structured network goals, policies, or configurations.

The surveyed literature exhibits four recurrent combinations of functional capabilities. The simplest frameworks are centered on \gls{randr}, using \gls{llm}s to interpret user intents and generate structured service configurations, with \gls{ac} and \gls{ds} limited to a human-feedback loop and a validation agent, respectively~\cite{mekrache2024llm}. A second group combines \gls{randr} with \gls{ac}, coupling intent understanding with adaptive optimization through reinforcement-learning-based network control, as exemplified by LAMeTA~\cite{Liu2026LAMeTA} and IAN~\cite{Guo2025IAN}. A third pattern introduces \gls{pands} in addition to \gls{randr} and \gls{ac}, allowing intents to be validated against predicted traffic volume so that they do not cause \gls{qos} drift~\cite{Habib2025LlmBasedIntentProcessing}. Finally, ARC~\cite{shokrnezhad2025autonomous} introduces decision-level planning, decomposition, and orchestration, while the intent-management framework of~\cite{mekrache2024intent} adds cross-domain intent decomposition to a validation agent similar to that of~\cite{mekrache2024llm} and proposes intent negotiation; both thereby incorporate decision-level functions associated with \gls{ds}.

Several recent systems further enrich intent management with external knowledge and planning mechanisms. ARC~\cite{shokrnezhad2025autonomous} employs \gls{rag}~\cite{lewis2020retrieval} to augment LLM reasoning and uses hierarchical planning before delegating optimization tasks to RL agents. Similarly, IAN~\cite{Guo2025IAN} combines retrieval-enhanced intent understanding with \gls{drl}-based resource optimization. These systems illustrate the growing integration of semantic reasoning, planning, and adaptive decision making within intent-driven network automation.

Taken together, intent management remains largely dominated by \gls{randr}, with recent works progressively incorporating prediction, reinforcement-learning-based adaptive optimization, and orchestration capabilities.

\textbf{Network planning, deployment, and configuration.} Network planning and deployment are quintessential slow-timescale applications because deployment decisions typically remain effective over months or years and rely on population and user-density data, geographical information, propagation models, and coverage and capacity objectives~\cite{sapkota20245g} rather than instantaneous radio conditions. As shown in Table~\ref{tab:planning_mapping}, this application class exhibits some of the richest combinations of functional capabilities. \gls{randr} and \gls{pands} support design-space exploration and predictive evaluation, while \gls{ac} can exploit computationally intensive optimization and learning procedures since deployment strategies are updated infrequently. \gls{ds} then selects deployment configurations that satisfy coverage, capacity, cost, and operational constraints.

Historically, network planning has been dominated by model-based and optimization-driven methodologies. Classical approaches rely on propagation models and metaheuristic optimization techniques, including Genetic Algorithms, Particle Swarm Optimization, Grey Wolf Optimization, and Simulated Annealing, to determine base-station placement and network configurations under deployment constraints~\cite{sapkota20245g,yangyang2004particle}. From the functional perspective adopted in this survey, these approaches are primarily associated with \gls{pands} and \gls{ds}, relying on predictive evaluation and optimization to identify feasible deployment strategies.

More recent works incorporate foundation models into the planning process. LMCO~\cite{qiu2024large} uses an \gls{llm} to generate candidate configurations, reason over deployment objectives, and guide design-space exploration in conjunction with ray-tracing simulation, while the agentic planning framework of~\cite{quan2025large} uses \gls{llm} agents to drive a commercial planning tool through radio-map generation and base-station planning, reducing manual operations by over 90\%. These approaches instantiate the \gls{randr}+\gls{pands}+\gls{ds} pattern by combining semantic reasoning, predictive analysis, and deployment decision synthesis.

Other works emphasize adaptive optimization. The framework proposed in~\cite{Sevim2024LLMAssistedWirelessDeployment} fine-tunes an \gls{llm}-based actor with reinforcement learning to choose the location and orientation of a base station, while~\cite{al2024multi} formulates base-station placement as a multi-objective deep reinforcement-learning problem, thereby emphasizing the \gls{ac} block. By contrast, the world-model-based approach of~\cite{hribar2026radioenvironmentmappingworld} has no \gls{ac} component; it leverages learned predictive models to evaluate hypothetical measurements and support radio-environment-map reconstruction, providing one of the clearest examples of a \gls{pands}-centric realization in this application domain.

The literature shows a gradual shift toward richer combinations of functional capabilities, particularly reasoning, prediction, and adaptive optimization.

\textbf{Energy and capacity management.}
Energy and capacity management are slow-timescale applications because decisions such as cell/\gls{ru} sleeping, carrier activation, and capacity provisioning are driven by traffic and load variations occurring over minutes, hours, or longer periods rather than instantaneous radio conditions~\cite{masrur2025energy}. Consequently, these tasks rely heavily on forecasting future demand, evaluating long-term consequences, and adapting network behavior over extended horizons, making \gls{pands} and \gls{ac} the dominant functional capabilities (Table~\ref{tab:energy_mapping}).

Two dominant capability patterns emerge from the literature. The first combines predictive models with downstream control or planning. MobiWorld~\cite{chai2025MobiWorld} employs a multimodal \gls{wm} to generate imagined trajectories for tasks such as base-station sleeping and user offloading, thereby combining \gls{randr}, \gls{pands}, and \gls{ac}. Similarly,~\cite{bouton2026mobile} learns a \gls{wm} from historical network data and performs planning over imagined trajectories using a chance-constrained optimization procedure. Rather than learning a control policy, the controller repeatedly optimizes future actions using the learned dynamics model, making it a representative example of the \gls{pands}+\gls{ds} pattern.

The second pattern is dominated by adaptive control. Reinforcement-learning-based sleep-mode optimization~\cite{el2022energy}, multi-agent \gls{drl} for mmWave energy management~\cite{masrur2025energy}, and recent centralized and federated learning solutions for radio-unit sleeping in Open RAN~\cite{liang2026scalable} directly learn activation and sleeping strategies from interaction with the network. These approaches primarily instantiate the \gls{ac} block, treating energy management as a sequential decision-making problem.

This application class is primarily characterized by the interaction between \gls{pands} and \gls{ac}, with recent world-model-based approaches introducing planning and decision-synthesis capabilities.

\textbf{Operational intelligence and assurance.}
Operational intelligence and assurance comprise supervisory applications that support network reliability, correctness, and operational awareness~\cite{10230112}, as well as explainability~\cite{Ali2023Security}; machine-learning alert filtering, demonstrated on cloud monitoring systems, is one example of such an application~\cite{voutsas2024mitigating}. Unlike the previous applications, these functions do not primarily generate network actions. Instead, they analyze specifications, telemetry, logs, alarms, and network behavior to produce explanations, verification results, and diagnostic insights. Consequently, they are predominantly associated with the \gls{randr} block (Table~\ref{tab:assurance_mapping}).

The literature mainly follows two directions. The first focuses on explainability and verification of AI-enabled networks. For example,~\cite{Ali2023Security} employs an \gls{llm} to turn the training statistics of a \gls{drl}-based anti-jamming agent into human-readable reports, combining \gls{randr} with an underlying \gls{ac} mechanism. Similarly,~\cite{zhang2025automatedDetection} uses a reinforcement-learning-trained small \gls{llm} to identify contradictions in 5G network specifications, where learning is used to improve reasoning performance rather than to control the network itself.

The second direction focuses on fault diagnosis and root-cause analysis. Domain-adapted reasoning models have been proposed for 5G root-cause analysis~\cite{Sana_2025_TeleLogs}, while~\cite{11618692} combines agent-based reasoning with hyperbolic spatio-temporal graph learning for root-cause localization in cloud-edge microservice environments. More recently,~\cite{wang2026fault} employs adaptive graph reasoning to support fault diagnosis in wireless networks. Despite using different learning paradigms, these approaches serve a common purpose: transforming heterogeneous operational information into interpretable diagnostic knowledge.

Operational intelligence remains predominantly a \gls{randr}-centric application, with learning mainly used to enhance reasoning, explainability, and diagnosis.

Overall, slow-timescale applications are dominated by \gls{randr} and \gls{pands}, reflecting their reliance on semantic understanding, forecasting, planning, and network-wide optimization. Although individual applications instantiate different combinations of functional capabilities, systems that jointly realize \gls{randr}, \gls{pands}, \gls{ac}, and \gls{ds} remain relatively uncommon.

%% file: sections/multi_timescale_applications/medium-time-scale.tex
\subsection{Medium-Timescale Control Applications} \label{subsec:medium_timescale} Medium-timescale applications form the intermediate layer between slow strategic optimization and fast radio-resource control; prior work likewise separates network optimization by timescale, although with different boundaries~\cite{li2022analytics,bui2017survey}. Operating over time horizons ranging from tens of milliseconds to seconds, these applications adapt network behavior to evolving traffic, mobility, interference, and service conditions. This timescale closely corresponds to the control loops of the O-RAN near-real-time \gls{ric} (10\,ms--1\,s)~\cite{oran2026oad}, where decisions must react to changing network conditions while retaining sufficient time for prediction and adaptive optimization. 

Consequently, medium-timescale applications naturally call for the interaction between \gls{pands} and \gls{ac}, although the surveyed works most often instantiate \gls{ac}. Prediction is required to anticipate evolving network conditions and their impact on future performance, while adaptive control continuously adjusts network behavior in response to observations and feedback. \gls{randr} increasingly appears through foundation-model-based reasoning and contextual understanding, whereas \gls{ds} remains less common and is typically associated with resource-allocation, orchestration, and cell-shaping functions. 

The following applications illustrate how different combinations of \gls{randr}, \gls{pands}, \gls{ac}, and \gls{ds} emerge in medium-timescale RAN control.

\textbf{Mobility and load management.} Mobility and load management comprise handover management, load balancing, dual connectivity, and mobility robustness optimization~\cite{haghrah2023survey,tashan2022mobility}, as well as user association~\cite{Tao2026LargeVisionModel}. Since these functions must react to evolving traffic and mobility conditions while anticipating future network states, they naturally call for \gls{pands} and \gls{ac}, although the surveyed works mostly instantiate \gls{ac} and \gls{randr} (Table~\ref{tab:mobility_management_mapping}). 
The literature exhibits three dominant combinations of functional capabilities. The first relies on digital twins or channel predictors, either to train reinforcement-learning association policies for load balancing~\cite{Tao2026LargeVisionModel} or to forecast channel quality and trigger proactive handovers~\cite{li2026llm}. The second formulates mobility management as an adaptive-control problem, where reinforcement learning tunes handover parameters for mobility robustness in simulation, either offline from previously collected datasets~\cite{alizadeh2025offline} or by interacting with the simulated network~\cite{kumar2025advanced}. The third introduces foundation models and structured network knowledge into mobility management, either to generate handover decisions directly~\cite{lien2026foundation} or to support knowledge-driven handover coordination~\cite{rao2025optimization}. 

Mobility management is primarily dominated by \gls{ac} and \gls{randr}, with prediction used mainly to support proactive handover. Explicit \gls{ds} capabilities remain largely absent.

\begin{table}[t] 
\caption{Instantiation of the four functional capabilities in mobility and load management.}
\label{tab:mobility_management_mapping} 
\centering 
\renewcommand{\arraystretch}{1.1} 
\begin{tabular}{lcccc} 
\toprule 
\textbf{Work} & 
\textbf{R\&R} & 
\textbf{P\&S} & 
\textbf{AC} & 
\textbf{DS} \\ 
\midrule
Tao et al.~\cite{Tao2026LargeVisionModel} 
& \xmark & \cmark & \cmark & \xmark \\ 
Li et al.~\cite{li2026llm} 
& \cmark & \cmark & \xmark & \xmark \\ 
Alizadeh et al.~\cite{alizadeh2025offline}, \cite{kumar2025advanced}
& \xmark & \xmark & \cmark & \xmark \\ 
Lien et al.~\cite{lien2026foundation} 
& \cmark & \xmark & \cmark & \xmark \\ 
Rao et al.~\cite{rao2025optimization} 
& \cmark & \xmark & \xmark & \xmark \\ 
\bottomrule 
\end{tabular} 
\end{table}

\textbf{Interference and cell-shaping control.} Interference and cell-shaping control regulate the spatial footprint of the network through decisions such as antenna tilting~\cite{forsberg2023network}, beam shaping~\cite{wu2006tilting}, power adaptation~\cite{vaezi2023deep}, RIS configuration~\cite{ghassemi2026FoundationModelAided}, and adaptive initial-access design~\cite{aroua2025adaptive}. These functions continuously reshape coverage, interference conditions, and traffic distribution across neighboring cells, making them a natural medium-timescale control problem (Table~\ref{tab:cell_shaping_mapping}).

The literature exhibits three main capability patterns. Early approaches relied on explicit optimization and heuristic control mechanisms to adapt coverage and balance traffic through antenna tilting and beam shaping \cite{wu2006tilting}. These solutions primarily instantiate the \gls{ds} block by directly selecting deployable network configurations.

More recent work formulates cell shaping as an adaptive-control problem. Graph-based multi-agent reinforcement learning adapts antenna tilts to interference conditions under safety constraints~\cite{forsberg2023network}, and reinforcement learning adapts initial-access beam codebooks to changing user distributions~\cite{aroua2025adaptive}, making \gls{ac} the dominant capability. These methods continuously adapt network behavior rather than relying on fixed optimization rules.

A third direction incorporates foundation models into radio-environment shaping. In foundation-model-assisted power control, an \gls{llm} is prompted with past states, decisions, and rewards~\cite{zhou2025prompting}, while in blockage-aware RIS control, channel representations learned by a foundation model are fed, together with user location and blockage status, to a hierarchical deep reinforcement learning agent~\cite{ghassemi2026FoundationModelAided}. These systems primarily instantiate the \gls{randr}+\gls{ac} pattern and illustrate the growing role of reasoning capabilities in medium-timescale radio control.

This application class illustrates the evolution from deterministic optimization toward adaptive and reasoning-enhanced control, with \gls{ac} emerging as the dominant capability, \gls{ds} still present in most works, and \gls{randr} increasingly complementing it.

\begin{table}[t] 
\caption{Instantiation of the four functional capabilities in interference and cell-shaping control.}
\label{tab:cell_shaping_mapping} 
\centering 
\renewcommand{\arraystretch}{1.1} \begin{tabular}{lcccc} \toprule \textbf{Work} & \textbf{R\&R} & \textbf{P\&S} & \textbf{AC} & \textbf{DS} \\ \midrule Wu et al.~\cite{wu2006tilting} & \xmark & \xmark & \xmark & \cmark \\ Forsberg et al.~\cite{forsberg2023network} & \xmark & \xmark & \cmark & \cmark \\ Zhou et al.~\cite{zhou2025prompting} & \cmark & \xmark & \cmark & \xmark \\ Ghassemi et al.~\cite{ghassemi2026FoundationModelAided} & \cmark & \xmark & \cmark & \xmark \\ Aroua et al.~\cite{aroua2025adaptive} & \xmark & \xmark & \cmark & \cmark \\ \bottomrule \end{tabular} \end{table}

\textbf{Dynamic resource, slice, and traffic-aware adaptation.} This class of applications dynamically allocates radio, compute, and network resources across heterogeneous services while adapting to evolving traffic conditions. Unlike mobility management, which determines user--cell associations, or interference-control mechanisms, which reshape the radio environment, these approaches focus on how resources should be distributed among users, network slices, and AI workloads. As shown in Table~\ref{tab:resource_slice_mapping}, this application class is one of the few medium-timescale domains that spans all four functional capabilities.

The literature exhibits four main patterns of capability integration. The first combines \gls{randr} and \gls{ac}, using foundation models to enrich resource-allocation and slice-management policies through semantic reasoning and domain knowledge \cite{lotfi2025prompt,lotfi2025oran}. The second adds \gls{ds} to \gls{randr} and \gls{ac}, where foundation models interact with optimization or orchestration mechanisms to generate executable resource allocations~\cite{wu2025llm}.

The third introduces predictive capabilities into resource management. SliceWM~\cite{zhou2025world} learns network-slice dynamics and trains allocation policies using imagined trajectories, providing a representative example of the \gls{pands}+\gls{ac} pattern.

Finally, systems such as T-ELLM~\cite{tan2025tool} move toward richer integrations of semantic reasoning, adaptive control, and solver-based resource allocation, with a learned virtual environment used to train its policy offline. Apart from these four patterns, several task-offloading, model-placement, and mobile-LLM-serving frameworks jointly optimize decisions such as offloading, computing-resource shares, model caching, prompt compression, and transmission power, mainly through \gls{rl}- or \gls{llm}-driven adaptive control~\cite{Wu2026LlmModelBasedTaskOffloading,Xu2026ServingLongContextLLMs,You2026JPPO}.

Dynamic resource and slice adaptation therefore provides one of the clearest examples of how the four capabilities can interact in practice, acting as a natural meeting point between AI-native services and AI-native network control.

\begin{table}[t] 
\caption{Instantiation of the four functional capabilities in dynamic resource, slice, and traffic-aware adaptation.}
\label{tab:resource_slice_mapping} 
\centering 
\renewcommand{\arraystretch}{1.1} 
\begin{tabular}{lcccc} 
\toprule 
\textbf{Work} & 
\textbf{R\&R} & 
\textbf{P\&S} & 
\textbf{AC} & 
\textbf{DS} \\ 
\midrule

PA-MRL~\cite{lotfi2025prompt} 
& \cmark & \xmark & \cmark & \xmark \\ 
ORAN-GUIDE~\cite{lotfi2025oran} 
& \cmark & \xmark & \cmark & \xmark \\
SliceWM~\cite{zhou2025world} 
& \xmark & \cmark & \cmark & \xmark \\ 
T-ELLM~\cite{tan2025tool} 
& \cmark & \cmark & \cmark & \cmark \\ 
LLM-xApp~\cite{wu2025llm} 
& \cmark & \xmark & \cmark & \cmark \\ 
Task Offloading~\cite{Wu2026LlmModelBasedTaskOffloading} 
& \cmark & \xmark & \cmark & \cmark \\ 
T2DRL~\cite{Xu2026ServingLongContextLLMs} 
& \xmark & \xmark & \cmark & \xmark \\ 
JPPO++~\cite{You2026JPPO} 
& \cmark & \xmark & \cmark & \xmark \\ 
\bottomrule 
\end{tabular} 
\end{table}

\textbf{Security and resilient control.} Security is often a cross-layer concern, as attackers can target several protocol layers at once~\cite{lu2022reinforcement}. In \gls{ran}s, security also spans multiple control timescales. At the medium timescale, security primarily manifests as a network-control problem, where abnormal or malicious behavior must be detected and mitigation policies adapted to evolving network conditions. This observe--detect--mitigate loop naturally emphasizes adaptive decision making and policy reconfiguration (Table~\ref{tab:security_mapping}). 

The surveyed literature relies mainly on the \gls{ac} and \gls{pands} blocks. Zero-trust and federated-learning frameworks protect O-RAN optimization and resource-management functions against adversarial behavior through adaptive security mechanisms and distributed learning \cite{el2025secure}. In addition, an AI-driven traffic-steering xApp that acts on forecasts of cell load and user trajectories has been run over an E2 interface whose quantum-safe IPsec protection adds only 0.818~ms of average latency~\cite{behravesh2025ai}. 

A second direction introduces foundation models to improve security interpretation, reasoning, and explainability. For example, \cite{Mojdeh2025TowardsSecureOran} investigates the role of \gls{llm}s in identifying vulnerabilities, supporting resilient control strategies, and strengthening security-aware O-RAN orchestration. More broadly, recent studies explore the opportunities and risks associated with intelligent controllers in AI-native Open RAN systems \cite{soltani2025intelligent}.

This application class relies mainly on \gls{ac} and \gls{pands}, with \gls{randr} introduced when foundation models are used for security interpretation, reasoning, or explainability. \gls{pands} appears mainly as load and mobility forecasting, and explicit \gls{ds} capabilities remain largely absent, highlighting opportunities for threat forecasting, coordinated mitigation, and security-aware decision synthesis.

\begin{table}[t] 
\caption{Instantiation of the four functional capabilities in security and resilient control.} \label{tab:security_mapping} \centering \renewcommand{\arraystretch}{1.1} \begin{tabular}{lcccc} \toprule \textbf{Work} & \textbf{R\&R} & \textbf{P\&S} & \textbf{AC} & \textbf{DS} \\ \midrule Karbalaee Motalleb et al.~\cite{Mojdeh2025TowardsSecureOran} & \cmark & \xmark & \cmark & \xmark \\ El-Hajj~\cite{el2025secure} & \xmark & \cmark & \cmark & \xmark \\ Behravesh et al.~\cite{behravesh2025ai} & \xmark & \cmark & \xmark & \xmark \\ \bottomrule \end{tabular} 
\end{table}

Medium-timescale applications are dominated by \gls{ac}, reflecting the need to continuously adapt network behavior to evolving traffic, mobility, and interference conditions. While \gls{randr} is increasingly introduced through foundation-model-based approaches, \gls{pands} appears mainly in prediction-driven applications, and \gls{ds} remains largely confined to resource-allocation, orchestration, and cell-shaping functions. Consequently, most existing solutions realize only a subset of the four functional capabilities.

%% file: sections/multi_timescale_applications/fast-time-scale.tex
\subsection{Fast-Timescale Control Applications} \label{subsec:fast_timescale} 

Fast-timescale control forms the lowest layer of the RAN control hierarchy, with decision horizons ranging from sub-milliseconds to a few milliseconds \cite{lacava2025dapps}. It operates directly on radio observations such as the rapidly varying channel state and channel-quality feedback~\cite{gao2026ai,xu2025fully}, interference measurements, buffer status reports, and HARQ feedback~\cite{3gpp38214,3gpp38321}. Representative functions include link adaptation, transmit-power control, beamforming and precoding, HARQ, frequency scheduling, and per-slot resource allocation~\cite{11771692, 3gpp38211,3gpp38213,3gpp38214,mismar2019deep}.

Although often studied separately, these functions are tightly coupled in practice: scheduling decisions alter interference conditions, interference affects link adaptation and power control, while beamforming directly influences the channel conditions seen by scheduling and resource-allocation algorithms. Furthermore, all operate under comparable latency constraints and are ultimately driven by the same objective of optimizing instantaneous radio-resource utilization. For this reason, we consider them collectively as a single application class referred to as \textit{fast radio-resource control}.

This layer faces stringent latency and computational constraints: 6G \gls{urllc} targets sub-millisecond latency, and its mechanisms add processing burdens at base stations with finite computational resources~\cite{bajracharya2026achieving}. We argue that these constraints significantly limit the use of computationally intensive reasoning and long-horizon prediction. As a result, \gls{ac} emerges as the dominant functional capability, while \gls{randr}, \gls{pands}, and \gls{ds} are practical only when their computational overhead remains compatible with real-time operation.

\textbf{Fast radio-resource control.} Fast radio-resource control is predominantly an adaptive-control problem~(Table~\ref{tab:fast_control_mapping}).
\gls{rl}-based approaches have been proposed for dynamic link adaptation and power control under stringent reliability and latency requirements \cite{guo2023reinforcement,salehi2025towards}. Safe \gls{rl} has also been investigated for \gls{eeirp}-constrained power control in upper-6 GHz spectrum sharing \cite{11771692}. A Lagrangian \gls{ppo} agent performs adaptive power control, while a rule-based compliance layer validates candidate actions and replaces constraint-violating outputs with a safe fallback allocation. This example combines \gls{ac} and \gls{ds} within a fast-timescale control loop and illustrates a hybrid AI approach in which learning is complemented by explicit decision-governance mechanisms. Similarly, AI-driven scheduling and resource-allocation frameworks continuously adapt transmission decisions to rapidly varying channel and traffic conditions \cite{OnlineFS}. AI-assisted beamforming and spatial control have also been investigated for MU-MIMO and RIS-enhanced systems under imperfect channel state information and user mobility, but with models trained in advance rather than adapted from feedback~\cite{farhadi2025ai,ismail2026ai}.

While \gls{ac} remains the dominant capability, recent work has begun to incorporate additional functional capabilities. An \gls{llm} used as a few-shot power-allocation solver brings \gls{randr} into transmission decisions, with a conventional fallback for invalid or weaker outputs (a simple form of \gls{ds}), but needs seconds rather than milliseconds per decision~\cite{lee2026llm}. A recurrent long short-term memory (LSTM) predictor enables limited \gls{pands} capabilities by forecasting the user's next angular position to steer an RIS beam~\cite{ismail2026ai}. In addition, a deep Q-learning scheduler coordinates its parallel per-sub-band user selections through a shared learned representation and value decomposition~\cite{OnlineFS}, but this coordination is learned inside the Q-network rather than performed by an explicit \gls{ds} mechanism.

\begin{table}[t] 
\caption{Instantiation of the four functional capabilities in fast radio-resource control.}
\label{tab:fast_control_mapping} 
\centering \renewcommand{\arraystretch}{1.1} \begin{tabular}{lcccc} \toprule \textbf{Work} & \textbf{R\&R} & \textbf{P\&S} & \textbf{AC} & \textbf{DS} \\ \midrule Guo et al.~\cite{guo2023reinforcement} & \xmark & \xmark & \cmark & \xmark \\ Salehi et al.~\cite{salehi2025towards} & \xmark & \xmark & \cmark & \xmark \\ Giovanidis et al.~\cite{OnlineFS} & \xmark & \xmark & \cmark & \xmark \\ Ismail et al.~\cite{ismail2026ai} & \xmark & \cmark & \xmark & \xmark \\ Lee et al.~\cite{lee2026llm} & \cmark & \xmark & \xmark & \cmark \\ 
Aroua et al.~\cite{11771692} & \xmark & \xmark & \cmark & \cmark \\ 
\bottomrule 
\end{tabular} \end{table}

In fast-timescale applications, \gls{ac} is the most frequent functional capability, reflecting the need for continuous adaptation under stringent latency constraints. While \gls{randr} and \gls{pands} have begun to appear in selected applications, explicit \gls{ds} mechanisms remain comparatively uncommon. Most existing systems deploy actions directly from adaptive policies, with feasibility typically assessed through offline analysis or simulation rather than through dedicated synthesis mechanisms. 

Recent work has nevertheless demonstrated that \gls{ds} can be integrated into fast-timescale control loops when operational constraints must be guaranteed. For example, feasibility-aware governance layers can validate candidate actions and replace constraint-violating outputs with safe fallback decisions before deployment \cite{11771692}. Such approaches combine learning-based adaptation with explicit \gls{ds}, illustrating a hybrid AI paradigm in which adaptive controllers generate candidate actions while governance mechanisms ensure operational or regulatory compliance. 

Despite these advances, \gls{ds} in fast-timescale RAN control remains largely underexplored. Existing solutions primarily focus on enforcing local constraints, while broader functions such as reconciliation of multiple controllers, conflict resolution, cross-agent coordination, and multi-objective governance have received comparatively little attention. Addressing these challenges will be essential for coordinating capabilities across multiple control loops and timescales in future AI-native RANs.

%% file: sections/multi_timescale_applications/multi-time-scale.tex
\subsection{Cross-Timescale and Multi-Capability Systems}
\label{Cross-Time-Scale}

The application classes discussed thus far primarily focus on individual control horizons. However, AI-native RANs increasingly require coordination across multiple timescales, where strategic decisions influence Near-RT and real-time operations, while observations from lower layers continuously refine higher-level policies. Such settings motivate systems that combine multiple functional capabilities and coordinate decision making across heterogeneous control loops.

The systems summarized in Table~\ref{tab:cross_timescale_mapping} represent some of the closest existing realizations of the functional perspective developed in this survey. Unlike most application-specific solutions, which typically instantiate only a subset of the four capabilities, several of these systems integrate reasoning, prediction, adaptation, and decision orchestration across multiple layers of the RAN control hierarchy.

A multi-scale agentic O-RAN framework~\cite{navidan2026toward} distributes intelligence across Non-RT, Near-RT, and real-time control loops through cooperating LLM agents, SLM agents, and wireless foundation models. Higher-layer agents perform intent interpretation and policy generation, realizing \gls{randr}, while Digital-Twin-assisted evaluation provides \gls{pands}. Adaptive optimization is delegated to Near-RT agents, corresponding to \gls{ac}, whereas policy governance and cross-layer coordination contribute to \gls{ds}. The framework therefore provides an explicit realization of all four functional capabilities across the O-RAN control hierarchy.

Similarly, Agentic AI-RAN~\cite{he2026agentic} structures network control around Plan--Act--Observe--Reflect loops, memory, Digital Twin support, and skill orchestration. Intent grounding and planning contribute to \gls{randr}, Digital-Twin-assisted what-if analysis realizes \gls{pands}, adaptive control loops implement \gls{ac}, and skill sequencing, rollback, and self-management mechanisms provide \gls{ds}. As a result, the system spans all four functional capabilities across Non-RT, Near-RT, and real-time control loops.

AgentRAN~\cite{elkael2026agentran} introduces a hierarchy of LLM-based agents generated from natural-language intents. Intent interpretation and decomposition realize \gls{randr}, while distributed scheduling, power-control, and resource-management agents provide \gls{ac}. Hierarchical coordination enables high-level objectives to be translated into executable actions, corresponding to \gls{ds}. Unlike the Digital-Twin-enabled frameworks, AgentRAN provides \gls{pands} through a learned predictor: in its AI-RAN Factory experiment, an xApp predicts whether an interference alert will occur in the next second, and a predicted alert triggers preemptive blocking of the overlapping physical resource blocks (PRBs). AgentRAN also coordinates decisions from sub-millisecond operations to longer-term orchestration.

In contrast, the meta-hierarchical reinforcement-learning framework of~\cite{lotfi2025meta} focuses primarily on adaptive resource management. A high-level controller allocates resources across slices, while lower-level agents perform intra-slice scheduling and optimization. Consequently, the system is predominantly \gls{ac}-centric, without explicit semantic reasoning or predictive simulation capabilities.

Beyond conventional RAN control, the AI-RAN convergence architecture of~\cite{polese2026beyond} extends orchestration to the joint management of communication services and AI workloads over shared infrastructure. Its AI-RAN Orchestrator translates the operator policy into resource allocations, and AI workloads are admitted only if the required resources are compatible with the orchestrator's policy, which corresponds to \gls{ds}; the article does not specify reasoning, prediction, or learning-based adaptation mechanisms. This system shows how the four functional capabilities can extend beyond traditional RAN optimization toward AI-RAN convergence.

Finally, E-ARC~\cite{aroua2026ran} targets autonomous energy-aware optimization through semantic intent interpretation, Digital-Twin-assisted validation, and rApp orchestration. A Semantic Coordination agent interprets high-level objectives, realizing \gls{randr}, while a Digital Twin agent evaluates candidate strategies and configurations, providing \gls{pands}. The Semantic Coordination agent then selects the rApps and their Digital-Twin-validated parameters, and the Configuration, Deployment and Monitoring agent configures, deploys, and monitors these rApps to achieve the desired energy-performance trade-off, thereby realizing \gls{ds}. The adaptive-control capability is subsequently provided by the selected control applications themselves, such as traffic-steering and UX-aware optimization rApps.

Overall, Table~\ref{tab:cross_timescale_mapping} highlights a clear evolution from adaptive-control-centric solutions toward systems that increasingly integrate reasoning, prediction, adaptation, and orchestration across multiple timescales. Nevertheless, systems that explicitly realize all four functional capabilities across multiple RAN control loops remain relatively uncommon, highlighting a key challenge for future AI-native RAN control systems.

\begin{table*}[t] \caption{Cross-timescale systems and their instantiation of the four functional capabilities.} \label{tab:cross_timescale_mapping} \centering \renewcommand{\arraystretch}{1.1} \begin{tabular}{lccccc} \toprule \textbf{System} & \textbf{R\&R} & \textbf{P\&S} & \textbf{AC} & \textbf{DS} & \textbf{Control Timescales} \\ \midrule Multi-scale Agentic O-RAN~\cite{navidan2026toward} & \cmark & \cmark & \cmark & \cmark & Non-RT / Near-RT / RT \\ Agentic AI-RAN~\cite{he2026agentic} & \cmark & \cmark & \cmark & \cmark & Non-RT / Near-RT / RT \\ AgentRAN~\cite{elkael2026agentran} & \cmark & \cmark & \cmark & \cmark & Sub-ms to minutes \\ Meta-HRL~\cite{lotfi2025meta} & \xmark & \xmark & \cmark & \xmark & Slice-level / scheduling-level \\ AI-RAN Convergence~\cite{polese2026beyond} & \xmark & \xmark & \xmark & \cmark & Multi-timescale orchestration \\ E-ARC~\cite{aroua2026ran} & \cmark & \cmark & \cmark & \cmark & Non-RT (SMO / Non-RT RIC) \\ \bottomrule \end{tabular} \end{table*}

%% file: sections/benchmarks_datasets/main2.tex
\section{Benchmarks, Datasets, and Evaluation}
\label{sec:benchmarksDatasets}

Section~\ref{sec:applications} showed that \gls{ran} applications instantiate different, partial subsets of the functional blocks. Since each application is evaluated on its own terms, meaningful comparison requires common evaluation resources. Existing evaluation resources can be classified along two independent dimensions. 

The first dimension is \emph{decision-making depth}, which captures how much of the decision-making pipeline is covered by an evaluation resource. Resources range from those supporting individual functional blocks, such as reasoning or prediction, to those enabling the evaluation of complete closed-loop control. Along this dimension, we distinguish five resource types that form an increasing continuum of decision-making depth, progressing from passive observations to full closed-loop interaction:
A data substrate provides network data without defining a task. A capability benchmark evaluates a specific decision-making capability, such as state understanding, reasoning, or prediction, independently of the broader control process. A state-conditioned decision benchmark evaluates judgments or decisions generated from fixed network states without executing them in the network. A dynamics model estimates the consequences of candidate actions by modeling network evolution, but does not include an acting agent. Finally, an interactive decision environment allows an agent to repeatedly observe, decide, and act while receiving feedback from the network, thereby supporting closed-loop evaluation. Only the latter captures the complete decision-making process required for end-to-end AI-native RAN control.

The second dimension is \emph{benchmark reproducibility}, which captures whether results obtained using a given evaluation resource can be reliably reproduced and meaningfully compared across studies. Reproducibility requires more than access to a common dataset or environment; it requires a well-defined evaluation protocol. In particular, a benchmark should specify a task, an evaluation metric, a fixed data split and seed protocol\footnote{A test \emph{split} fixes which examples of a dataset are used for evaluation. A \emph{seed} initializes a simulator's random-number generator and therefore fixes stochastic elements such as traffic arrivals, channel conditions, and user locations. A seed protocol specifies the seeds that must be used during evaluation, ensuring that different studies are assessed under identical conditions.}, and one or more reported baselines against which new methods can be compared. Resources that do not define a task and evaluation procedure should therefore be viewed as infrastructure rather than benchmarks.

Importantly, benchmark reproducibility and decision-making depth are independent dimensions. An evaluation resource may support sophisticated closed-loop interaction while lacking a standardized evaluation protocol, or conversely provide a rigorous benchmarking methodology while evaluating only a narrow aspect of the decision-making process. For example, Colosseum~\cite{Bonati2021Colosseum} enables interactive closed-loop experimentation but does not prescribe a specific evaluation task, whereas TeleQnA~\cite{Maatouk_2026_TeleQnA} defines a complete benchmarking protocol while remaining a single-turn evaluation benchmark.
Table~\ref{tab:benchmarks_at_a_glance} classifies the surveyed resources according to their decision-making depth and indicates the functional blocks they primarily support. The corresponding subsections discuss the reproducibility characteristics of their evaluation protocols. Standardization efforts and emerging benchmarking initiatives that are not yet publicly available are discussed qualitatively rather than included in the comparison table.

This analysis reveals a key gap in the current ecosystem. Well-defined and reproducible evaluation protocols are increasingly available for individual capabilities and one-shot decision-making tasks, while several environments support closed-loop experimentation. However, few resources combine closed-loop control with standardized tasks, common metrics, reference baselines, and reproducible evaluation procedures. As a result, the community still lacks widely accepted benchmarks for rigorous and comparable evaluation of end-to-end AI-native RAN control.

\begin{table*}[!t]
\caption{Evaluation resources for hybrid AI in RAN control.}
\label{tab:benchmarks_at_a_glance}
\centering
\footnotesize
\renewcommand{\arraystretch}{1.12}
\begin{tabular}{p{3.4cm} p{1.9cm} p{2.5cm} p{1.2cm} p{4.0cm}}
\toprule
\textbf{Resource} &
\textbf{Type} &
\textbf{Block(s)} &
\textbf{Time} &
\textbf{Access / Notes} \\
\midrule

\rowcolor{gray!10}
\multicolumn{5}{l}{\textbf{Level 1: Data Substrates} --- observations only, no declared task} \\

SPEC5G~\cite{karim2023Spec5G}
& Substrate & R\&R & --- 
& Open; tables/equations removed \\

TSpec-LLM~\cite{Nikbakht2024TSpecLlm}
& Substrate & R\&R & --- 
& Open; preserves structured content \\

Tele-Data~\cite{maatouk2025TeleLlm}
& Substrate & R\&R & --- 
& Open \\

OpenTelecom~\cite{zou2025telecomgpt}
& Substrate & R\&R & --- 
& Release status not stated \\

CommData~\cite{Jiang2026CommGPT}
& Substrate & R\&R & --- 
& Release status not stated, Text-only corpus \\

DeepMIMO~\cite{Alkhateeb2019DeepMimo}
& Substrate & P\&S & Fast 
& Open; ray-traced channels \\

DeepSense 6G~\cite{Alkhateeb2023DeepSense6G}
& Substrate & P\&S & Fast 
& Open; real measurements \\

Milan~\cite{Barlacchi2015MilanDataset}
& Substrate & P\&S & Slow 
& Open; collected in 2013 \\

NetMob23~\cite{Orlando2023NetMob23Dataset}
& Substrate & P\&S & Slow 
& Challenge-based access \\

OpenRAN Gym traces~\cite{Bonati2023OpenRANGym,Polese2023ColORan}
& Substrate & AC & Medium 
& Open; KPM traces under fixed policies \\

Twinning~\cite{Bonati2024Twinning}
& Substrate & P\&S & Medium 
& Open; $>$500 h, 450 GB \\

\midrule

\rowcolor{gray!10}
\multicolumn{5}{l}{\textbf{Level 2: Capability Benchmarks} --- one query, one scored answer} \\

TeleQnA~\cite{Maatouk_2026_TeleQnA}
& Capability & R\&R & --- 
& Open; 10k MCQ \\

ORAN-Bench-13K~\cite{Gajjar_2025_ORAN_Bench_13K}
& Capability & R\&R & --- 
& Open; 13.9k MCQ \\

TeleMath~\cite{Colle_2026_TeleMath}
& Capability & R\&R & --- 
& Open; numerical answers \\

TeleTables~\cite{Ezzakri_2025_TeleTables}
& Capability & R\&R & --- 
& Open \\

Tele-Eval~\cite{maatouk2025TeleLlm}
& Capability & R\&R & --- 
& Open, no fixed split or seed protocol \\

srsRANBench~\cite{Gajjar_2025_ORANSight2}
& Capability & R\&R & --- 
& Open; code-generation tasks \\

TeleCom-Bench~\cite{Xiao2026TeleComBench}
& Capability & R\&R & --- 
& Partially released \\

CSI-RRG~\cite{Cheng2025CsiRrg}
& Capability & P\&S & Fast 
& Open, 3,060 scenarios; TDD/FDD \\

\midrule

\rowcolor{gray!10}
\multicolumn{5}{l}{\textbf{Level 3: State-Conditioned Decision Benchmarks}} \\

TeleLogs~\cite{Sana_2025_TeleLogs}
& State-cond. Decision & R\&R & --- 
& Open, synthetic troubleshooting logs \\

TelecomTS~\cite{Feng2025TelecomTS}
& State-cond. Decision & R\&R, P\&S & Medium 
& Open, real 5G observability data \\

\midrule

\rowcolor{gray!10}
\multicolumn{5}{l}{\textbf{Level 4: Dynamics Models}} \\

Sionna / Sionna RT~\cite{hoydis2023sionnaopensource,Jakob2023SionnaRt}
& Dynamics Model & P\&S & Fast 
& Open; differentiable simulation \\

AODT~\cite{nvidia_aodt}
& Dynamics Model & P\&S & Slow 
& Hybrid, platform-gated access \\

\midrule

\rowcolor{gray!10}
\multicolumn{5}{l}{\textbf{Level 5: Interactive Environments}} \\

ns-O-RAN~\cite{Lacava2023nsOran}
& Environment & AC & Medium 
& Open; E2 interface \\

Colosseum~\cite{Bonati2021Colosseum}
& Environment & AC & Medium 
& Shared testbed \\

SCOPE~\cite{Bonati2021Scope}
& Environment & AC & Medium 
& Open, runtime control APIs \\

ns-O-RAN Gym~\cite{Bordin2025NsOranGym}
& Environment & AC & Medium 
& Open, released task and baselines; no fixed split/seed \\

Wireless Suite~\cite{nokia-wireless-suite}
& Environment & AC & Fast 
& Open, complete evaluation protocol; no RIC \\

FORLORN~\cite{forlorn}
& Environment & AC & Medium 
& Open, complete evaluation protocol; no RIC \\

NetworkGym~\cite{Haider2024NetworkGym}
& Environment & AC & Medium 
& Open, complete evaluation protocol, adjacent domain \\

EdgeRIC~\cite{edgeric}
& Environment & AC & Fast 
& Open, seeding shipped disabled \\

\midrule

\rowcolor{gray!20}
\textbf{Missing Resource: Reproducible Closed-Loop RAN Benchmark}
& \textbf{Environment}
& \textbf{All}
& \textbf{Multi}
& \textbf{Currently unavailable} \\

\bottomrule
\end{tabular}

\vspace{1mm}
\begin{minipage}{\textwidth}
\footnotesize
\textbf{Blocks:} R\&R = Representation \& Reasoning,
P\&S = Prediction \& Simulation,
DS = Decision Synthesis,
AC = Adaptive Control.

\textbf{Time:}
Fast, medium, and slow correspond to the representative decision
horizon of the target application class.
\end{minipage}
\end{table*}

\subsection{Data Substrates: Corpora, Measurements, and Traces} \label{sec:benchmarksDatasetsPretraining} Data substrates represent the shallowest level of decision-making depth. They provide observations of telecom systems but do not define tasks, metrics, or evaluation protocols. Consequently, they serve as infrastructure for model development rather than benchmarks. We group existing data substrates into three families: telecom knowledge corpora, network measurements, and control-conditioned traces. 

The first family comprises telecom knowledge corpora derived from standards, technical documentation, and scientific publications. SPEC5G~\cite{karim2023Spec5G} established an initial standards-centric corpus, while TSpec-LLM~\cite{Nikbakht2024TSpecLlm} extended coverage across multiple 3GPP releases and preserved structured content such as tables and equations. Tele-Data~\cite{maatouk2025TeleLlm}, OpenTelecom~\cite{zou2025telecomgpt}, and CommData~\cite{Jiang2026CommGPT} further expanded scale, diversity, and alignment support. Collectively, these resources provide a knowledge substrate for \gls{randr} capabilities. 

The second family consists of measured or physically generated network observations. DeepMIMO~\cite{Alkhateeb2019DeepMimo} provides configurable ray-traced channel datasets, while DeepSense6G~\cite{Alkhateeb2023DeepSense6G} complements them with real-world mmWave and sensor measurements. For traffic modeling, the Milan dataset~\cite{Barlacchi2015MilanDataset} remains a widely used reference, whereas NetMob23~\cite{Orlando2023NetMob23Dataset} offers more recent service-level traffic observations. These datasets primarily support prediction and simulation by exposing the statistical characteristics and dynamics of wireless networks. 

The third family captures measurements collected under known scheduling, resource-allocation, or control configurations. OpenRAN Gym~\cite{Bonati2023OpenRANGym} and ColO-RAN~\cite{Polese2023ColORan} provide slice-level performance measurements under predefined scheduling and resource-allocation policies, while the Twinning dataset~\cite{Bonati2024Twinning} contributes more than 500 hours of multi-layer measurements and protocol logs collected on an Open RAN platform. These resources are the closest to decision-making data, yet observations remain conditioned on fixed configurations rather than on a sequence of agent interactions. 

Despite their diversity, all three families remain observational resources. They support representation learning, reasoning, prediction, and offline analysis, but they neither define evaluation tasks nor capture complete observation--action--feedback trajectories. Consequently, they should be viewed as enabling infrastructure rather than benchmarks, forming the foundation upon which higher-depth evaluation resources are built.

\subsection{Capability Benchmarks} 
\label{sec:benchmarksDatasetsKnowledge}

Capability benchmarks evaluate individual functional blocks in isolation by comparing model outputs against ground truth. Unlike data substrates, they define explicit tasks, evaluation metrics, and reference results, making them among the most reproducible evaluation resources currently available. Existing benchmarks predominantly target the \gls{randr} block, with additional coverage of the \gls{pands} block and more limited attention to the \gls{ac} and \gls{ds} blocks.

Most existing benchmarks focus on \gls{randr}. TeleQnA~\cite{Maatouk_2026_TeleQnA} has emerged as a reference benchmark, comprising 10{,}000 multiple-choice questions spanning telecom concepts and standards. Its finding that model performance degrades substantially on standards-specific questions continues to shape subsequent evaluations. ORAN-Bench-13K~\cite{Gajjar_2025_ORAN_Bench_13K} extends this paradigm to O-RAN knowledge, covering 116 specification documents and nearly 14{,}000 questions organized by difficulty. TeleMath~\cite{Colle_2026_TeleMath} focuses on mathematical reasoning through numerical-answer tasks, enabling unambiguous scoring and highlighting the benefits of reasoning-oriented adaptations. TeleTables~\cite{Ezzakri_2025_TeleTables} further evaluates understanding of structured 3GPP tables, revealing persistent challenges in both information retrieval and interpretation. 

Tele-Eval~\cite{maatouk2025TeleLlm} broadens evaluation beyond multiple-choice formats by introducing open-ended telecom question answering scored using embedding similarity and an \gls{llm}-based judge. However, despite reporting baseline results, it does not specify a fixed split or seed protocol. Moreover, because its evaluation items are synthesized from the same corpus family used for continual pretraining, the separation between training and evaluation is less clear-cut than in standardized benchmarks. 

A second category evaluates operationally oriented reasoning tasks that lie closer to \gls{ds} than traditional question-answering benchmarks. srsRANBench~\cite{Gajjar_2025_ORANSight2} evaluates code generation over an operational O-RAN stack, requiring models to translate telecom knowledge into executable artifacts. TeleCom-Bench~\cite{Xiao2026TeleComBench} further moves toward operational decision support by evaluating industrial telecom workflows. Its results reveal a substantial gap between linguistic understanding and procedural execution, with performance dropping from approximately 90\% on language-centric tasks to roughly 30\% on workflow-oriented tasks. While neither benchmark executes actions in a live environment, both assess capabilities that lie at the interface between reasoning and decision generation. 

The \gls{pands} block is represented by CSI-RRG~\cite{Cheng2025CsiRrg}, which evaluates channel prediction across more than 3{,}000 scenarios spanning channel models, delay spreads, mobility conditions, noise characteristics, and both TDD and FDD deployments. By combining publicly released datasets with standardized evaluation procedures, CSI-RRG demonstrates that reproducible capability benchmarking extends beyond language-centric tasks to prediction-driven network functions. 

With the exception of Tele-Eval's incomplete evaluation protocol and TeleCom-Bench's partial public release, these benchmarks define a task, a fixed test set, evaluation metrics, and reference results. They therefore satisfy the reproducibility criteria introduced earlier and currently constitute the most mature benchmarking ecosystem within AI-native telecom research. Their comparability is further strengthened by initiatives such as the GSMA Open Telco Harness~\cite{GsmaOpenTelco}, which reports leaderboard rankings and statistical confidence intervals under a common evaluation framework. 

Taken together, these benchmarks provide rigorous and reproducible evaluation of individual functional blocks. However, they remain single-turn assessments that evaluate capabilities in isolation. They do not capture the interaction between observations, decisions, actions, and subsequent network evolution. Consequently, they provide valuable measures of representation, reasoning, prediction, and decision-support capabilities, but cannot by themselves assess end-to-end decision-making behavior in AI-native RAN systems.

\subsection{State-Conditioned Decision Benchmarks}

State-conditioned decision benchmarks occupy an intermediate position between capability benchmarks and interactive environments. Unlike capability benchmarks, their outputs depend on a specific network state rather than on generic knowledge or reasoning alone. However, they stop short of action execution and therefore do not close the control loop. Instead, they evaluate state-dependent judgments that guide subsequent control decisions.

TeleLogs~\cite{Sana_2025_TeleLogs} formulates troubleshooting over synthetic drive-test logs as a network-diagnosis task, with annotations spanning eight root causes. Given network observations, a model must infer the underlying network condition responsible for the observed behavior. Although the benchmark does not require action selection, the inferred diagnosis directly informs subsequent operational decisions. The reported results show that current reasoning models struggle with these tasks, motivating the supervised and reinforcement-learning fine-tuning strategy proposed by the authors.

TelecomTS~\cite{Feng2025TelecomTS} provides a complementary benchmark based on multimodal observations collected from a real 5G testbed. It defines anomaly-detection, root-cause-analysis, and question-answering tasks over operational network data while preserving the absolute scale and variability of measured conditions. Compared with synthetic troubleshooting logs, TelecomTS exposes models to the complexity of real deployments, where current foundation-model baselines exhibit notable weaknesses, particularly under abrupt and highly dynamic network conditions.

Together, TeleLogs and TelecomTS evaluate state-conditioned judgments over both synthetic and measured network observations. They move beyond isolated capability evaluation by grounding inference in a specific network state, yet they stop before generating or executing actions. As a result, they assess decision-support functions rather than closed-loop decision making, highlighting the gap between understanding network conditions and controlling network behavior.

\subsection{Environments and Dynamics Models} 
Interactive decision environments represent the deepest level of decision-making depth. Unlike datasets and benchmarks, they allow an agent to observe network state, execute actions, and receive feedback from the resulting network evolution. ns-O-RAN~\cite{Lacava2023nsOran} couples ns-3 to a near-real-time RIC through the E2 interface, Colosseum~\cite{Bonati2021Colosseum} provides large-scale wireless-network emulation over 256 software-defined radios, and SCOPE~\cite{Bonati2021Scope} supplies the runtime control and data-collection framework underlying OpenRAN Gym and ColO-RAN. Together, these platforms enable closed-loop evaluation of adaptive RAN control strategies.

Despite this capability, two limitations persist. First, existing environments primarily support control-oriented interaction and provide limited support for higher-level \gls{randr} over intents, configurations, and operational knowledge. Second, they typically lack standardized tasks and evaluation protocols, making comparisons across studies difficult. PandORA~\cite{Tsampazi2024PandORA} partially addresses this issue by evaluating 23 DRL-based xApps on Colosseum across multiple decision timescales, demonstrating the feasibility of common benchmarking infrastructure. ns-O-RAN Gym~\cite{Bordin2025NsOranGym,NsOranGymRepo} moves further by defining a cell on/off energy-saving task, specifying throughput, energy, and reliability metrics, reporting baseline results, and releasing the environment. However, the absence of a fixed split or seed protocol still limits reproducibility.

A complementary class of resources provides network dynamics without maintaining an interaction loop. Sionna~\cite{hoydis2023sionnaopensource} offers differentiable communication-system simulation, while its ray-tracing extension~\cite{Jakob2023SionnaRt} incorporates physically grounded propagation models. NVIDIA's Aerial Omniverse Digital Twin~\cite{nvidia_aodt} extends this approach to city-scale deployments, enabling high-fidelity evaluation of network behavior under realistic operating conditions. Unlike interactive environments, these platforms do not maintain a persistent agent-facing state and therefore do not support closed-loop control. Within the taxonomy adopted in this survey, they are best interpreted as dynamics models serving the \gls{pands} block.

\subsection{Standardized Evaluation Methodology} 

An additional source of evaluation structure comes from standardization rather than research artifacts. 3GPP TS~28.105~\cite{3gpp28105} specifies AI/ML management procedures for 5G systems, while the Release-18 study TR~28.908~\cite{3gpp28908} identifies more than forty AI/ML use cases spanning the model lifecycle. At the air-interface level, TR~38.843~\cite{3gpp38843} defines evaluation methodologies and performance indicators for AI/ML-enabled radio functions. Collectively, these documents establish common lifecycle stages, management procedures, and evaluation criteria, thereby contributing to benchmark reproducibility across studies. 

However, these specifications define evaluation methodology rather than benchmark artifacts. They provide neither datasets nor environments, and they do not attach tasks, baselines, or scored evaluation protocols. Consequently, they serve as reference frameworks for evaluating Decision Synthesis capabilities rather than as benchmarks themselves. For this reason, they are discussed qualitatively rather than included in Table~\ref{tab:benchmarks_at_a_glance}.

\textit{Critical Analysis.} 
The primary evaluation gap is not a lack of data, simulators, or experimental platforms, but the absence of resources that simultaneously offer high decision-making depth and strong reproducibility. Capability and state-conditioned decision benchmarks already provide standardized tasks, metrics, and reference results, while several environments support realistic closed-loop interaction with \gls{ran} systems. However, these two properties rarely coexist within the same resource.

Most interactive \gls{ran} environments support closed-loop control but do not define complete evaluation protocols with fixed tasks, baselines, and reproducibility guidelines. Conversely, existing benchmarks provide rigorous and reproducible evaluation procedures but remain limited to single-turn assessments of individual capabilities. Among the surveyed resources, ns-O-RAN Gym comes closest to bridging this gap by defining a task, evaluation metrics, baseline policies, and a released environment. Nevertheless, the absence of a fixed split or seed protocol still limits direct comparison across studies.

This gap appears methodological rather than infrastructural. Adjacent efforts such as NetworkGym~\cite{Haider2024NetworkGym} demonstrate that interactive network environments can be paired with standardized tasks and reproducible evaluation procedures, while 6GAgentGym~\cite{Chen2026SixGAgentGym} illustrates how learned network dynamics can themselves support closed-loop evaluation. The immediate opportunity is therefore not necessarily to build new environments, but to transform existing \gls{ran} infrastructures into reproducible closed-loop benchmarks. Such resources would provide both the interaction required to evaluate end-to-end decision making and the methodological rigor needed for meaningful comparison across studies.

%% file: sections/future_directions/main2.tex
\input{sections/future_directions/main-f}

%% file: sections/future_directions/main-f.tex
\section{Lessons Learned and Future Directions}
\label{sec:lessonsLearned} 
The surveyed literature suggests that the evolution toward AI-native \gls{ran}s is not primarily a question of improving individual AI techniques, but of building coherent decision-making systems capable of operating across heterogeneous network functions, timescales, and objectives. 

Through the functional lens proposed in this survey, four major lessons emerge, each highlighting important directions for future research. 

\subsection{AI-Native RAN Requires Hybrid Intelligence}

A central observation is that AI-native \gls{ran} control is fundamentally a distributed, multi-capability decision-making problem. Modern RANs comprise multiple interacting control functions and intelligent agents, each responsible for optimizing specific objectives such as scheduling, mobility management, interference mitigation, energy efficiency, or resource allocation. Although these functions often target distinct operational goals, they remain tightly coupled through shared network resources, dynamics, and performance objectives. As a result, effective network operation cannot be achieved through isolated decision-making processes. 

Supporting autonomous operation therefore requires systems capable of representing and reasoning about network conditions, anticipating their evolution, adapting through interaction, and synthesizing decisions that satisfy operational constraints. These requirements correspond to the four functional capabilities identified in this survey: Representation and Reasoning (R\&R), Prediction and Simulation (P\&S), Adaptive Control (AC), and Decision Synthesis (DS). 

Importantly, these capabilities are not instantiated only once within the RAN. A given capability may be required simultaneously by multiple control functions operating at different locations, scopes, or timescales. For example, several agents may maintain their own representations of network state, perform localized reasoning, or execute adaptive control policies for distinct optimization tasks. Consequently, the challenge extends beyond implementing individual capabilities to coordinating their interaction, maintaining consistency across distributed decision makers, and aligning local decisions with system-wide objectives. 

The surveyed literature further indicates that no single AI paradigm naturally spans all four capabilities. Foundation models support knowledge representation, reasoning, and generalization; world models provide predictive and counterfactual simulation; reinforcement learning enables adaptation through interaction; and optimization methods ensure feasibility and constraint satisfaction. These paradigms should therefore be viewed as complementary rather than competing approaches. 

Future research should focus on Hybrid AI architectures that integrate multiple capabilities and paradigms within a unified decision-making framework. A key challenge is to establish explicit interfaces between components, enabling information, uncertainty estimates, predictions, decisions, and feedback to be exchanged consistently throughout the control process while supporting coordination among multiple distributed agents and control functions.

\subsection{Coordination Across Timescales and Agents Is the Central Systems Challenge}

RAN control operates across heterogeneous timescales, ranging from near-real-time scheduling and link adaptation to longer-horizon network optimization and energy management. Decisions taken at one timescale influence the operating conditions, objectives, and feasible actions of controllers operating at others. At the same time, the increasing deployment of distributed xApps, rApps, and intelligent controllers introduces interactions among multiple decision-making agents that may share resources, compete for control authority, or pursue partially conflicting objectives. 

This makes coordination a first-class systems problem rather than a secondary implementation concern. A key challenge for future AI-native \gls{ran}s is to enable autonomous operation across multiple control loops while preserving alignment with network-wide objectives and operational constraints. Rather than optimizing individual functions in isolation, intelligent agents must continuously exchange information, anticipate the impact of their decisions on other controllers, and adapt their behavior to the evolving network context. 

Future research should therefore investigate architectures for autonomous multi-agent and multi-timescale control, including hierarchical decision-making, distributed planning, conflict detection and resolution, and objective alignment mechanisms. Particular attention should be given to ensuring that local decisions remain consistent with global network goals despite uncertainty, partial observability, communication constraints, and the dynamic nature of network environments. Ultimately, achieving scalable autonomy in AI-native \gls{ran}s will require mechanisms that coordinate independent agents while maintaining coherent behavior across fast, medium, and slow control timescales.

\subsection{World Models Enable Proactive and Safer Control} 

A recurring observation throughout the surveyed literature is that predictive simulation remains substantially less explored than reasoning and adaptive control. While many existing approaches react to observed network conditions, relatively few explicitly evaluate the consequences of candidate actions before deployment. This limitation is particularly important in AI-native \gls{ran}s, where multiple intelligent agents operating at different timescales continuously influence both the network state and one another.

World models offer a promising path toward proactive network intelligence. Rather than merely predicting future network states, RAN world models can serve as predictive mirrors of the network, capturing its evolution under the actions of multiple interacting controllers. Such models represent not only the physical and logical state of the network, but also the effects of decisions taken by agents operating across heterogeneous timescales, from near-real-time control loops to longer-horizon optimization and planning functions. In this sense, world models extend conventional digital twins from descriptive platforms into decision-support systems capable of anticipating the network-wide consequences of distributed decision making.

Recent advances in multi-timescale world modeling suggest a promising foundation for this vision. By maintaining representations of network dynamics at different temporal abstractions, world models can capture both fast operational effects and slower structural trends while preserving their interactions. Such capabilities are particularly relevant for AI-native \gls{ran}s, where decisions taken by long-term optimization processes influence the operating conditions of short-term controllers and vice versa.

Beyond prediction and planning, world models can also support governance within autonomous control architectures. By simulating the consequences of candidate actions before deployment, they provide a mechanism for assessing compliance with operational constraints, service-level objectives, safety requirements, and network policies. In this role, the world model acts not only as a predictive engine but also as a supervisory layer that enables policy validation, risk assessment, safe exploration, and intervention when potentially harmful decisions are detected.

This perspective motivates uncertainty-aware world models that jointly represent network dynamics, agent interactions, policy constraints, and multi-timescale behavior. Such models can support counterfactual reasoning, what-if analysis, planning over imagined trajectories, and governance-aware decision validation. More broadly, they open the path toward predict--verify--govern--act--adapt control loops, in which candidate decisions are first evaluated in a learned predictive environment, assessed against operational objectives and safety constraints, and only then enforced in the operational network. Such capabilities provide a foundation for trustworthy, coordinated, and scalable autonomy in future AI-native \gls{ran}s.

\subsection{Evaluation and Reproducibility Are Becoming the Primary Bottlenecks} 

The ecosystem supporting AI-native \gls{ran} research has expanded considerably, including datasets, benchmarks, simulators, digital twins, and experimental platforms. However, evaluation remains fragmented across tasks, assumptions, metrics, and experimental settings. As a result, direct comparison between approaches is often difficult, and reproducing published results remains challenging. 

This observation highlights the need for standardized and reproducible methodologies for evaluating integrated intelligence in closed-loop environments. Such methodologies should assess not only network performance but also robustness, adaptability, constraint satisfaction, scalability, computational efficiency, and coordination among multiple agents and capabilities. Equally important is the establishment of benchmark suites that combine realistic network dynamics with transparent evaluation protocols, common baselines, and representative deployment scenarios. 

Taken together, these lessons point toward AI-native RANs in which distributed controllers reason about network state,
anticipate network evolution, coordinate decisions across agents and timescales, and adapt through closed-loop interaction. Achieving this vision depends on the convergence of Hybrid AI architectures, predictive world models, multi-agent coordination mechanisms, and reproducible evaluation frameworks. Such capabilities provide the foundation for network intelligence that is proactive, trustworthy, and scalable.

%% file: bare_jrnl.bbl
\begin{thebibliography}{100}
\providecommand{\url}[1]{#1}
\csname url@samestyle\endcsname
\providecommand{\newblock}{\relax}
\providecommand{\bibinfo}[2]{#2}
\providecommand{\BIBentrySTDinterwordspacing}{\spaceskip=0pt\relax}
\providecommand{\BIBentryALTinterwordstretchfactor}{4}
\providecommand{\BIBentryALTinterwordspacing}{\spaceskip=\fontdimen2\font plus
\BIBentryALTinterwordstretchfactor\fontdimen3\font minus
  \fontdimen4\font\relax}
\providecommand{\BIBforeignlanguage}[2]{{%
\expandafter\ifx\csname l@#1\endcsname\relax
\typeout{** WARNING: IEEEtran.bst: No hyphenation pattern has been}%
\typeout{** loaded for the language `#1'. Using the pattern for}%
\typeout{** the default language instead.}%
\else
\language=\csname l@#1\endcsname
\fi
#2}}
\providecommand{\BIBdecl}{\relax}
\BIBdecl

\bibitem{luong2019applications}
N.~C. Luong, D.~T. Hoang, S.~Gong, D.~Niyato, P.~Wang, Y.-C. Liang, and D.~I.
  Kim, ``Applications of deep reinforcement learning in communications and
  networking: A survey,'' \emph{IEEE Communications Surveys \& Tutorials},
  vol.~21, no.~4, pp. 3133--3174, 2019.

\bibitem{chen2020reinforcement}
Y.~Chen, Y.~Liu, M.~Zeng, U.~Saleem, Z.~Lu, X.~Wen, D.~Jin, Z.~Han, T.~Jiang,
  and Y.~Li, ``Reinforcement learning meets wireless networks: A layering
  perspective,'' \emph{IEEE Internet of Things Journal}, vol.~8, no.~1, pp.
  85--111, 2021.

\bibitem{popovski20185g}
P.~Popovski, K.~F. Trillingsgaard, O.~Simeone, and G.~Durisi, ``{5G} wireless
  network slicing for {eMBB}, {URLLC}, and {mMTC}: A communication-theoretic
  view,'' \emph{IEEE Access}, vol.~6, pp. 55\,765--55\,779, 2018.

\bibitem{parvez2018survey}
I.~Parvez, A.~Rahmati, I.~Guvenc, A.~I. Sarwat, and H.~Dai, ``A survey on low
  latency towards {5G}: {RAN}, core network and caching solutions,'' \emph{IEEE
  Communications Surveys \& Tutorials}, vol.~20, no.~4, pp. 3098--3130, 2018.

\bibitem{olwal2016survey}
T.~O. Olwal, K.~Djouani, and A.~M. Kurien, ``A survey of resource management
  toward {5G} radio access networks,'' \emph{IEEE Communications Surveys \&
  Tutorials}, vol.~18, no.~3, pp. 1656--1686, 2016.

\bibitem{oran_architecture_2026}
{O-RAN Alliance}, ``{O-RAN Architecture Description},'' European
  Telecommunications Standards Institute (ETSI), ETSI Technical Specification
  TS 103 982 V14.0.0, Mar. 2026.

\bibitem{mustafovski2026ai}
R.~Mustafovski, G.~Marinova, B.~Qehaja, E.~Hajrizi, S.~Gagica, and
  V.~Guliashki, ``{AI}-driven network optimization for the {5G}-to-{6G}
  transition: A taxonomy-based survey and reference framework,'' \emph{Future
  Internet}, vol.~18, no.~3, p. 155, 2026.

\bibitem{ahmadi20195g}
S.~Ahmadi, \emph{{5G} {NR}: Architecture, technology, implementation, and
  operation of {3GPP} new radio standards}.\hskip 1em plus 0.5em minus
  0.4em\relax Academic Press, 2019.

\bibitem{peterson20205g}
L.~Peterson and O.~Sunay, \emph{{5G} Mobile Networks: A Systems Approach}, ser.
  Synthesis Lectures on Network Systems.\hskip 1em plus 0.5em minus 0.4em\relax
  Morgan \& Claypool Publishers, 2020.

\bibitem{yan2025near}
P.~Yan, J.~Lu, H.~Zeng, and Y.~T. Hou, ``Near-real-time resource slicing for
  {QoS} optimization in {5G} {O-RAN} using deep reinforcement learning,''
  \emph{IEEE Transactions on Networking}, vol.~34, pp. 1596--1611, 2026.

\bibitem{navidan2026toward}
H.~Navidan, M.~Cheraghinia, J.~Fontaine, M.~Seif, E.~De~Poorter, H.~V. Poor,
  I.~Moerman, and A.~Shahid, ``Toward autonomous {O-RAN}: A multi-scale agentic
  {AI} framework for real-time network control and management,'' \emph{IEEE
  Network}, pp. 1--9, 2026, early Access.

\bibitem{heo2025dynamic}
D.~Heo, D.~N. Rim, and H.~Choi, ``Dynamic preference multi-objective
  reinforcement learning for internet network management,'' \emph{arXiv
  preprint arXiv:2506.13153}, 2025.

\bibitem{khetarpal2022continualrl}
K.~Khetarpal, M.~Riemer, I.~Rish, and D.~Precup, ``Towards continual
  reinforcement learning: A review and perspectives,'' \emph{J. Artificial
  Intelligence Research}, vol.~75, pp. 1401--1476, 2022.

\bibitem{korkmaz2024survey}
E.~Korkmaz, ``A survey analyzing generalization in deep reinforcement
  learning,'' \emph{arXiv preprint arXiv:2401.02349}, 2024.

\bibitem{pateria2021hierarchical}
S.~Pateria, B.~Subagdja, A.-H. Tan, and C.~Quek, ``Hierarchical reinforcement
  learning: A comprehensive survey,'' \emph{ACM Computing Surveys}, vol.~54,
  no.~5, pp. 109:1--109:35, 2022.

\bibitem{machado2023temporal}
\BIBentryALTinterwordspacing
M.~C. Machado, A.~Barreto, D.~Precup, and M.~Bowling, ``Temporal abstraction in
  reinforcement learning with the successor representation,'' \emph{Journal of
  Machine Learning Research}, vol.~24, no.~80, pp. 1--69, 2023. [Online].
  Available: \url{http://jmlr.org/papers/v24/21-1213.html}
\BIBentrySTDinterwordspacing

\bibitem{yu2025reinforcement}
Y.~Yu, ``Reinforcement learning with anticipation: A hierarchical approach for
  long-horizon tasks,'' \emph{arXiv preprint arXiv:2509.05545}, 2025.

\bibitem{huh2023multi}
D.~Huh and P.~Mohapatra, ``Multi-agent reinforcement learning: A comprehensive
  survey,'' \emph{arXiv preprint arXiv:2312.10256}, 2023.

\bibitem{ma2023hierarchical}
T.~Ma, K.~Peng, H.~Rong, Y.~Qian, and N.~Al-Nabhan, ``Hierarchical coordination
  multi-agent reinforcement learning with spatio-temporal abstraction,''
  \emph{IEEE Transactions on Emerging Topics in Computational Intelligence},
  vol.~8, no.~1, pp. 533--547, 2024.

\bibitem{sun2025survey}
J.~Sun, C.~Zheng, E.~Xie, Z.~Liu, R.~Chu, J.~Qiu, J.~Xu, M.~Ding, H.~Li,
  M.~Geng \emph{et~al.}, ``A survey of reasoning with foundation models:
  Concepts, methodologies, and outlook,'' \emph{ACM Computing Surveys},
  vol.~57, no.~11, pp. 1--43, 2025.

\bibitem{zhang2026exploring}
D.~Zhang, W.~Shi, and X.~Jia, ``Exploring scientific principles and laws of
  artificial intelligence, world model, and artificial general intelligence
  ({AGI}) in future intelligence networking: Paradigms, architectures, and
  innovations,'' \emph{IEEE Communications Surveys \& Tutorials}, vol.~28, pp.
  5456--5495, 2026.

\bibitem{yang2025decision}
N.~Yang, M.~Fan, W.~Wang, and H.~Zhang, ``Decision-making large language model
  for wireless communication: A comprehensive survey on key techniques,''
  \emph{IEEE Communications Surveys \& Tutorials}, vol.~28, pp. 3055--3088,
  2026.

\bibitem{Polese_2023_Understanding_ORAN}
M.~Polese, L.~Bonati, S.~D'Oro, S.~Basagni, and T.~Melodia, ``Understanding
  {O-RAN}: Architecture, interfaces, algorithms, security, and research
  challenges,'' \emph{IEEE Communications Surveys \& Tutorials}, vol.~25,
  no.~2, pp. 1376--1411, 2023.

\bibitem{duran2025toward}
K.~Duran, L.~V. Cakir, Y.~Yigit, K.~Huseynov, S.~R. Kusu, M.~A. Ert{\"u}rk, and
  B.~Canberk, ``Toward digital twin-as-a-service ({DTaaS}) platforms: A survey
  on architecture, design requirements, and performance metrics,'' \emph{IEEE
  Communications Surveys \& Tutorials}, vol.~28, pp. 1845--1878, 2026.

\bibitem{li2022applications}
T.~Li, K.~Zhu, N.~C. Luong, D.~Niyato, Q.~Wu, Y.~Zhang, and B.~Chen,
  ``Applications of multi-agent reinforcement learning in future internet: A
  comprehensive survey,'' \emph{IEEE Communications Surveys \& Tutorials},
  vol.~24, no.~2, pp. 1240--1279, 2022.

\bibitem{zhou2026convergence}
N.~Zhou, Y.~Yao, X.~Liang, S.~Yao, J.~Wan, C.~Yang, W.~Gui, and W.~Gao,
  ``Convergence of reinforcement learning and time-sensitive networking for
  future industrial {AI} agent communication: Fundamentals, challenges, and
  opportunities,'' \emph{IEEE Communications Surveys \& Tutorials}, vol.~28,
  pp. 6136--6173, 2026.

\bibitem{Cai_2025_Tutorial_on_LLM_RL_Wireless}
L.~Cai, W.~Fu, Y.~Huang, R.~Zhang, Y.~Liu, J.~Kang, Z.~Xiong, T.~Jiang,
  D.~Niyato, X.~Wang, S.~Mao, and X.~Shen, ``Tutorial on large language
  model-enhanced reinforcement learning for wireless networks,'' \emph{IEEE
  Communications Surveys \& Tutorials}, vol.~28, pp. 6665--6693, 2026.

\bibitem{Zhang_2026_Multi_Modal_FM_Survey}
H.~Zhang, M.~Farzanullah, M.~Ghassemi, A.~Bin~Sediq, A.~Afana, and
  M.~Erol-Kantarci, ``Multi-modal data-enhanced foundation models for
  prediction and control in wireless networks: A survey,'' \emph{IEEE
  Communications Surveys \& Tutorials}, vol.~28, pp. 4359--4393, 2026.

\bibitem{Mienye_2026_DRL_FM_Survey}
I.~D. Mienye, E.~Esenogho, and C.~Modisane, ``Deep reinforcement learning in
  the era of foundation models: A survey,'' \emph{Computers}, vol.~15, no.~1,
  p.~40, 2026.

\bibitem{Wei_2025_LLM_For_Wireless_Management}
B.~Wei, R.~Jiang, R.~Zhang, Y.~Liu, D.~Niyato, Y.~Sun, Y.~Lu, Y.~Li, S.~Mao,
  C.~Yuen, M.~Di~Renzo, and M.~Peng, ``Large language models for optimization
  in next-generation wireless network management: A survey,'' \emph{IEEE
  Communications Surveys \& Tutorials}, 2026, early access.

\bibitem{zhao2025edge}
C.~Zhao, G.~Liu, R.~Zhang, Y.~Liu, J.~Wang, J.~Kang, D.~Niyato, Z.~Li, X.~Shen,
  Z.~Han, S.~Sun, C.~Yuen, and D.~I. Kim, ``Edge general intelligence through
  world models, large language models, and agentic {AI}: Fundamentals,
  solutions, and challenges,'' \emph{IEEE Transactions on Cognitive
  Communications and Networking}, vol.~12, pp. 5649--5675, 2026.

\bibitem{liu2024survey}
Y.-F. Liu, T.-H. Chang, M.~Hong, Z.~Wu, A.~M.-C. So, E.~A. Jorswieck, and
  W.~Yu, ``A survey of recent advances in optimization methods for wireless
  communications,'' \emph{IEEE Journal on Selected Areas in Communications},
  vol.~42, no.~11, pp. 2992--3031, 2024.

\bibitem{barker2026ainative}
\BIBentryALTinterwordspacing
R.~Barker, A.~Ebrahimi~Dorcheh, T.~Seyfi, M.~R. Uddin, A.~Mohammadhosseini,
  J.~Boone, S.~Streit, D.~Schlesener, and F.~Afghah, ``{AI}-native {Open RAN}:
  A roadmap from {xApps} and {rApps} to autonomous network agents,'' 2026.
  [Online]. Available: \url{https://arxiv.org/abs/2609.15704}
\BIBentrySTDinterwordspacing

\bibitem{ameur2026agentic}
\BIBentryALTinterwordspacing
M.~Ameur, A.~Mekrache, B.~Brik, and A.~Ksentini, ``{LLM}-powered agentic {AI}
  for {5G}/{6G} networks: A tutorial and survey on architectures, protocols,
  and standardization,'' 2026. [Online]. Available:
  \url{https://arxiv.org/abs/2607.16066}
\BIBentrySTDinterwordspacing

\bibitem{belter2026assured}
\BIBentryALTinterwordspacing
B.~Belter and M.~Glabowski, ``Assured {AI}-native network control loops: State
  of the art, research challenges and the missing runtime assurance layer,''
  2026. [Online]. Available: \url{https://arxiv.org/abs/2609.11996}
\BIBentrySTDinterwordspacing

\bibitem{sahin2026ainos}
\BIBentryALTinterwordspacing
O.~Sahin, V.~Sathya, and L.~Zhang, ``A survey and tutorial on {AI}-native
  network operating systems for {6G} communication networks: Architectures,
  control loops, memory, and governed autonomy,'' SSRN preprint 7174399, 2026.
  [Online]. Available:
  \url{https://papers.ssrn.com/sol3/papers.cfm?abstract_id=7174399}
\BIBentrySTDinterwordspacing

\bibitem{11320975}
X.~Lin, L.~Kundu, S.~Cammerer, Y.~Huang, C.~Dick, C.~Santhosam, R.~Gadiyar,
  R.~Wiesmayr, and C.~Studer, ``Ai-native 6g: Empowering intelligent ran with
  accelerated compute,'' \emph{IEEE Wireless Communications}, 2025.

\bibitem{bertsekas2019reinforcement}
D.~P. Bertsekas, \emph{Reinforcement Learning and Optimal Control}.\hskip 1em
  plus 0.5em minus 0.4em\relax Belmont, MA: Athena Scientific, 2019.

\bibitem{peng2015recent}
M.~Peng, C.~Wang, J.~Li, H.~Xiang, and V.~Lau, ``Recent advances in underlay
  heterogeneous networks: Interference control, resource allocation, and
  self-organization,'' \emph{IEEE Communications Surveys \& Tutorials},
  vol.~17, no.~2, pp. 700--729, 2015.

\bibitem{sutton2018reinforcement}
R.~S. Sutton and A.~G. Barto, \emph{Reinforcement Learning: An Introduction},
  2nd~ed.\hskip 1em plus 0.5em minus 0.4em\relax MIT Press, 2018.

\bibitem{arulkumaran2017deep}
K.~Arulkumaran, M.~P. Deisenroth, M.~Brundage, and A.~A. Bharath, ``Deep
  reinforcement learning: A brief survey,'' \emph{IEEE Signal Processing
  Magazine}, vol.~34, no.~6, pp. 26--38, 2017.

\bibitem{alizadeh2025offline}
P.~Alizadeh, A.~Giovanidis, P.~Ramachandra, V.~Koutsoukis, and O.~Arouk,
  ``Offline reinforcement learning for mobility robustness optimization,''
  \emph{arXiv preprint arXiv:2506.22793}, 2025.

\bibitem{guo2023reinforcement}
C.~Guo, Z.~Li, L.~Liang, and G.~Y. Li, ``Reinforcement learning-based power
  control for reliable mission-critical wireless transmission,'' \emph{IEEE
  Internet of Things Journal}, vol.~10, no.~23, pp. 20\,868--20\,883, 2023.

\bibitem{ha2018worldmodels}
D.~Ha and J.~Schmidhuber, ``World models,'' \emph{arXiv preprint
  arXiv:1803.10122}, 2018.

\bibitem{hafner2020mastering}
D.~Hafner, T.~Lillicrap, M.~Norouzi, and J.~Ba, ``Mastering atari with discrete
  world models,'' in \emph{International Conference on Learning Representations
  (ICLR)}, 2021, arXiv:2010.02193.

\bibitem{hafner2023mastering}
D.~Hafner, J.~Pasukonis, J.~Ba, and T.~Lillicrap, ``Mastering diverse domains
  through world models,'' \emph{arXiv preprint arXiv:2301.04104}, 2023,
  published as ``Mastering diverse control tasks through world models'', Nature
  640:647--653, 2025, doi:10.1038/s41586-025-08744-2.

\bibitem{Moerland_2023_Model_based_RL}
T.~M. Moerland, J.~Broekens, A.~Plaat, and C.~M. Jonker, ``Model-based
  reinforcement learning: A survey,'' \emph{Foundations and Trends in Machine
  Learning}, vol.~16, no.~1, pp. 1--118, 2023.

\bibitem{bouton2026mobile}
M.~Bouton, I.~Mitsioni, S.~Lindst{\aa}hl, and J.~Jeong, ``Mobile network
  control with a world model,'' in \emph{2026 IEEE/IFIP Network Operations and
  Management Symposium (NOMS)}.\hskip 1em plus 0.5em minus 0.4em\relax IEEE,
  2026, pp. 1--11.

\bibitem{zhou2025world}
H.~Zhou, S.~Lin, W.~Wang, D.~Gao, and J.~Zhang, ``World-model-based adaptive
  network slicing,'' in \emph{GLOBECOM 2025 - 2025 IEEE Global Communications
  Conference}.\hskip 1em plus 0.5em minus 0.4em\relax IEEE, 2025, pp.
  1627--1632.

\bibitem{bommasani2021opportunities}
R.~Bommasani, D.~A. Hudson, E.~Adeli, R.~Altman, S.~Arora, S.~von Arx, M.~S.
  Bernstein, J.~Bohg, A.~Bosselut, E.~Brunskill \emph{et~al.}, ``On the
  opportunities and risks of foundation models,'' \emph{arXiv preprint
  arXiv:2108.07258}, 2021.

\bibitem{Habib2025LlmBasedIntentProcessing}
M.~A. Habib, P.~E. Iturria~Rivera, Y.~Ozcan, M.~Elsayed, M.~Bavand,
  R.~Gaigalas, and M.~Erol-Kantarci, ``{LLM}-based intent processing and
  network optimization using attention-based hierarchical reinforcement
  learning,'' in \emph{2025 IEEE Wireless Communications and Networking
  Conference (WCNC)}, 2025, pp. 1--6.

\bibitem{lotfi2025prompt}
F.~Lotfi, H.~Rajoli, and F.~Afghah, ``Prompt-tuned {LLM}-augmented {DRL} for
  dynamic {O-RAN} network slicing,'' 2025, published version: ``Scalable
  LLM-Augmented DRL with Context-Aware Prompt Learning for O-RAN Slicing,''
  Proc. IEEE ICASSP 2026, pp. 21436--21440,
  doi:10.1109/ICASSP55912.2026.11461553.

\bibitem{reed2022gato}
\BIBentryALTinterwordspacing
S.~Reed, K.~{\.Z}o{\l}na, E.~Parisotto, S.~G{\'o}mez~Colmenarejo, A.~Novikov
  \emph{et~al.}, ``A generalist agent,'' \emph{Transactions on Machine Learning
  Research}, 2022. [Online]. Available:
  \url{https://openreview.net/forum?id=1ikK0kHjvj}
\BIBentrySTDinterwordspacing

\bibitem{driess2023palme}
\BIBentryALTinterwordspacing
D.~Driess, F.~Xia, M.~S.~M. Sajjadi, C.~Lynch, A.~Chowdhery, B.~Ichter,
  A.~Wahid, J.~Tompson, Q.~Vuong, T.~Yu, W.~Huang, Y.~Chebotar, P.~Sermanet,
  D.~Duckworth, S.~Levine, V.~Vanhoucke, K.~Hausman, M.~Toussaint, K.~Greff,
  A.~Zeng, I.~Mordatch, and P.~Florence, ``{PaLM-E}: An embodied multimodal
  language model,'' in \emph{Proceedings of the 40th International Conference
  on Machine Learning (ICML)}, ser. Proceedings of Machine Learning Research,
  vol. 202.\hskip 1em plus 0.5em minus 0.4em\relax PMLR, 2023, pp. 8469--8488.
  [Online]. Available: \url{https://proceedings.mlr.press/v202/driess23a.html}
\BIBentrySTDinterwordspacing

\bibitem{chen2021decisiontransformer}
L.~Chen, K.~Lu, A.~Rajeswaran, K.~Lee, A.~Grover, M.~Laskin, P.~Abbeel,
  A.~Srinivas, and I.~Mordatch, ``Decision transformer: Reinforcement learning
  via sequence modeling,'' in \emph{Advances in Neural Information Processing
  Systems}, vol.~34, 2021, pp. 15\,084--15\,097.

\bibitem{micheli2023transformers}
V.~Micheli, E.~Alonso, and F.~Fleuret, ``Transformers are sample-efficient
  world models,'' in \emph{International Conference on Learning Representations
  (ICLR)}, 2023.

\bibitem{qiu2024large}
K.~Qiu, S.~Bakirtzis, I.~Wassell, H.~Song, J.~Zhang, and K.~Wang, ``Large
  language model-based wireless network design,'' \emph{IEEE Wireless
  Communications Letters}, vol.~13, no.~12, pp. 3340--3344, 2024.

\bibitem{chai2025MobiWorld}
\BIBentryALTinterwordspacing
H.~Chai, Y.~Yuan, and Y.~Li, ``{MobiWorld}: World models for mobile wireless
  network,'' arXiv:2507.09462, 2025. [Online]. Available:
  \url{https://arxiv.org/abs/2507.09462}
\BIBentrySTDinterwordspacing

\bibitem{amos2017optnet}
B.~Amos and J.~Z. Kolter, ``{OptNet}: Differentiable optimization as a layer in
  neural networks,'' in \emph{Proceedings of the 34th International Conference
  on Machine Learning (ICML)}, ser. Proceedings of Machine Learning Research,
  vol.~70.\hskip 1em plus 0.5em minus 0.4em\relax PMLR, 2017, pp. 136--145.

\bibitem{kotary2021endtoend}
J.~Kotary, F.~Fioretto, P.~Van~Hentenryck, and B.~Wilder, ``End-to-end
  constrained optimization learning: A survey,'' in \emph{Proceedings of the
  Thirtieth International Joint Conference on Artificial Intelligence
  (IJCAI-21)}, 2021, pp. 4475--4482.

\bibitem{sudhakara2025constrained}
S.~Sudhakara and P.~Rajak, ``Constrained network slice assignment via large
  language models,'' \emph{arXiv preprint arXiv:2512.00040}, 2025, neurIPS 2025
  Workshop AI4NextG.

\bibitem{tan2025tool}
C.~Tan, R.~Wen, R.~Li, Z.~Zhao, E.~Hossain, and H.~Zhang, ``Tool-aided
  evolutionary {LLM} for generative policy toward efficient resource management
  in wireless federated learning,'' \emph{IEEE Journal on Selected Areas in
  Communications}, vol.~44, pp. 2904--2921, 2026.

\bibitem{chua2018deep}
K.~Chua, R.~Calandra, R.~McAllister, and S.~Levine, ``Deep reinforcement
  learning in a handful of trials using probabilistic dynamics models,'' in
  \emph{Advances in Neural Information Processing Systems}, vol.~31, 2018, pp.
  4754--4765.

\bibitem{OnlineFS}
A.~Giovanidis, M.~Leconte, S.~Aroua, T.~Kvernvik, and D.~Sandberg, ``Online
  frequency scheduling by learning parallel actions,'' in \emph{2024 3rd
  International Conference on 6G Networking (6GNet)}, 2024, pp. 153--160.

\bibitem{wu2006tilting}
J.~Wu, J.~Bigham, P.~Jiang, and J.~P. Neophytou, ``Tilting and beam-shaping for
  traffic load balancing in {WCDMA} network,'' in \emph{2006 European
  Conference on Wireless Technologies}.\hskip 1em plus 0.5em minus 0.4em\relax
  IEEE, 2006, pp. 63--66.

\bibitem{forsberg2023network}
A.~Larsson~Forsberg, A.~Nikou, A.~V. Feljan, and J.~Tumova, ``Network parameter
  control in cellular networks through graph-based multi-agent constrained
  reinforcement learning,'' in \emph{2023 IEEE 19th International Conference on
  Automation Science and Engineering (CASE)}.\hskip 1em plus 0.5em minus
  0.4em\relax IEEE, 2023, pp. 1--7.

\bibitem{xu2025fully}
Y.~Xu, Z.~Liu, B.~Qian, H.~Du, J.~Chen, J.~Kang, H.~Zhou, and D.~Niyato,
  ``Fully-decoupled {RAN} for feedback-free multi-base station transmission in
  {MIMO-OFDM} system,'' \emph{IEEE Journal on Selected Areas in
  Communications}, vol.~43, no.~3, pp. 780--794, 2025.

\bibitem{lacava2023programmable}
A.~Lacava, M.~Polese, R.~Sivaraj, R.~Soundrarajan, B.~S. Bhati, T.~Singh,
  T.~Zugno, F.~Cuomo, and T.~Melodia, ``Programmable and customized
  intelligence for traffic steering in {5G} networks using open {RAN}
  architectures,'' \emph{IEEE Transactions on Mobile Computing}, vol.~23,
  no.~4, pp. 2882--2897, 2024.

\bibitem{o_ran_whitepaper_2020}
\BIBentryALTinterwordspacing
{O-RAN Alliance}, ``{O-RAN Use Cases and Deployment Scenarios: Towards Open and
  Smart RAN},'' White Paper, Feb. 2020. [Online]. Available:
  \url{https://mediastorage.o-ran.org/white-papers/O-RAN.WG1.Use-Cases-and-Deployment-Scenarios-White-Paper-2020-02.pdf}
\BIBentrySTDinterwordspacing

\bibitem{fourati2021comprehensive}
H.~Fourati, R.~Maaloul, L.~Chaari, and M.~Jmaiel, ``Comprehensive survey on
  self-organizing cellular network approaches applied to {5G} networks,''
  \emph{Computer Networks}, vol. 199, p. 108435, 2021.

\bibitem{3gpp38213}
{3rd Generation Partnership Project (3GPP)}, ``{NR; Physical layer procedures
  for control},'' 3GPP, Technical Specification TS 38.213 V19.1.0, Sep. 2025,
  release 19.

\bibitem{3gpp38214}
------, ``{NR; Physical layer procedures for data},'' 3GPP, Technical
  Specification TS 38.214 V19.5.0, Sep. 2026, release 19.

\bibitem{3gpp38321}
\BIBentryALTinterwordspacing
------, ``{NR; Medium Access Control (MAC) protocol specification},'' 3GPP,
  Technical Specification TS 38.321 V19.0.0, Sep. 2025, release 19. [Online].
  Available: \url{https://www.3gpp.org/ftp/Specs/archive/38_series/38.321/}
\BIBentrySTDinterwordspacing

\bibitem{mismar2019deep}
F.~B. Mismar, B.~L. Evans, and A.~Alkhateeb, ``Deep reinforcement learning for
  {5G} networks: Joint beamforming, power control, and interference
  coordination,'' \emph{IEEE Transactions on Communications}, vol.~68, no.~3,
  pp. 1581--1592, 2020.

\bibitem{wang2018handover}
Z.~Wang, L.~Li, Y.~Xu, H.~Tian, and S.~Cui, ``Handover control in wireless
  systems via asynchronous multiuser deep reinforcement learning,'' \emph{IEEE
  Internet of Things Journal}, vol.~5, no.~6, pp. 4296--4307, 2018.

\bibitem{addali2019dynamic}
K.~M. Addali, S.~Y. {Bani Melhem}, Y.~Khamayseh, Z.~Zhang, and M.~Kadoch,
  ``Dynamic mobility load balancing for {5G} small-cell networks based on
  utility functions,'' \emph{IEEE Access}, vol.~7, pp. 126\,998--127\,011,
  2019.

\bibitem{rodriguez20205g}
V.~Q. Rodriguez, F.~Guillemin, and A.~Boubendir, ``{5G} {E2E} network slicing
  management with {ONAP},'' in \emph{2020 23rd Conference on Innovation in
  Clouds, Internet and Networks and Workshops (ICIN)}.\hskip 1em plus 0.5em
  minus 0.4em\relax IEEE, 2020, pp. 87--94.

\bibitem{hoffmann2024energy}
M.~Hoffmann and M.~Dryja{\'n}ski, ``Energy efficiency in open {RAN}: {RF}
  channel reconfiguration use case,'' \emph{IEEE Access}, vol.~12, pp.
  118\,493--118\,501, 2024.

\bibitem{aroua2025adaptive}
\BIBentryALTinterwordspacing
S.~Aroua, C.~A. Bovolis, B.~Göransson, A.~Giovanidis, M.~Leconte, and
  A.~Destounis, ``Adaptive design of mmwave initial access codebooks using
  reinforcement learning,'' in \emph{2026 IEEE 103rd Vehicular Technology
  Conference (VTC2026-Spring)}, 2026. [Online]. Available:
  \url{https://arxiv.org/abs/2510.25271}
\BIBentrySTDinterwordspacing

\bibitem{schrittwieser2020muzero}
J.~Schrittwieser, I.~Antonoglou, T.~Hubert, K.~Simonyan, L.~Sifre, S.~Schmitt,
  A.~Guez, E.~Lockhart, D.~Hassabis, T.~Graepel, T.~Lillicrap, and D.~Silver,
  ``Mastering {Atari}, {Go}, chess and shogi by planning with a learned
  model,'' \emph{Nature}, vol. 588, no. 7839, pp. 604--609, 2020.

\bibitem{wang2025founder}
Y.~Wang, R.~Yu, S.~Wan, L.~Gan, and D.-C. Zhan, ``{FOUNDER}: Grounding
  foundation models in world models for open-ended embodied decision making,''
  in \emph{Proceedings of the 42nd International Conference on Machine Learning
  (ICML)}, 2025.

\bibitem{yang2023foundationdecision}
S.~Yang, O.~Nachum, Y.~Du, J.~Wei, P.~Abbeel, and D.~Schuurmans, ``Foundation
  models for decision making: Problems, methods, and opportunities,''
  \emph{arXiv preprint arXiv:2303.04129}, 2023.

\bibitem{elmachtoub2022smart}
A.~N. Elmachtoub and P.~Grigas, ``Smart ``predict, then optimize'',''
  \emph{Management Science}, vol.~68, no.~1, pp. 9--26, 2022.

\bibitem{marasinghe2026perception}
D.~Marasinghe, S.~Gimhana, N.~Rajapaksha, P.~Susarla, N.~Vaara, T.~Sivalingam,
  and N.~Rajatheva, ``Perception-aided ai-ran for industrial physical ai
  deployments,'' in \emph{2026 International Conference on Future and
  Intelligent Networking (FINE)}.\hskip 1em plus 0.5em minus 0.4em\relax IEEE,
  2026, pp. 89--96.

\bibitem{battaglia2018relational}
P.~W. Battaglia, J.~B. Hamrick, V.~Bapst, A.~Sanchez-Gonzalez, V.~Zambaldi,
  M.~Malinowski, A.~Tacchetti, D.~Raposo, A.~Santoro, R.~Faulkner
  \emph{et~al.}, ``Relational inductive biases, deep learning, and graph
  networks,'' \emph{arXiv preprint arXiv:1806.01261}, 2018.

\bibitem{kaelbling1998planning}
L.~P. Kaelbling, M.~L. Littman, and A.~R. Cassandra, ``Planning and acting in
  partially observable stochastic domains,'' \emph{Artificial Intelligence},
  vol. 101, no. 1-2, pp. 99--134, 1998.

\bibitem{samejima1999adaptive}
K.~Samejima and T.~Omori, ``Adaptive internal state space construction method
  for reinforcement learning of a real-world agent,'' \emph{Neural Networks},
  vol.~12, no. 7-8, pp. 1143--1155, 1999.

\bibitem{wierstra2010recurrent}
D.~Wierstra, A.~F{\"o}rster, J.~Peters, and J.~Schmidhuber, ``Recurrent policy
  gradients,'' \emph{Logic Journal of the IGPL}, vol.~18, no.~5, pp. 620--634,
  2010.

\bibitem{baisero2018learning}
A.~Baisero and C.~Amato, ``Learning internal state models in partially
  observable environments,'' in \emph{Reinforcement Learning under Partial
  Observability, NeurIPS Workshop}, 2018.

\bibitem{igl2018deep}
M.~Igl, L.~Zintgraf, T.~A. Le, F.~Wood, and S.~Whiteson, ``Deep variational
  reinforcement learning for {POMDPs},'' in \emph{Proceedings of the 35th
  International Conference on Machine Learning}, ser. Proceedings of Machine
  Learning Research, vol.~80.\hskip 1em plus 0.5em minus 0.4em\relax PMLR,
  2018, pp. 2117--2126.

\bibitem{madbhavi2023graph}
R.~Madbhavi, B.~Natarajan, and B.~Srinivasan, ``Graph neural network-based
  distribution system state estimators,'' \emph{IEEE Transactions on Industrial
  Informatics}, vol.~19, no.~12, pp. 11\,630--11\,639, 2023.

\bibitem{jin2112graph}
Y.~Jin, F.~Vannella, M.~Bouton, J.~Jeong, and E.~{Al Hakim}, ``A graph
  attention learning approach to antenna tilt optimization,'' in \emph{2022 1st
  International Conference on 6G Networking (6GNet)}.\hskip 1em plus 0.5em
  minus 0.4em\relax IEEE, 2022, pp. 1--5.

\bibitem{zhang2024neuro}
X.~Zhang and V.~S. Sheng, ``Neuro-symbolic {AI}: Explainability, challenges,
  and future trends,'' \emph{arXiv preprint arXiv:2411.04383}, 2024.

\bibitem{zou2025telecomgpt}
H.~Zou, Q.~Zhao, Y.~Tian, L.~Bariah, F.~Bader, T.~Lestable, and M.~Debbah,
  ``{TelecomGPT}: A framework to build telecom-specific large language
  models,'' \emph{IEEE Transactions on Machine Learning in Communications and
  Networking}, vol.~3, pp. 948--975, 2025.

\bibitem{alikhani2024large}
S.~Xu, C.~Kurisummoottil~Thomas, O.~Hashash, N.~Muralidhar, W.~Saad, and
  N.~Ramakrishnan, ``Large multi-modal models ({LMMs}) as universal foundation
  models for {AI}-native wireless systems,'' \emph{IEEE Network}, vol.~38,
  no.~5, pp. 10--20, 2024.

\bibitem{jiao2025ai2mmum}
T.~Jiao, Y.~Xu, Z.~Xiao, Y.~Huang, C.~Ye, Y.~Feng, L.~Cai, J.~Chang, F.~Liu,
  D.~He, Y.~Guan, and W.~Zhang, ``{AI$^2$MMUM}: {AI}-{AI} oriented multi-modal
  universal model leveraging telecom domain large model,'' \emph{IEEE Wireless
  Communications Letters}, vol.~14, no.~8, pp. 2651--2655, 2025.

\bibitem{gao2023retrieval}
Y.~Gao, Y.~Xiong, X.~Gao, K.~Jia, J.~Pan, Y.~Bi, Y.~Dai, J.~Sun, M.~Wang, and
  H.~Wang, ``Retrieval-augmented generation for large language models: A
  survey,'' 2023.

\bibitem{brown2020language}
T.~Brown, B.~Mann, N.~Ryder, M.~Subbiah, J.~Kaplan, P.~Dhariwal,
  A.~Neelakantan, P.~Shyam, G.~Sastry, A.~Askell \emph{et~al.}, ``Language
  models are few-shot learners,'' in \emph{Advances in Neural Information
  Processing Systems}, vol.~33, 2020, pp. 1877--1901.

\bibitem{sahoo2024systematic}
P.~Sahoo, A.~K. Singh, S.~Saha, V.~Jain, S.~Mondal, and A.~Chadha, ``A
  systematic survey of prompt engineering in large language models: Techniques
  and applications,'' \emph{arXiv preprint arXiv:2402.07927}, 2024.

\bibitem{nardini2024enabling}
G.~Nardini and G.~Stea, ``Enabling simulation services for digital twins of
  {5G}/{B5G} mobile networks,'' \emph{Computer Communications}, vol. 213, pp.
  33--48, 2024.

\bibitem{adday2024investigating}
G.~H. Adday, S.~K. Subramaniam, Z.~A. Zukarnain, and N.~Samian, ``Investigating
  and analyzing simulation tools of wireless sensor networks: A comprehensive
  survey,'' \emph{IEEE Access}, vol.~12, pp. 22\,938--22\,977, 2024.

\bibitem{tran2025network}
D.-H. Tran, N.~Waheed, Y.~M. Saputra, X.~Lin, C.~T. Nguyen, T.~S. Abdu, V.~N.
  Vo, V.-Q. Pham, M.~Alsenwi, A.~B.~M. Adam, S.~Chatzinotas, E.~Lagunas,
  H.~Tran, T.~H. Dac, and N.~V. Huynh, ``Network digital twin for {6G} and
  beyond: An end-to-end view across multi-domain network ecosystems,''
  \emph{IEEE Open Journal of the Communications Society}, vol.~6, pp.
  6866--6911, 2025.

\bibitem{zaki2026reference}
A.~Zaki-Hindi, P.~Soto, G.~Castellanos, T.~H. Pritom, G.~Volpe, I.~Zellagui,
  B.~Hensel, J.~Jimenez, M.~Marvulli, L.~Santos \emph{et~al.}, ``A reference
  functional architecture for network digital twins in {6G} systems,''
  \emph{IEEE Open Journal of the Communications Society}, vol.~7, pp.
  2068--2101, 2026.

\bibitem{young2000stochastic}
P.~C. Young, ``Stochastic, dynamic modelling and signal processing: Time
  variable and state dependent parameter estimation,'' in \emph{Nonlinear and
  Nonstationary Signal Processing}, W.~J. Fitzgerald, R.~L. Smith, A.~T.
  Walden, and P.~C. Young, Eds.\hskip 1em plus 0.5em minus 0.4em\relax
  Cambridge: Cambridge University Press, 2000, pp. 74--114.

\bibitem{rashid2009opportunistic}
M.~M. Rashid, M.~J. Hossain, E.~Hossain, and V.~K. Bhargava, ``Opportunistic
  spectrum scheduling for multiuser cognitive radio: a queueing analysis,''
  \emph{IEEE Transactions on Wireless Communications}, vol.~8, no.~10, pp.
  5259--5269, 2009.

\bibitem{cui2019traffic}
Z.~Cui, K.~Henrickson, R.~Ke, and Y.~Wang, ``Traffic graph convolutional
  recurrent neural network: A deep learning framework for network-scale traffic
  learning and forecasting,'' \emph{IEEE Transactions on Intelligent
  Transportation Systems}, vol.~21, no.~11, pp. 4883--4894, 2020.

\bibitem{habib2024transformer}
M.~A. Habib, P.~E. Iturria-Rivera, Y.~Ozcan, M.~Elsayed, M.~Bavand,
  R.~Gaigalas, and M.~Erol-Kantarci, ``Transformer-based wireless traffic
  prediction and network optimization in {O-RAN},'' in \emph{2024 IEEE
  International Conference on Communications Workshops (ICC Workshops)}.\hskip
  1em plus 0.5em minus 0.4em\relax IEEE, 2024, pp. 1--6.

\bibitem{coronado2022zero}
E.~Coronado, R.~Behravesh, T.~Subramanya, A.~Fern{\'a}ndez-Fern{\'a}ndez, M.~S.
  Siddiqui, X.~Costa-P{\'e}rez, and R.~Riggio, ``Zero touch management: A
  survey of network automation solutions for {5G} and {6G} networks,''
  \emph{IEEE Communications Surveys \& Tutorials}, vol.~24, no.~4, pp.
  2535--2578, 2022.

\bibitem{annaswamy2023adaptive}
A.~M. Annaswamy, ``Adaptive control and intersections with reinforcement
  learning,'' \emph{Annual Review of Control, Robotics, and Autonomous
  Systems}, vol.~6, no.~1, pp. 65--93, 2023.

\bibitem{stooke2020responsive}
A.~Stooke, J.~Achiam, and P.~Abbeel, ``Responsive safety in reinforcement
  learning by {PID} {Lagrangian} methods,'' in \emph{Proceedings of the 37th
  International Conference on Machine Learning (ICML)}, ser. Proceedings of
  Machine Learning Research, vol. 119.\hskip 1em plus 0.5em minus 0.4em\relax
  PMLR, 2020, pp. 9133--9143.

\bibitem{aastrom1995adaptive}
K.~J. {\AA}str{\"o}m, ``Adaptive control,'' in \emph{Mathematical System
  Theory: The Influence of R. E. Kalman}, A.~C. Antoulas, Ed.\hskip 1em plus
  0.5em minus 0.4em\relax Berlin, Heidelberg: Springer, 1991, pp. 437--450.

\bibitem{chen2017online}
T.~Chen, Q.~Ling, and G.~B. Giannakis, ``An online convex optimization approach
  to proactive network resource allocation,'' \emph{IEEE Transactions on Signal
  Processing}, vol.~65, no.~24, pp. 6350--6364, 2017.

\bibitem{mayne2000constrained}
D.~Q. Mayne, J.~B. Rawlings, C.~V. Rao, and P.~O.~M. Scokaert, ``Constrained
  model predictive control: Stability and optimality,'' \emph{Automatica},
  vol.~36, no.~6, pp. 789--814, 2000.

\bibitem{masero2020coalitional}
E.~Masero, L.~A. Fletscher, and J.~M. Maestre, ``A coalitional model predictive
  control for the energy efficiency of next-generation cellular networks,''
  \emph{Energies}, vol.~13, no.~24, p. 6546, 2020.

\bibitem{zhao2024meta}
B.~Zhao, J.~Wu, Y.~Ma, and C.~Yang, ``Meta-learning for wireless
  communications: A survey and a comparison to {GNNs},'' \emph{IEEE Open
  Journal of the Communications Society}, vol.~5, pp. 1987--2015, 2024.

\bibitem{dahrouj2021overview}
H.~Dahrouj, R.~Alghamdi, H.~Alwazani, S.~Bahanshal, A.~A. Ahmad, A.~Faisal,
  R.~Shalabi, R.~Alhadrami, A.~Subasi, M.~T. Al-Nory, O.~Kittaneh, and J.~S.
  Shamma, ``An overview of machine learning-based techniques for solving
  optimization problems in communications and signal processing,'' \emph{IEEE
  Access}, vol.~9, pp. 74\,908--74\,938, 2021.

\bibitem{giannopoulos2025comix}
A.~E. Giannopoulos, S.~T. Spantideas, G.~Levis, A.~S. Kalafatelis, and
  P.~Trakadas, ``{COMIX}: Generalized conflict management in {O-RAN}
  {xApps}---architecture, workflow, and a power control case,'' \emph{IEEE
  Access}, vol.~13, pp. 116\,684--116\,700, 2025.

\bibitem{singh2026regulations}
S.~K. Singh, ``Regulations and standards in wireless communication,'' in
  \emph{Advanced Wireless Communication Systems: A Comprehensive Guide}.\hskip
  1em plus 0.5em minus 0.4em\relax Bentham Science Publishers, 2026, pp.
  30--81.

\bibitem{11771692}
S.~Aroua and L.~Giupponi, ``Learning to coexist: Safe reinforcement learning
  for terrestrial-incumbent spectrum sharing in the upper 6 ghz band,'' in
  \emph{2026 IEEE Conference on Standards for Communications and Networking
  (CSCN)}, 2026.

\bibitem{Adamczyk2023ConflictMitigationORAN}
\BIBentryALTinterwordspacing
C.~Adamczyk and A.~Kliks, ``Conflict mitigation framework and conflict
  detection in o-ran near-rt ric,'' \emph{IEEE Communications Magazine},
  vol.~61, no.~12, pp. 199--205, Dec. 2023. [Online]. Available:
  \url{http://dx.doi.org/10.1109/MCOM.018.2200752}
\BIBentrySTDinterwordspacing

\bibitem{polese2026beyond}
M.~Polese, N.~Mohamadi, S.~D'Oro, L.~Bonati, and T.~Melodia, ``Beyond
  connectivity: An open architecture for {AI-RAN} convergence in {6G},''
  \emph{IEEE Communications Magazine}, vol.~64, no.~7, pp. 90--95, 2026.

\bibitem{boyd2004convex}
S.~Boyd and L.~Vandenberghe, \emph{Convex Optimization}.\hskip 1em plus 0.5em
  minus 0.4em\relax Cambridge, UK: Cambridge University Press, 2004.

\bibitem{dalal2018safelayer}
\BIBentryALTinterwordspacing
G.~Dalal, K.~Dvijotham, M.~Vecerik, T.~Hester, C.~Paduraru, and Y.~Tassa,
  ``Safe exploration in continuous action spaces,'' 2018. [Online]. Available:
  \url{https://arxiv.org/abs/1801.08757}
\BIBentrySTDinterwordspacing

\bibitem{alshiekh2018shielding}
M.~Alshiekh, R.~Bloem, R.~Ehlers, B.~K{\"o}nighofer, S.~Niekum, and U.~Topcu,
  ``Safe reinforcement learning via shielding,'' in \emph{Proceedings of the
  32nd AAAI Conference on Artificial Intelligence (AAAI)}, vol.~32, no.~1,
  2018.

\bibitem{networkPlanning}
W.~El-Beaino, A.~M. El-Hajj, and Z.~Dawy, ``On radio network planning for next
  generation {5G} networks: A case study,'' in \emph{2015 International
  Conference on Communications, Signal Processing, and their Applications
  (ICCSPA'15)}, 2015, pp. 1--6.

\bibitem{gures2020comprehensive}
E.~Gures, I.~Shayea, A.~Alhammadi, M.~Ergen, and H.~Mohamad, ``A comprehensive
  survey on mobility management in {5G} heterogeneous networks: Architectures,
  challenges and solutions,'' \emph{IEEE Access}, vol.~8, pp.
  195\,883--195\,913, 2020.

\bibitem{haque2023survey}
M.~E. Haque, F.~Tariq, M.~R.~A. Khandaker, K.-K. Wong, and Y.~Zhang, ``A survey
  of scheduling in {5G} {URLLC} and outlook for emerging {6G} systems,''
  \emph{IEEE Access}, vol.~11, pp. 34\,372--34\,396, 2023.

\bibitem{mekrache2024llm}
A.~Mekrache and A.~Ksentini, ``{LLM}-enabled intent-driven service
  configuration for next generation networks,'' in \emph{2024 IEEE 10th
  International Conference on Network Softwarization (NetSoft)}.\hskip 1em plus
  0.5em minus 0.4em\relax IEEE, 2024, pp. 253--257.

\bibitem{shokrnezhad2025autonomous}
M.~Shokrnezhad and T.~Taleb, ``An autonomous network orchestration framework
  integrating large language models with continual reinforcement learning,''
  \emph{IEEE Communications Magazine}, vol.~63, no.~8, pp. 78--84, 2025.

\bibitem{Guo2025IAN}
L.~Guo, L.~Zhang, J.~Wang, J.~Wu, Y.~Yan, H.~Sun, B.~He, Q.~Qi, and J.~Liao,
  ``Intent-based autonomous network framework guided by large language model,''
  \emph{IEEE Transactions on Automation Science and Engineering}, vol.~22, pp.
  22\,185--22\,197, 2025.

\bibitem{Liu2026LAMeTA}
Y.~Liu, G.~Liu, J.~Wang, R.~Zhang, D.~Niyato, G.~Sun, Z.~Xiong, and Z.~Han,
  ``{LAMeTA}: Intent-aware agentic network optimization via a large {AI}
  model-empowered two-stage approach,'' \emph{IEEE Journal on Selected Areas in
  Communications}, vol.~44, pp. 2462--2478, 2026.

\bibitem{mekrache2024intent}
A.~Mekrache, A.~Ksentini, and C.~Verikoukis, ``Intent-based management of
  next-generation networks: An {LLM}-centric approach,'' \emph{IEEE Network},
  vol.~38, no.~5, pp. 29--36, 2024.

\bibitem{quan2025large}
H.~Quan, W.~Ni, T.~Zhang, X.~Ye, Z.~Xie, S.~Wang, Y.~Liu, and H.~Song, ``Large
  language model agents for radio map generation and wireless network
  planning,'' \emph{IEEE Networking Letters}, vol.~7, no.~3, pp. 166--170,
  2025.

\bibitem{Sevim2024LLMAssistedWirelessDeployment}
N.~Sevim, M.~Ibrahim, and S.~Ekin, ``Large language models ({LLMs}) assisted
  wireless network deployment in urban settings,'' in \emph{2024 IEEE 100th
  Vehicular Technology Conference (VTC2024-Fall)}, 2024, pp. 1--7.

\bibitem{hribar2026radioenvironmentmappingworld}
J.~Hribar, L.~Milosheski, and R.~Shinkuma, ``Radio environment mapping with
  world models for active measurement control: Should networks dream of optimal
  control?'' in \emph{2026 Joint European Conference on Networks and
  Communications \& 6G Summit (EuCNC/6G Summit)}.\hskip 1em plus 0.5em minus
  0.4em\relax IEEE, 2026, pp. 14--19.

\bibitem{al2024multi}
A.~Al-Tahmeesschi, J.~Talvitie, M.~L{\'o}pez-Ben{\'\i}tez, H.~Ahmadi, and
  L.~Ruotsalainen, ``Multi-objective deep reinforcement learning for {5G} base
  station placement to support localisation for future sustainable traffic,''
  in \emph{2024 Joint European Conference on Networks and Communications \& 6G
  Summit (EuCNC/6G Summit)}.\hskip 1em plus 0.5em minus 0.4em\relax IEEE, 2024,
  pp. 493--498.

\bibitem{el2022energy}
A.~El~Amine, J.-P. Chaiban, H.~A.~H. Hassan, P.~Dini, L.~Nuaymi, and R.~Achkar,
  ``Energy optimization with multi-sleeping control in {5G} heterogeneous
  networks using reinforcement learning,'' \emph{IEEE Transactions on Network
  and Service Management}, vol.~19, no.~4, pp. 4310--4322, 2022.

\bibitem{masrur2025energy}
S.~Masrur, {\.I}.~G{\"u}ven{\c{c}}, and D.~L{\'o}pez-P{\'e}rez,
  ``Energy-efficient sleep mode optimization in {5G} {mmWave} networks via
  multi-agent deep reinforcement learning,'' \emph{IEEE Transactions on Green
  Communications and Networking}, vol.~10, pp. 1495--1511, 2026.

\bibitem{liang2026scalable}
X.~Liang, A.~Al-Tahmeesschi, S.~Chetty, C.~Cavdar, B.~Canberk, and H.~Ahmadi,
  ``Scalable machine learning-based approaches for energy saving in densely
  deployed {Open RAN},'' \emph{IEEE Transactions on Green Communications and
  Networking}, vol.~10, pp. 2710--2722, 2026.

\bibitem{Ali2023Security}
\BIBentryALTinterwordspacing
A.~S. Ali, D.~M. Manias, A.~Shami, and S.~Muhaidat, ``Leveraging large language
  models for {DRL}-based anti-jamming strategies in zero touch networks,''
  arXiv preprint arXiv:2308.09376, 2023. [Online]. Available:
  \url{https://arxiv.org/abs/2308.09376}
\BIBentrySTDinterwordspacing

\bibitem{zhang2025automatedDetection}
W.~Zhang, Q.~Wei, H.~Chen, and Y.~Wang, ``Automated detection of contradictions
  in {5G} network specifications using reinforcement learning-trained small
  {LLM},'' \emph{EURASIP Journal on Wireless Communications and Networking},
  vol. 2025, no.~1, p.~85, 2025.

\bibitem{Sana_2025_TeleLogs}
\BIBentryALTinterwordspacing
M.~Sana, N.~Piovesan, A.~De~Domenico, Y.~Kang, H.~Zhang, M.~Debbah, and
  F.~Ayed, ``Reasoning language models for root cause analysis in {5G} wireless
  networks,'' 2025. [Online]. Available: \url{https://arxiv.org/abs/2507.21974}
\BIBentrySTDinterwordspacing

\bibitem{11618692}
Y.~He, Y.~Yang, L.~Rui, and C.~Wu, ``Hyperbolic spatio-temporal graph learning
  with agent reasoning for root cause localization in cloud-edge
  microservices,'' \emph{IEEE Transactions on Cognitive Communications and
  Networking}, vol.~12, pp. 10\,810--10\,825, 2026.

\bibitem{wang2026fault}
J.~Wang, N.~Gu, K.~Zhang, Y.~Bao, J.~Yang, X.~Yin, Z.~Zhang, Z.~Zhang, L.~Liu,
  Q.~Yu, Y.~Gao, F.~Tan, L.~Huang, and J.~Xie, ``Fault root cause analysis via
  adaptive chain of graph reasoning in wireless networks,'' in \emph{Computing
  and Pattern Recognition (ICCPR)}, ser. Communications in Computer and
  Information Science.\hskip 1em plus 0.5em minus 0.4em\relax Springer Nature
  Singapore, 2026, pp. 443--455.

\bibitem{lewis2020retrieval}
P.~Lewis, E.~Perez, A.~Piktus, F.~Petroni, V.~Karpukhin, N.~Goyal,
  H.~K{\"u}ttler, M.~Lewis, W.-t. Yih, T.~Rockt{\"a}schel, S.~Riedel, and
  D.~Kiela, ``Retrieval-augmented generation for knowledge-intensive {NLP}
  tasks,'' in \emph{Advances in Neural Information Processing Systems
  (NeurIPS)}, vol.~33, 2020, pp. 9459--9474.

\bibitem{sapkota20245g}
B.~Sapkota, R.~Ghimire, P.~Pujara, S.~Ghimire, U.~Shrestha, R.~Ghimire, B.~R.
  Dawadi, and S.~R. Joshi, ``{5G} network deployment planning using
  metaheuristic approaches,'' \emph{Telecom}, vol.~5, no.~3, pp. 588--608,
  2024.

\bibitem{yangyang2004particle}
Y.~Zhang, C.~Ji, P.~Yuan, M.~Li, C.~Wang, and G.~Wang, ``Particle swarm
  optimization for base station placement in mobile communication,'' in
  \emph{Proc. IEEE International Conference on Networking, Sensing and Control
  (ICNSC)}, vol.~1.\hskip 1em plus 0.5em minus 0.4em\relax Taipei, Taiwan:
  IEEE, 2004, pp. 428--432.

\bibitem{10230112}
A.~Leivadeas and M.~Falkner, ``Autonomous network assurance in intent based
  networking: Vision and challenges,'' in \emph{2023 32nd International
  Conference on Computer Communications and Networks (ICCCN)}, 2023, pp. 1--10.

\bibitem{voutsas2024mitigating}
F.~Voutsas, J.~Violos, and A.~Leivadeas, ``Mitigating alert fatigue in cloud
  monitoring systems: A machine learning perspective,'' \emph{Computer
  Networks}, vol. 250, p. 110543, 2024.

\bibitem{li2022analytics}
Y.~Li, D.~Tujkovic, and P.-H. Huang, ``Analytics and machine learning powered
  wireless network optimization and planning,'' \emph{arXiv preprint
  arXiv:2209.06352}, 2022.

\bibitem{bui2017survey}
N.~Bui, M.~Cesana, S.~A. Hosseini, Q.~Liao, I.~Malanchini, and J.~Widmer, ``A
  survey of anticipatory mobile networking: Context-based classification,
  prediction methodologies, and optimization techniques,'' \emph{IEEE
  Communications Surveys \& Tutorials}, vol.~19, no.~3, pp. 1790--1821, 2017.

\bibitem{oran2026oad}
{O-RAN ALLIANCE}, ``{O-RAN} architecture description,'' O-RAN ALLIANCE e.V.,
  Technical Specification O-RAN.WG1.TS.OAD-R005-v17.00, 2026.

\bibitem{haghrah2023survey}
A.~Haghrah, M.~P. Abdollahi, H.~Azarhava, and J.~M. Niya, ``A survey on the
  handover management in {5G}-{NR} cellular networks: aspects, approaches and
  challenges,'' \emph{EURASIP Journal on Wireless Communications and
  Networking}, vol. 2023, no.~1, p.~52, 2023.

\bibitem{tashan2022mobility}
W.~Tashan, I.~Shayea, S.~Aldirmaz-Colak, M.~Ergen, M.~H. Azmi, and
  A.~Alhammadi, ``Mobility robustness optimization in future mobile
  heterogeneous networks: A survey,'' \emph{IEEE Access}, vol.~10, pp.
  45\,522--45\,541, 2022.

\bibitem{Tao2026LargeVisionModel}
Z.~Tao, W.~Xu, and X.~You, ``Large vision model-enhanced digital twin with deep
  reinforcement learning for user association and load balancing in dynamic
  wireless networks,'' \emph{IEEE Journal on Selected Areas in Communications},
  vol.~44, pp. 2718--2732, 2026.

\bibitem{li2026llm}
W.~Li, W.~Chen, B.~Ai, Y.~Zhang, Y.~Peng, and M.~Matthaiou, ``{LLM}-empowered
  wireless channel quality prediction for reliable cell handover,'' \emph{IEEE
  Wireless Communications}, pp. 1--8, 2026, early access.

\bibitem{kumar2025advanced}
T.~S. Kumar, M.~Roslee, J.~Jayapradha, Y.~Ullah, C.~Sudhamani, S.~M.~I. Mitani,
  A.~F. Osman, and F.~Z. Ali, ``Advanced handover optimization ({AHO}) using
  deep reinforcement learning in {5G} networks,'' \emph{Journal of King Saud
  University Computer and Information Sciences}, vol.~37, no.~6, p. 115, 2025.

\bibitem{lien2026foundation}
S.-Y. Lien, Y.-H. Huang, C.-C. Tseng, Y.~Cao, H.-H. Chin, and D.-J. Deng,
  ``Foundation model-based mobility management for {6G} mobile networks,''
  \emph{IEEE Internet of Things Journal}, vol.~13, no.~11, pp.
  24\,185--24\,204, 2026.

\bibitem{rao2025optimization}
A.~Rao, V.~Gowtham, S.~Shivaprasad, F.~Schreiner, M.~Corici, and T.~Magedanz,
  ``Optimization of handover through knowledge graph enabled coordination
  between {LLM} and {O-RAN},'' in \emph{2025 IEEE Future Networks World Forum
  (FNWF)}.\hskip 1em plus 0.5em minus 0.4em\relax IEEE, 2025, pp. 1--7.

\bibitem{vaezi2023deep}
M.~Vaezi, X.~Lin, H.~Zhang, W.~Saad, and H.~V. Poor, ``Deep reinforcement
  learning for interference management in {UAV}-based {3D} networks: Potentials
  and challenges,'' \emph{IEEE Communications Magazine}, vol.~62, no.~2, pp.
  134--140, 2024.

\bibitem{ghassemi2026FoundationModelAided}
M.~Ghassemi, H.~Zhang, A.~Afana, A.~Bin~Sediq, and M.~Erol-Kantarci,
  ``Foundation model-aided hierarchical deep reinforcement learning for
  blockage-aware link in {RIS}-assisted networks,'' in \emph{Proc. IEEE
  International Conference on Communications (ICC)}, 2026, pp. 1--6.

\bibitem{zhou2025prompting}
H.~Zhou, C.~Hu, D.~Yuan, Y.~Yuan, D.~Wu, X.~Liu, and J.~Zhang, ``Prompting
  wireless networks: Reinforced in-context learning for power control,''
  \emph{arXiv preprint arXiv:2506.06526}, 2025.

\bibitem{lotfi2025oran}
F.~Lotfi, H.~Rajoli, and F.~Afghah, ``{ORAN-GUIDE}: {RAG}-driven prompt
  learning for {LLM}-augmented reinforcement learning in {O-RAN} network
  slicing,'' \emph{arXiv preprint arXiv:2506.00576}, 2025.

\bibitem{wu2025llm}
X.~Wu, J.~Farooq, Y.~Wang, and J.~Chen, ``{LLM-xApp}: A large language model
  empowered radio resource management {xApp} for {5G} {O-RAN},'' in
  \emph{Proceedings 2025 Workshop on Security and Privacy of Next-Generation
  Networks (FutureG), co-located with NDSS}.\hskip 1em plus 0.5em minus
  0.4em\relax San Diego, CA, USA: Internet Society, 2025.

\bibitem{Wu2026LlmModelBasedTaskOffloading}
Q.~Wu, Y.~Xie, P.~Fan, D.~Qin, K.~Wang, N.~Cheng, and K.~B. Letaief, ``Large
  language model-based task offloading and resource allocation for digital twin
  edge computing networks,'' \emph{IEEE Transactions on Mobile Computing},
  vol.~25, no.~7, pp. 10\,290--10\,301, 2026.

\bibitem{Xu2026ServingLongContextLLMs}
M.~Xu, D.~Niyato, and C.~G. Brinton, ``Serving long-context {LLMs} at the
  mobile edge: Test-time reinforcement learning-based model caching and
  inference offloading,'' \emph{IEEE Transactions on Networking}, vol.~34, pp.
  3808--3823, 2026.

\bibitem{You2026JPPO}
F.~You, H.~Du, K.~Huang, and A.~Jamalipour, ``{JPPO++}: Joint power and
  denoising-inspired prompt optimization for mobile {LLM} services,''
  \emph{IEEE Transactions on Mobile Computing}, vol.~25, no.~8, pp.
  12\,159--12\,174, 2026.

\bibitem{lu2022reinforcement}
X.~Lu, L.~Xiao, P.~Li, X.~Ji, C.~Xu, S.~Yu, and W.~Zhuang, ``Reinforcement
  learning-based physical cross-layer security and privacy in {6G},''
  \emph{IEEE Communications Surveys \& Tutorials}, vol.~25, no.~1, pp.
  425--466, 2023.

\bibitem{el2025secure}
M.~El-Hajj, ``Secure and trustworthy open radio access network ({O-RAN})
  optimization: A zero-trust and federated learning framework for {6G}
  networks,'' \emph{Future Internet}, vol.~17, no.~6, p. 233, 2025.

\bibitem{behravesh2025ai}
R.~Behravesh, D.~Coroama, M.~Nikroo, M.~Gholami, N.~Agrawala, and R.~N. Mitra,
  ``{AI}-driven traffic steering {xApp} for {Open RAN}: A cloud-native
  quantum-safe approach,'' in \emph{2025 IEEE 11th International Conference on
  Network Softwarization (NetSoft)}.\hskip 1em plus 0.5em minus 0.4em\relax
  IEEE, 2025, pp. 49--54.

\bibitem{Mojdeh2025TowardsSecureOran}
M.~{Karbalaee Motalleb}, C.~Benzaid, T.~Taleb, M.~Katz, V.~Shah-Mansouri, and
  J.~Kim, ``Towards secure intelligent {O-RAN} architecture: Vulnerabilities,
  threats and promising technical solutions using {LLMs},'' \emph{Digital
  Communications and Networks}, vol.~12, no.~6, pp. 916--925, 2026.

\bibitem{soltani2025intelligent}
S.~Soltani, A.~Amanloo, M.~Shojafar, and R.~Tafazolli, ``Intelligent control in
  {6G} open {RAN}: Security risk or opportunity?'' \emph{IEEE Open Journal of
  the Communications Society}, vol.~6, pp. 840--880, 2025.

\bibitem{lacava2025dapps}
A.~Lacava, L.~Bonati, N.~Mohamadi, R.~Gangula, F.~Kaltenberger, P.~Johari,
  S.~D'Oro, F.~Cuomo, M.~Polese, and T.~Melodia, ``{dApps}: Enabling real-time
  {AI}-based open {RAN} control,'' \emph{Computer Networks}, vol. 269, p.
  111342, 2025.

\bibitem{gao2026ai}
Y.~Gao, Z.~Lu, X.~Wu, W.~Yu, S.~Liu, J.~Du, Y.~Jin, S.~Zhang, X.~Chu, and
  S.~Xu, ``{AI}-driven channel state information ({CSI}) extrapolation for
  {6G}: Current situations, challenges, and future research,'' \emph{IEEE
  Communications Surveys \& Tutorials}, vol.~28, pp. 4485--4518, 2026.

\bibitem{3gpp38211}
{3rd Generation Partnership Project (3GPP)}, ``{NR; Physical Channels and
  Modulation},'' 3GPP, Technical Specification TS 38.211, V19.1.0, Sep. 2025,
  release 19.

\bibitem{bajracharya2026achieving}
R.~Bajracharya and R.~Shrestha, ``Achieving ultra-reliable low latency
  communication in {5G} and beyond,'' \emph{Sensors}, vol.~26, no.~11, p. 3485,
  2026.

\bibitem{salehi2025towards}
F.~Salehi, A.~Mahmood, S.~F. Abedin, K.~Thar, and M.~Gidlund, ``Towards
  ultra-reliable {6G} in-{X} subnetworks: Dynamic link adaptation by deep
  reinforcement learning,'' \emph{arXiv preprint arXiv:2507.12031}, 2025.

\bibitem{farhadi2025ai}
H.~Farhadi, P.~Gowgi, and D.~Sandberg, ``{AI}/{ML}-driven beamforming with
  imperfect channel state information for multi-user {MIMO} transmission,'' in
  \emph{2025 IEEE International Conference on Acoustics, Speech, and Signal
  Processing Workshops (ICASSPW)}.\hskip 1em plus 0.5em minus 0.4em\relax IEEE,
  2025, pp. 1--5.

\bibitem{ismail2026ai}
M.~I. Ismail, R.~Elsayed, M.~Ismail, M.~I. Ibrahem, Z.~M. Fadlullah, and
  M.~Fouda, ``{AI}-driven secure {RIS} beam prediction for {O-RAN}-enabled
  {B5G} networks,'' in \emph{2026 IEEE 2nd International Conference on Secure
  IoT, Assured and Trusted Computing (SATC)}.\hskip 1em plus 0.5em minus
  0.4em\relax IEEE, 2026, pp. 1--7.

\bibitem{lee2026llm}
W.~Lee and J.~Park, ``{LLM}-empowered resource allocation in wireless
  communications systems,'' \emph{IEEE Access}, vol.~14, pp. 15\,260--15\,272,
  2026.

\bibitem{he2026agentic}
Z.~He, Y.~Luo, X.~Liu, M.~B. Mashhadi, M.~Shojafar, M.~Debbah, and
  R.~Tafazolli, ``Agentic {AI-RAN}: Enabling intent-driven, explainable and
  self-evolving open {RAN} intelligence,'' \emph{IEEE Network}, pp. 1--10,
  2026, early access.

\bibitem{elkael2026agentran}
M.~Elkael, S.~D'Oro, L.~Bonati, M.~Polese, Y.~Lee, K.~Furueda, and T.~Melodia,
  ``{AgentRAN}: An agentic {AI} architecture for autonomous control of open
  {6G} networks,'' \emph{IEEE Communications Magazine}, pp. 1--7, 2026, early
  access.

\bibitem{lotfi2025meta}
F.~Lotfi and F.~Afghah, ``Meta hierarchical reinforcement learning for scalable
  resource management in {O-RAN},'' \emph{arXiv preprint arXiv:2512.13715},
  2025.

\bibitem{aroua2026ran}
S.~Aroua, A.~I. Aravanis, I.~Chatzistefanidis, H.~Abbar, A.-K. Dang,
  A.~Giovanidis, S.-E. {El Ayoubi}, S.~Senecal, M.~V. Konchylaki, and
  N.~Nikaein, ``From {RAN} control to agentic intelligence: Architecture and
  vision for energy efficient {AI-RAN},'' \emph{arXiv preprint
  arXiv:2606.21955}, 2026.

\bibitem{Bonati2021Colosseum}
L.~Bonati, P.~Johari, M.~Polese, S.~D'Oro, S.~Mohanti, M.~Tehrani-Moayyed,
  D.~Villa, S.~Shrivastava, C.~Tassie, K.~Yoder, A.~Bagga, P.~Patel, V.~Petkov,
  M.~Seltser, F.~Restuccia, A.~Gosain, K.~R. Chowdhury, S.~Basagni, and
  T.~Melodia, ``{Colosseum}: Large-scale wireless experimentation through
  hardware-in-the-loop network emulation,'' in \emph{2021 IEEE International
  Symposium on Dynamic Spectrum Access Networks (DySPAN)}.\hskip 1em plus 0.5em
  minus 0.4em\relax Los Angeles, CA, USA: IEEE, 2021, pp. 105--113.

\bibitem{Maatouk_2026_TeleQnA}
A.~Maatouk, F.~Ayed, N.~Piovesan, A.~{De Domenico}, M.~Debbah, and Z.-Q. Luo,
  ``{TeleQnA}: A benchmark dataset to assess large language models
  telecommunications knowledge,'' \emph{IEEE Network}, vol.~40, no.~2, pp.
  253--260, 2026.

\bibitem{karim2023Spec5G}
I.~Karim, K.~S. Mubasshir, M.~M. Rahman, and E.~Bertino, ``{SPEC5G}: A dataset
  for {5G} cellular network protocol analysis,'' in \emph{Findings of the
  Association for Computational Linguistics: IJCNLP-AACL 2023
  (Findings)}.\hskip 1em plus 0.5em minus 0.4em\relax Nusa Dua, Bali:
  Association for Computational Linguistics, 2023, pp. 20--38.

\bibitem{Nikbakht2024TSpecLlm}
R.~Nikbakht, M.~Benzaghta, and G.~Geraci, ``{TSpec-LLM}: An open-source dataset
  for {LLM} understanding of {3GPP} specifications,'' in \emph{2024 IEEE
  Globecom Workshops (GC Wkshps)}, 2024, pp. 1--6.

\bibitem{maatouk2025TeleLlm}
A.~Maatouk, K.~Chirino~Ampudia, R.~Ying, and L.~Tassiulas, ``Tele-{LLMs}: A
  series of specialized large language models for telecommunications,''
  \emph{IEEE Access}, vol.~14, pp. 86\,424--86\,441, 2026.

\bibitem{Jiang2026CommGPT}
F.~Jiang, W.~Zhu, L.~Dong, K.~Wang, K.~Yang, C.~Pan, and O.~A. Dobre,
  ``{CommGPT}: A graph and retrieval-augmented multimodal communication
  foundation model,'' \emph{IEEE Communications Magazine}, vol.~64, no.~3, pp.
  46--52, 2026.

\bibitem{Alkhateeb2019DeepMimo}
A.~Alkhateeb, ``{DeepMIMO}: A generic deep learning dataset for millimeter wave
  and massive {MIMO} applications,'' in \emph{Proc. Information Theory and
  Applications Workshop (ITA)}, San Diego, CA, 2019, pp. 1--8.

\bibitem{Alkhateeb2023DeepSense6G}
A.~Alkhateeb, G.~Charan, T.~Osman, A.~Hredzak, J.~Morais, U.~Demirhan, and
  N.~Srinivas, ``{DeepSense} {6G}: A large-scale real-world multi-modal sensing
  and communication dataset,'' \emph{IEEE Communications Magazine}, vol.~61,
  no.~9, pp. 122--128, 2023.

\bibitem{Barlacchi2015MilanDataset}
\BIBentryALTinterwordspacing
G.~Barlacchi, M.~De~Nadai, R.~Larcher, A.~Casella, C.~Chitic, G.~Torrisi,
  F.~Antonelli, A.~Vespignani, A.~Pentland, and B.~Lepri, ``A multi-source
  dataset of urban life in the city of {Milan} and the {Province of
  Trentino},'' \emph{Scientific Data}, vol.~2, no.~1, p. 150055, 2015.
  [Online]. Available: \url{https://doi.org/10.1038/sdata.2015.55}
\BIBentrySTDinterwordspacing

\bibitem{Orlando2023NetMob23Dataset}
\BIBentryALTinterwordspacing
O.~E. Mart{\'i}nez-Durive, S.~Mishra, C.~Ziemlicki, S.~Rubrichi, Z.~Smoreda,
  and M.~Fiore, ``The {NetMob23} dataset: A high-resolution multi-region
  service-level mobile data traffic cartography,'' 2023. [Online]. Available:
  \url{https://arxiv.org/abs/2305.06933}
\BIBentrySTDinterwordspacing

\bibitem{Bonati2023OpenRANGym}
\BIBentryALTinterwordspacing
L.~Bonati, M.~Polese, S.~D'Oro, S.~Basagni, and T.~Melodia, ``{OpenRAN Gym}:
  {AI/ML} development, data collection, and testing for {O-RAN} on {PAWR}
  platforms,'' \emph{Computer Networks}, vol. 220, p. 109502, 2023. [Online].
  Available:
  \url{https://www.sciencedirect.com/science/article/pii/S1389128622005369}
\BIBentrySTDinterwordspacing

\bibitem{Polese2023ColORan}
M.~Polese, L.~Bonati, S.~D'Oro, S.~Basagni, and T.~Melodia, ``{ColO-RAN}:
  Developing machine learning-based {xApps} for {Open RAN} closed-loop control
  on programmable experimental platforms,'' \emph{IEEE Transactions on Mobile
  Computing}, vol.~22, no.~10, pp. 5787--5800, Oct. 2023.

\bibitem{Bonati2024Twinning}
L.~Bonati, R.~Shirkhani, C.~Fiandrino, S.~Maxenti, S.~D'Oro, M.~Polese, and
  T.~Melodia, ``Twinning commercial network traces on experimental open {RAN}
  platforms,'' in \emph{Proc. 18th ACM Workshop on Wireless Network Testbeds,
  Experimental Evaluation and Characterization (WiNTECH), co-located with ACM
  MobiCom}, 2024, pp. 1914--1921.

\bibitem{Gajjar_2025_ORAN_Bench_13K}
P.~Gajjar and V.~K. Shah, ``{ORAN-Bench-13K}: An open source benchmark for
  assessing {LLMs} in open radio access networks,'' in \emph{2025 IEEE 22nd
  Consumer Communications \& Networking Conference (CCNC)}, 2025, pp. 1--4.

\bibitem{Colle_2026_TeleMath}
V.~Colle, M.~Sana, N.~Piovesan, A.~De~Domenico, F.~Ayed, and M.~Debbah,
  ``{TeleMath}: A benchmark for large language models in telecom mathematical
  problem solving,'' \emph{IEEE Network}, pp. 1--7, 2026.

\bibitem{Ezzakri_2025_TeleTables}
\BIBentryALTinterwordspacing
A.~Ezzakri, N.~Piovesan, M.~Sana, A.~{De Domenico}, F.~Ayed, and H.~Zhang,
  ``{TeleTables}: A benchmark for large language models in telecom table
  interpretation,'' 2025, arXiv:2601.04202v1. [Online]. Available:
  \url{https://arxiv.org/abs/2601.04202v1}
\BIBentrySTDinterwordspacing

\bibitem{Gajjar_2025_ORANSight2}
P.~Gajjar and V.~K. Shah, ``{ORANSight-2.0}: Foundational {LLMs} for {O-RAN},''
  \emph{IEEE Transactions on Machine Learning in Communications and
  Networking}, vol.~3, pp. 903--920, 2025.

\bibitem{Xiao2026TeleComBench}
J.~Xiao, Y.~Lin, H.~Qiu, R.~Ma, C.~Zhong, D.~Xu, X.~Long, Q.~Hao, D.~Zou,
  Z.~Yang, Y.~Gao, and F.~Tan, ``{TeleCom-Bench}: How far are large language
  models from industrial telecommunication applications?'' in \emph{Proc. 32nd
  ACM SIGKDD Conference on Knowledge Discovery and Data Mining (KDD)}, 2026,
  pp. 10\,056--10\,067.

\bibitem{Cheng2025CsiRrg}
S.~Cheng, R.~Zandehshahvar, H.~Zhao, D.~A. Garcia-Ulloa, A.~Villena-Rodriguez,
  C.~N. Manch{\'o}n, and P.~Van~Hentenryck, ``{CSI-4CAST}: A hybrid deep
  learning model for {CSI} prediction with comprehensive robustness and
  generalization testing,'' \emph{IEEE Transactions on Machine Learning in
  Communications and Networking}, vol.~4, pp. 882--912, 2026.

\bibitem{Feng2025TelecomTS}
\BIBentryALTinterwordspacing
A.~Feng, A.~Varvarigos, I.~Panitsas, D.~Fernandez, J.~Wei, Y.~Guo, J.~Chen,
  A.~Maatouk, L.~Tassiulas, and R.~Ying, ``{TelecomTS}: A multi-modal
  observability dataset for time series and language analysis,'' in
  \emph{Proceedings of the 43rd International Conference on Machine Learning
  (ICML)}, 2026. [Online]. Available: \url{https://arxiv.org/abs/2510.06063}
\BIBentrySTDinterwordspacing

\bibitem{hoydis2023sionnaopensource}
\BIBentryALTinterwordspacing
J.~Hoydis, S.~Cammerer, F.~{Ait Aoudia}, A.~Vem, N.~Binder, G.~Marcus, and
  A.~Keller, ``Sionna: An open-source library for next-generation physical
  layer research,'' 2022. [Online]. Available:
  \url{https://arxiv.org/abs/2203.11854}
\BIBentrySTDinterwordspacing

\bibitem{Jakob2023SionnaRt}
J.~Hoydis, F.~A. Aoudia, S.~Cammerer, M.~Nimier-David, N.~Binder, G.~Marcus,
  and A.~Keller, ``{Sionna} {RT}: Differentiable ray tracing for radio
  propagation modeling,'' in \emph{2023 IEEE Globecom Workshops (GC Wkshps)},
  2023, pp. 317--321.

\bibitem{nvidia_aodt}
{NVIDIA Corporation}, ``Aerial {O}mniverse {D}igital {T}win ({AODT})
  documentation,'' [Online]. Available:
  \url{https://docs.nvidia.com/aerial/aodt/}, accessed: Oct. 2026.

\bibitem{Lacava2023nsOran}
A.~Lacava, M.~Bordin, M.~Polese, R.~Sivaraj, T.~Zugno, F.~Cuomo, and
  T.~Melodia, ``{ns-O-RAN}: Simulating {O-RAN} {5G} systems in {ns-3},'' in
  \emph{Proceedings of the 2023 Workshop on {ns-3} ({WNS3} '23)}.\hskip 1em
  plus 0.5em minus 0.4em\relax Arlington, VA, USA: ACM, 2023, pp. 35--44.

\bibitem{Bonati2021Scope}
L.~Bonati, S.~D'Oro, S.~Basagni, and T.~Melodia, ``{SCOPE}: An open and
  softwarized prototyping platform for {NextG} systems,'' in \emph{Proc. 19th
  Annual International Conference on Mobile Systems, Applications, and Services
  (MobiSys '21)}.\hskip 1em plus 0.5em minus 0.4em\relax ACM, 2021, pp.
  415--426.

\bibitem{Bordin2025NsOranGym}
M.~Bordin, A.~Lacava, M.~Polese, S.~Satish, M.~A. Nittoor, R.~Sivaraj,
  F.~Cuomo, and T.~Melodia, ``Design and evaluation of deep reinforcement
  learning for energy saving in open {RAN},'' in \emph{Proc. IEEE 22nd Consumer
  Communications \& Networking Conference (CCNC)}, 2025, pp. 1--6.

\bibitem{nokia-wireless-suite}
{Nokia}, ``{Wireless Suite: A Collection of Problems in Wireless
  Telecommunications},'' \url{https://github.com/nokia/wireless-suite}, 2020,
  bSD 3-Clause; repository archived (read-only) 14 Sep. 2026; accessed Sep.
  2026.

\bibitem{forlorn}
V.~Edvardsen, G.~Spreemann, and J.~Van~den Abeele, ``{FORLORN: A Framework for
  Comparing Offline Methods and Reinforcement Learning for Optimization of RAN
  Parameters},'' in \emph{Proc. 18th ACM International Symposium on QoS and
  Security for Wireless and Mobile Networks (Q2SWinet)}, 2022, pp. 37--44.

\bibitem{Haider2024NetworkGym}
M.~Haider, M.~Yin, M.~Zhang, A.~Gupta, J.~Zhu, and Y.-X. Wang, ``{NetworkGym}:
  Reinforcement learning environments for multi-access traffic management in
  network simulation,'' in \emph{Advances in Neural Information Processing
  Systems (NeurIPS)}, 2024, pp. 106\,628--106\,648.

\bibitem{edgeric}
W.-H. Ko, U.~Ghosh, U.~Dinesha, R.~Wu, S.~Shakkottai, and D.~Bharadia,
  ``{EdgeRIC}: Empowering real-time intelligent optimization and control in
  {NextG} cellular networks,'' in \emph{Proc. 21st USENIX Symp. on Networked
  Systems Design and Implementation (NSDI)}.\hskip 1em plus 0.5em minus
  0.4em\relax USENIX Association, 2024, pp. 1315--1330,
  \url{https://github.com/ucsdwcsng/EdgeRIC-A-real-time-RIC}; accessed Sep.
  2026, commit 6780868.

\bibitem{GsmaOpenTelco}
{GSMA Labs}, ``{Open Telco}: Telco-specific {LLM} benchmarks,''
  \url{https://github.com/gsma-labs/evals}, 2025, evaluation harness built on
  Inspect AI (MIT); leaderboard at
  \url{https://huggingface.co/datasets/GSMA/leaderboard} (Apache-2.0). Accessed
  3 September 2026.

\bibitem{Tsampazi2024PandORA}
M.~Tsampazi, S.~D'Oro, M.~Polese, L.~Bonati, G.~Poitau, M.~Healy, M.~Alavirad,
  and T.~Melodia, ``{PandORA}: Automated design and comprehensive evaluation of
  deep reinforcement learning agents for {Open RAN},'' \emph{IEEE Transactions
  on Mobile Computing}, vol.~24, no.~4, pp. 3223--3240, 2025.

\bibitem{NsOranGymRepo}
{WiNES Lab}, ``{ns-O-RAN Gym}: A {Gymnasium} environment for {ns-O-RAN},''
  \url{https://github.com/wineslab/ns-o-ran-gym}, 2025, {GPL-3.0}; PyPI package
  nsoran; accessed 3 September 2026.

\bibitem{3gpp28105}
{3GPP}, ``Management and orchestration; artificial intelligence/machine
  learning ({AI/ML}) management,'' 3rd Generation Partnership Project (3GPP),
  Technical Specification TS 28.105, 2026, v20.0.0, Release 20.

\bibitem{3gpp28908}
------, ``Study on {Artificial Intelligence/Machine Learning (AI/ML)}
  management,'' 3rd Generation Partnership Project (3GPP), Technical Report TR
  28.908, 2024, release 18.

\bibitem{3gpp38843}
------, ``Study on {Artificial Intelligence (AI)/Machine Learning (ML)} for
  {NR} air interface,'' 3rd Generation Partnership Project (3GPP), Technical
  Report TR 38.843, Dec. 2023, release 18.

\bibitem{Chen2026SixGAgentGym}
\BIBentryALTinterwordspacing
J.~Chen, J.~Tang, X.~Yang, and Z.~Lv, ``{6GAgentGym}: Tool use, data synthesis,
  and agentic learning for network management,'' 2026. [Online]. Available:
  \url{https://arxiv.org/abs/2603.29656}
\BIBentrySTDinterwordspacing

\end{thebibliography}
